\documentclass[10pt]{article}
\usepackage{neurips_preprint}

\usepackage{amsmath,amssymb,amsthm,mathtools,bm,booktabs,tabularx}
\usepackage{microtype}
\usepackage{needspace}
\usepackage{float} 
\usepackage{xcolor}
\usepackage[hidelinks]{hyperref}
\usepackage{url}
\hypersetup{pdftitle={Prescribed Bayesian Efficiency under Lifetime Adjustment Costs}}
\allowdisplaybreaks[1]
\newcommand{\E}{\mathbb{E}}
\newcommand{\Pp}{\mathbb{P}}
\newcommand{\R}{\mathbb{R}}
\newcommand{\N}{\mathbb{N}}
\newcommand{\G}{\mathcal{G}}
\newcommand{\F}{\mathcal{F}}
\newcommand{\CC}{\mathcal{C}}
\newcommand{\HH}{\mathcal{H}}
\newcommand{\EE}{\mathcal{E}}
\newcommand{\Alloc}{\mathcal{P}}
\newcommand{\Dual}{\mathcal{D}}
\newcommand{\A}{\mathbf{A}}
\newcommand{\B}{\mathbf{B}}
\newcommand{\bb}{\mathbf{b}}
\newcommand{\X}{\mathbf{X}}
\newcommand{\bhat}{\widehat{\bb}}
\newcommand{\iter}{\mathbf{v}}
\newcommand{\winit}{\mathbf{w}_{\mathrm{init}}}
\newcommand{\radmed}{R^{\mathrm{med}}}
\newcommand{\dep}{d^{\mathrm{ep}}}
\newcommand{\wnorm}{\mathbf{u}}

\newcommand{\W}{\mathbf{W}}
\newcommand{\U}{\mathbf{U}}
\newcommand{\Uhat}{\widehat{\mathbf{U}}}
\newcommand{\Svec}{\mathbf{S}}
\newcommand{\Tvec}{\mathbf{T}}
\newcommand{\V}{\mathbf{V}}
\newcommand{\Qvec}{\mathbf{Q}}
\newcommand{\w}{\mathbf{w}}
\newcommand{\x}{\mathbf{x}}
\newcommand{\y}{\mathbf{y}}
\newcommand{\uvec}{\mathbf{u}}
\newcommand{\vvec}{\mathbf{v}}
\newcommand{\hvec}{\mathbf{h}}
\newcommand{\evec}{\mathbf{e}}
\newcommand{\Id}{\mathbf{I}}
\newcommand{\xiobs}{\bm{\xi}}
\newcommand{\ind}{\mathbf{1}}
\newcommand{\norm}[1]{\lVert#1\rVert}
\newcommand{\ip}[2]{\langle#1,#2\rangle}
\DeclareMathOperator{\ri}{ri}
\DeclareMathOperator{\rbd}{rbd}
\DeclareMathOperator{\dist}{dist}
\DeclareMathOperator{\diam}{diam}
\DeclareMathOperator{\aff}{aff}
\DeclareMathOperator{\supp}{supp}
\DeclareMathOperator{\Var}{Var}
\DeclareMathOperator{\Reg}{Reg}
\DeclareMathOperator{\Proj}{Proj}
\newcommand{\Law}{\mathcal{L}}
\newcommand{\phys}{\mathrm{phys}}

\newcommand{\good}{\mathcal{J}_{\mathrm{good}}}
\newcommand{\loc}{\mathsf{L}}
\newcommand{\ui}{\omega_{\mathrm{UI}}}
\newcommand{\trmod}{\omega_{\mathrm{tr}}}
\newcommand{\Ctwo}{C_{\mathrm{act},2}}
\DeclareMathOperator*{\argmin}{arg\,min}
\DeclareMathOperator{\Unif}{Unif}
\DeclareMathOperator{\Bernoulli}{Bernoulli}
\DeclareMathOperator{\Beta}{Beta}
\DeclareMathOperator{\interior}{int}
\newcommand{\Jac}{\mathrm{D}}
\newcommand{\DiniOne}{\mathcal I_1}
\newcommand{\DiniHalf}{\mathcal I_{1/2}}
\newtheorem{theorem}{Theorem}
\newtheorem{lemma}[theorem]{Lemma}
\newtheorem{proposition}[theorem]{Proposition}
\newtheorem{corollary}[theorem]{Corollary}
\newtheorem{assumption}{Assumption}
\theoremstyle{definition}

\newcommand{\Hgrad}{H_{\nabla}}
\newcommand{\dd}{\mathrm{d}}
\newcommand{\Borel}{\mathcal{B}}
\newcommand{\feasible}{\mathcal{A}}
\newcommand{\allocvec}{\mathbf{a}}

\title{Prescribed Bayesian Efficiency under Lifetime Adjustment Costs}
\author{Viet Bac Nguyen\\
Institute for Artificial Intelligence\\
VNU-University of Engineering and Technology\\
Hanoi, Vietnam\\
\texttt{bacngv@vnu.edu.vn}}
\hypersetup{pdfauthor={Viet Bac Nguyen}}

\begin{document}
\maketitle
\raggedbottom 
\begin{abstract}
A Bayesian learner may retain an issued action while updating its posterior, trading predictive loss against adjustment charges. We assume population curvature, boundary control, an order-$1/n$ lower bound on expected posterior uncertainty about the optimal action, and an observable unbiased gradient witness whose squared norms are uniformly integrable. Sample losses may be unbounded or nonconvex. Cumulative expected excess measures the loss increase relative to current predictive Bayes actions. For bounded nonnegative nondecreasing charges $c$ with $c(0)=0$ and nondecreasing profiles $G:\N\to[1,\infty)$ with $G(J)=o(J)$, we characterize joint attainability of cumulative expected excess $O(G(J))$ at horizons $T=2^J$ and finite expected lifetime charge. Here $J$ indexes logarithmic time. The criterion is finiteness of a prefix-constrained allocation of tracking-error allowances across action-distance scales, equivalently its decreasing-weight dual. Adaptive timing and private randomization do not enlarge the attainable region. For each feasible pair, a horizon-independent deterministic checkpoint policy attains the profile with lifetime-cost moments of every positive order. For excess $O((\log T)^r)$, $0<r<1$, every policy needs $\Omega((\log T)^{2-r})$ revisions in expectation, and a charge-independent schedule achieves a matching pathwise upper bound for every feasible charge. We verify the assumptions for logistic decision models with nondegenerate covariates, finite second moments and a locally positive smooth prior density, including fixed-ridge models with unbounded parameter priors.
\end{abstract}

\section{Introduction}\label{sec:intro}
How accurately can a Bayesian learner track its changing optimal decision while paying a finite expected adjustment cost over its entire lifetime? The posterior updates after every observation. Retaining the issued action avoids an adjustment charge but can increase predictive loss.

Consider $p\sim\Unif[0,1]$ and conditionally iid observations $Y_t\mid p\sim\Bernoulli(p)$. Write $S_n=\sum_{t=1}^nY_t$, with $S_0=0$. Under squared loss $f_y(w)=(w-y)^2$ on $[0,1]$, the predictive Bayes action after $n$ observations is $B_n=(S_n+1)/(n+2)$. Initially issue $A_0=B_0=1/2$. After $Y_1=1$, retaining $A_1=1/2$ incurs excess $(1/2-2/3)^2=1/36$ and no adjustment charge. If $c(x)$ denotes the charge for a change of distance $x$, issuing $A_1=2/3$ removes the excess and costs $c(1/6)$. Both choices use the same updated posterior. Here the Bayes action is also the posterior mean of $p$, a property of squared loss that does not extend to general decision losses.

More generally, a policy $A=(\A_n)_{n\ge0}$ issues $\A_n$ after $n$ observations and before observation $n+1$. Let $\B_n$ be the current predictive Bayes action. Write $\EE_T(A)$ for the sum, over the first $T$ predictions, of the expected loss gaps between $\A_n$ and $\B_n$. The random lifetime charge is $V_{c,\infty}(A)=\sum_{n\ge1}c(\norm{\A_n-\A_{n-1}})$. Fix a finite-valued nondecreasing target $G:\{1,2,\ldots\}\to[1,\infty)$ with $G(J)=o(J)$. We seek one infinite policy with $\EE_{2^J}(A)=O(G(J))$ and $\E V_{c,\infty}(A)<\infty$. The dyadic horizon $2^J$ makes $G$ a profile in logarithmic time. The condition $G(J)=o(J)$ selects policies whose additional predictive risk is negligible on the logarithmic scale of the Bayesian benchmark. The comparator is the predictive Bayes action conditional on observations available before the next prediction, rather than a latent-state oracle or a hindsight fit. Section~\ref{sec:performance} defines these criteria and explains when sublogarithmic excess preserves the leading risk of the Bayes policy.

Theorem~\ref{thm:main} shows that adaptive choice of revision times does not enlarge the attainable region: whenever any admissible policy reaches the prescribed excess profile at finite expected lifetime cost, a fixed sequence of checkpoint times does so too. At each checkpoint, the policy issues the current predictive Bayes action. The criterion is a deterministic allocation of tracking-error allowances across action-distance scales. More retention blocks reduce excess on each doubling epoch, but increase the number of potentially charged revisions. The allocation enforces this reciprocal tradeoff at every prefix of the profile.

The converse applies to every admissible policy, including adaptive timing and independent private randomization. It forces revisions on deterministic windows where the Bayes action moves with positive probability. The construction obtains checkpoint times from a feasible allocation and controls large jumps separately.

For the Bernoulli example with linear adjustment charges, always issuing the current Bayes action has divergent expected total movement. A deterministic retention schedule can instead have both finite expected lifetime movement and bounded cumulative expected excess. Section~\ref{sec:calibration} gives the exact comparison.

Section~\ref{sec:consequences} derives the Dini threshold $\int_0^1c(x)\,\dd x/x<\infty$ for some sublogarithmic excess profile and the square-root threshold $\int_0^1\sqrt{c(x)}\,\dd x/x<\infty$ for bounded total excess. These conditions test the decay of charges near zero. For $c_\alpha(x)=[\log(e/x)]^{-\alpha}$ on $0<x\le1$, with $c_\alpha(0)=0$ and $\alpha>0$, excess $O((\log T)^r)$, $0<r<1$, is feasible exactly when $\alpha+r>2$. Section~\ref{sec:consequences} also gives critical corrections and revision orders. Section~\ref{sec:applications} verifies logistic models and calibrates squared loss. The proof is in Section~\ref{sec:proof} and the appendices.

\subsection{Related work}
\paragraph{Costly adjustment and switching budgets.}
\citet{feldman1995} study infinite-horizon learning with nonvanishing adjustment costs. \citet{agrawal1988} consider bandits, where actions also determine acquired information. \citet{cohen2019} study fixed-charge retention relative to a changing Bayesian estimate in a linear Gaussian Kalman--Bucy model. Their preprint appendix also gives a discrete Bernoulli illustration based on the sample mean. Their small-cost asymptotics yield inaction bands and a testing interpretation. \citet{baley2025} distinguish belief updating from forecast revision under fixed revision costs and strategic alignment. Our observations are exogenous, the monotone charge function is fixed, and the requirement concerns finite expected cost over the whole infinite lifetime.

Switching-budget guarantees in online learning typically compare with a fixed hindsight or population action. \citet{altschuler2018} study finite action sets. For continuous actions, the corrected lazy online convex optimization analysis of \citet{sherman2023} provides logarithmically many switches with squared-logarithmic stochastic strongly convex pseudo-regret. Further lazy optimization guarantees are given by \citet{agarwal2023} and \citet{agarwal2024}, while \citet{wang2021} study a continuous movement budget. Here the comparator evolves with the posterior, and one infinite policy must meet every prefix of a prescribed profile $G$ satisfying $G(J)=o(J)$.

\paragraph{Tracking, Bayesian risk and deterministic allocation.}
\citet{fukasawa2014} derives sharp error--intervention tradeoffs for stochastic-integral discretization. \citet{cai2017} reduce tracking lower bounds to deterministic linear programming. These asymptotic refinement or small-cost results differ from a finiteness criterion for fixed irregular monotone charges and arbitrary sublinear profiles.

Minimum excess risk compares data-informed prediction with a latent-state oracle, as studied by \citet{xu2022} and \citet{hafez2021}. Our excess is the additional gap caused by retaining an issued action. \citet{ohn2026} studies approximate posterior prediction, whereas our posterior remains exact. The statistical proof combines stochastic approximation \citep{moulines2011}, robust independent-block aggregation \citep{minsker2015} and local information inequalities \citep{gill1995}.

Separable convex optimization with ascending constraints is studied by \citet{padakandla2009}. The finite allocation dual follows from Lagrange duality \citep[Chapter~5]{boyd2004}. We connect the infinite allocation program to feasibility over all admissible Bayesian policies and construct one retention schedule for each feasible pair.

\section{Model and main characterization}\label{sec:model}
\paragraph{Notation.}
Write $\R$ for the real numbers and $\N=\{1,2,\ldots\}$. Vectors and matrices are written in bold, except for the abstract random elements $\theta$ and $Z$. Scalar examples use plain type. Vectors are columns, $\ip{\x}{\y}=\x^\top\y$, and $\norm{\cdot}$ is the Euclidean norm. The operators $\ri$, $\rbd$ and $\aff$ denote relative interior, relative boundary and affine hull. For nonempty $S$, $\dist(\x,S)=\inf_{\y\in S}\norm{\x-\y}$. The symbol $\ind_E$ denotes the indicator of an event $E$.

\subsection{Bayesian decisions and information}
Let $\theta\sim\pi$ take values in a standard Borel space $\Theta$. Conditional on $\theta$, observations $Z_1,Z_2,\ldots$ are iid from a measurable kernel $P_\theta$ on a standard Borel space $\mathcal Z$. Their law does not depend on the issued actions. Write $(\Omega,\mathcal A,\Pp)$ for the joint Bayesian experiment, including the independent policy randomization introduced below, and $\E$ for its expectation. Write $\G_n=\sigma(Z_1,\ldots,Z_n)$, with trivial $\G_0$, and $\Pi_n=\Law(\theta\mid\G_n)$ for a regular conditional posterior, where $\Law$ denotes distribution.

The action set $K$ is nonempty, compact and convex, with $D_K:=\diam K\le1$ and positive finite affine dimension $d$. Identify $\aff(K)$ isometrically with $\R^d$. Gradients, interiors and boundaries are relative to this hull. The diameter bound fixes the unit of action distance, so every adjustment lies in the charge domain $[0,1]$. Section~\ref{sec:applications} gives the conversion from physical coordinates.

For loss $f_z(\w)$, define the population and predictive risks by
\[
 R_\theta(\w)=\int f_z(\w)P_\theta(\dd z),\qquad
 r_n(\w)=\int R_u(\w)\Pi_n(\dd u).
\]
The assumptions below guarantee finite risks and unique measurable minimizers. Three actions have different roles:
\[
 \bb_\theta=\argmin_{\w\in K}R_\theta(\w),\qquad
 \bar{\bb}_n=\E[\bb_\theta\mid\G_n],\qquad
 \B_n=\argmin_{\w\in K}r_n(\w).
\]
The latent optimum $\bb_\theta$ minimizes population risk. Its posterior mean $\bar{\bb}_n$ estimates that optimum under squared distance. The predictive Bayes action $\B_n$ minimizes next-observation loss, since $r_n(\w)=\E[f_{Z_{n+1}}(\w)\mid\G_n]$ for fixed $\w$. In general $\B_n\ne\bar{\bb}_n$: the minimizer of an averaged risk need not equal the average of its minimizers. For example, consider scalar quadratic risks $R_\theta(w)=a_\theta(w-b_\theta)^2+\gamma_\theta$ on a compact interval, with $b_\theta$ in that interval, a bounded measurable coefficient $a_\theta>0$, and an integrable action-independent offset $\gamma_\theta$. The first-order equation gives \(r_n'(w)=2\{w\E[a_\theta\mid\G_n]-\E[a_\theta b_\theta\mid\G_n]\}\). The zero of this derivative is a weighted mean with positive weights and therefore lies in the action interval. It consequently minimizes the predictive risk over that interval. Hence \(r_n'(B_n)=0\), and we obtain \(B_n=\E[a_\theta b_\theta\mid\G_n]/\E[a_\theta\mid\G_n]\), whereas \(\bar b_n=\E[b_\theta\mid\G_n]\).
Equal curvatures recover the posterior mean. For a posterior assigning probability $1/2$ to each of $(a_\theta,b_\theta)=(1,0)$ and $(2,1)$, however, $B_n=2/3$ whereas $\bar b_n=1/2$.

A private random seed $\Xi$, called the private tape, has law $P_\Xi$ and is independent of $(\theta,Z_1,Z_2,\ldots)$. It may contain a sequence of random draws that the policy uses adaptively. Put $\F_n=\G_n\vee\sigma(\Xi)$. Independence gives $\Law(\theta\mid\F_n)=\Pi_n$ and $\E[\bb_\theta\mid\F_n]=\bar{\bb}_n$.
The operators $\E_\theta$ and $\Pp_\theta$ condition the joint prior--data--tape law on $\theta$, whereas $P_\theta(\dd z)$ is the sampling kernel for one observation. Conditioning on $(\theta,\F_n)$ means conditioning on $\sigma(\theta)\vee\F_n$. Appendix~\ref{app:kernels} proves the conditioning identities, including evaluation at random actions.

An admissible policy $A=(\A_n)_{n\ge0}$ issues a $K$-valued, $\F_n$-measurable action after the first $n$ observations and before $Z_{n+1}$. It is defined for all $n$ and does not receive an evaluation horizon as input. The issued action can differ from both posterior-based actions above. Write $B=(\B_n)_{n\ge0}$ for the policy that always issues the current Bayes action.

A deterministic-checkpoint retention policy fixes integer times \(0=t_0<t_1<\cdots\to\infty\) and issues \(\A_n=\B_{t_k}\) for \(t_k\le n<t_{k+1}\). Only the checkpoint times are deterministic. The actions issued there use the available observations.

In sums over observation counts, $n<T$ means $0\le n<T$. Logarithms are natural unless a base is displayed. Generic finite constants $C$ may change between bounds, with dependencies specified locally. Named constants remain fixed. The relations $\lesssim$, $\gtrsim$ and $\asymp$ suppress fixed positive factors.

\subsection{Excess and lifetime charge}\label{sec:performance}
For integer $T\ge1$, define cumulative expected excess and its limit by
\[
 \EE_T(A)=\sum_{n=0}^{T-1}\E[r_n(\A_n)-r_n(\B_n)],\qquad
 \EE_\infty(A)=\lim_{T\to\infty}\EE_T(A).
\]
The mean-square tracking error $e_n(A)=\E\norm{\A_n-\B_n}^2$ is deterministic. Predictive strong convexity will bound its cumulative sum by a multiple of $\EE_T(A)$.
Let $\CC$ consist of bounded nonnegative nondecreasing $c:[0,1]\to[0,\infty)$ with $c(0)=0$. For $c\in\CC$, set
\[
 V_{c,T}(A)=\sum_{n=1}^{T-1}c(\norm{\A_n-\A_{n-1}}),\qquad
 V_{c,\infty}(A)=\lim_{T\to\infty}V_{c,T}(A).
\]
These charges are random, unlike $\EE_T$ and $e_n(A)$. There is no charge before $\A_0$ or for internal posterior updates. The random number of actual revisions is $N_A(T)=\sum_{1\le n<T}\ind_{\{\A_n\ne\A_{n-1}\}}$.

Let $\HH$ consist of finite-valued nondecreasing $G:\N\to[1,\infty)$ with $G(J)=o(J)$. A pair $(c,G)$ is \emph{attainable} if one admissible policy and a finite constant $C_A$ satisfy
\begin{equation}\label{eq:attain}
 \EE_{2^J}(A)\le C_AG(J)\quad(J\ge1),\qquad \E V_{c,\infty}(A)<\infty.
\end{equation}
The policy may depend on the entire fixed pair. Requiring the excess bound only eventually is equivalent, because the finite initial prefix is integrable and $G\ge1$. At arbitrary horizons, monotonicity gives
\begin{equation}\label{eq:all-horizons}
 \EE_T(A)\le C_AG(\lceil\log_2T\rceil),\qquad T\ge2.
\end{equation}
The ceiling retains the correct prefix for irregular $G$, for which $G(J+1)=O(G(J))$ need not hold. For $J=\lceil\log_2T\rceil\ge2$,
\[
 0\le\frac{\EE_T(A)}{\log T}
 \le\frac{C_AG(J)}{(J-1)\log2}
 =\frac{C_A}{\log2}\frac{G(J)}J\frac{J}{J-1}
 \longrightarrow0.
\]

Two deterministic benchmark quantities accompany the policy-dependent error \(e_n(A)\): \(q_n=\E\norm{\bb_\theta-\bar{\bb}_n}^2\) and \(g_n=\E[R_\theta(\B_n)-R_\theta(\bb_\theta)]\), for \(n\ge0\).
Their roles are:
\begin{center}
\begin{tabularx}{\linewidth}{@{}lX@{}}
\toprule
Quantity & Meaning \\
\midrule
$e_n(A)$ & Squared tracking error of the issued action relative to $\B_n$. \\
$q_n$ & Posterior uncertainty about the latent optimum $\bb_\theta$, averaged over the data. \\
$g_n$ & Expected population loss of $\B_n$ above the latent optimum. \\
\bottomrule
\end{tabularx}
\end{center}
In particular, $e_n(B)=0$ even when $q_n,g_n>0$. The proof uses $q_n$ to force benchmark motion and $g_n$ to control the loss of retaining an earlier benchmark action.

\paragraph{The two risk comparisons.}
Define the latent-oracle cumulative risk as \(\mathcal R_T^{\mathrm{or}}(A):=\sum_{n<T}\E[R_\theta(\A_n)-R_\theta(\bb_\theta)]\). For any \(\F_n\)-measurable action \(\A_n\), independence of the private tape and posterior conditioning give \(\E[R_\theta(\A_n)\mid\F_n]=r_n(\A_n)\) and \(\E R_\theta(\A_n)=\E r_n(\A_n)\).
Appendix~\ref{app:kernels} justifies these identities at random actions. Adding and subtracting $R_\theta(\B_n)$ now yields
\[
\begin{aligned}
 \mathcal R_T^{\mathrm{or}}(A)
 &=\sum_{n<T}\E[R_\theta(\A_n)-R_\theta(\B_n)]
   +\sum_{n<T}\E[R_\theta(\B_n)-R_\theta(\bb_\theta)]\\
 &=\sum_{n<T}\E[r_n(\A_n)-r_n(\B_n)]+\sum_{n<T}g_n
 =\EE_T(A)+\sum_{n<T}g_n.
\end{aligned}
\]
Thus $\EE_T(A)$ is exactly the additional loss relative to the current predictive Bayes actions. Under the assumptions below, Appendix~\ref{app:foundations} proves $g_n\asymp n^{-1}$. If, more specifically, $\sum_{n<T}g_n=a_{\mathrm{or}}\log T+o(\log T)$ for some constant $a_{\mathrm{or}}>0$, then $\EE_T(A)=o(\log T)$ preserves this same leading coefficient in $\mathcal R_T^{\mathrm{or}}(A)$. The characterization also covers Bayes risks without a logarithmic leading coefficient. For the distinct hindsight comparator, Section~\ref{sec:calibration} gives an exact squared-loss calibration, while Appendix~\ref{app:empirical} gives a model with linear regret.

\subsection{Statistical assumptions}\label{sec:assumptions}
Assumption~\ref{ass:curvature} converts excess loss into squared tracking error. Assumptions~\ref{ass:witness} and~\ref{ass:information} convert posterior action uncertainty into movement of $\B_n$ across observation windows. Boundary conditions identify events on which the predictive gradient is stationary. A \emph{prior-full set} is a measurable subset of $\Theta$ with prior probability one.

\begin{assumption}[Integrability and population curvature]\label{ass:curvature}
The loss $(z,\w)\mapsto f_z(\w)$ is jointly measurable and obeys $|f_z(\w)|\le H_0(z)$ on $K$ for a nonnegative measurable envelope with $\E H_0(Z_1)<\infty$. On one measurable prior-full set $\Theta_0$, $R_\theta$ has a continuously differentiable extension to a relative neighborhood of $K$. On $K$ it is $\mu$-strongly convex and its gradient is $L$-Lipschitz, for fixed $\mu>0$ and $L<\infty$.
\end{assumption}
The assumptions permit unbounded or nonconvex sample losses. The envelope ensures integrability. Put \(\kappa=\mu/2\). Predictive risks inherit strong convexity, and quadratic growth at their constrained minimizers gives \(r_n(\w)-r_n(\B_n)\ge\kappa\norm{\w-\B_n}^2\) for \(\w\in K\), hence \(\kappa\sum_{n<T}e_n(A)\le\EE_T(A)\).
Appendix~\ref{app:kernels} constructs common conditional versions and proves quadratic growth without assuming interior minimizers.

For deterministic counts $0\le a\le b$, the action $\B_a$ is $\G_b$-measurable. The same conditioning identity gives
\begin{align*}
 \E r_b(\B_a)
 &=\E\bigl[\E[R_\theta(\B_a)\mid\G_b]\bigr]
 =\E R_\theta(\B_a),\\
 \E[r_b(\B_a)-r_b(\B_b)]
 &=\E[R_\theta(\B_a)-R_\theta(\bb_\theta)]
   -\E[R_\theta(\B_b)-R_\theta(\bb_\theta)]
 =g_a-g_b.
\end{align*}
Combining this identity with quadratic growth gives the \emph{risk-energy bound}
\begin{equation}\label{eq:risk-energy}
 g_a-g_b=\E[r_b(\B_a)-r_b(\B_b)]
 \ge\kappa\E\norm{\B_b-\B_a}^2\ge0.
\end{equation}
Thus $g_n$ is nonincreasing. Its decrease controls both the loss of retaining an earlier Bayes action and the squared displacement of the benchmark.

\Needspace{5\baselineskip}
\begin{assumption}[Observable gradients and boundary alternatives]\label{ass:witness}
A jointly measurable parameter-free field $\xiobs:\mathcal Z\times K\to\R^d$, written $\xiobs_z(\w)$, is unbiased for $\theta\in\Theta_0$:
\[
 \int\xiobs_z(\w)P_\theta(\dd z)=\nabla R_\theta(\w)\quad(\w\in K),\qquad
 \sup_{\theta,\w}\int\norm{\xiobs_z(\w)}^2P_\theta(\dd z)\le M_2<\infty.
\]
All suprema in this assumption range over $\theta\in\Theta_0$ and $\w\in K$. In addition,
\[
 \tau(R):=\sup_{\theta,\w}\int\norm{\xiobs_z(\w)}^2
 \ind_{\{\norm{\xiobs_z(\w)}>R\}}P_\theta(\dd z)\longrightarrow0\quad(R\to\infty).
\]
We call this uniform square-integrability condition UI2. For the boundary conditions, use the common predictive gradient \(\nabla r_n(\w):=\int\nabla R_u(\w)\Pi_n(\dd u)\) for \(\w\in K\).
The moment bound makes this field finite. Appendix~\ref{app:kernels} proves that it is the gradient on $\ri K$, continuously extended to $K$, on one probability-one event for all $n$. Require one of the following alternatives:
\begin{align*}
 \textup{(I)}\quad&\bb_\theta\in\ri K\quad\pi\text{-almost surely},\\
 \textup{(S)}\quad&\nabla R_\theta(\bb_\theta)=0\quad\pi\text{-almost surely},\\[-2pt]
 &\nabla r_n(\B_n)=0\quad\text{almost surely for every }n\ge0.
\end{align*}
\end{assumption}
The labels (I) and (S) mean interiority and stationarity. Both give $\nabla R_\theta(\bb_\theta)=0$. Alternative (S) gives $\nabla r_n(\B_n)=0$ at every count. Under (I), the proof instead restricts to observable events on which $\B_n$ stays a fixed positive distance from $\rbd K$. Lemma~\ref{lem:geometry} retains a fixed fraction of action uncertainty on these events without a uniform interior margin for $\bb_\theta$. Appendix~\ref{app:boundary} gives a model satisfying (I) in which (S) fails at arbitrarily large deterministic counts. Intersect $\Theta_0$ with the measurable prior-full set for the chosen population condition.

We call $\xiobs$ a gradient witness: the same field is observable without knowing $\theta$ and has the required population gradient as its conditional mean. It may differ from a sample-loss gradient. For squared Bernoulli loss, $\xi_y(w)=2(w-y)$ and $\E_p\xi_Y(w)=2(w-p)=\nabla R_p(w)$. UI2 excludes rare values carrying a fixed amount of squared witness mass. Uniform second-moment boundedness alone does not imply it. A bound $\sup_{\theta,\w}\E_\theta\norm{\xiobs_{Z_1}(\w)}^p\le M_p$ with $p>2$ is sufficient, since $\tau(R)\le M_pR^{2-p}$, but is not required.

\begin{assumption}[Action information]\label{ass:information}
For the expected posterior action variance $q_n=\E\norm{\bb_\theta-\bar{\bb}_n}^2$ defined in Section~\ref{sec:performance}, there are fixed $v>0$ and an integer $n_0\ge1$ such that $q_n\ge v/n$ for every $n\ge n_0$.
\end{assumption}
The quantity \(q_n\) measures uncertainty about the optimum rather than the parameter, since distinct states may share an action. Convexity gives \(\bar{\bb}_n\in K\), and tape independence gives the posterior orthogonality identity \(\E\norm{\A_n-\bb_\theta}^2=q_n+\E\norm{\A_n-\bar{\bb}_n}^2\ge q_n\).
\paragraph{Risk and action-moment bounds.}
Lemma~\ref{lem:accuracy} (Appendix~\ref{app:robust}) and Appendix~\ref{app:action-moments} derive the following bounds from Assumptions~\ref{ass:curvature}--\ref{ass:witness}. There are fixed finite constants $C_0,C_B$ such that, with $\Ctwo=C_0/\kappa$, for $n\ge1$,
\begin{equation}\label{eq:basic-action-bounds}
 \begin{aligned}
 0\le g_n&\le C_0/n,\\
 \E\norm{\B_n-\bb_\theta}^2&\le \Ctwo/n,\\
 \E\norm{\B_n-\bb_\theta}^4&\le C_B/n^2.
 \end{aligned}
\end{equation}
Appendix~\ref{app:robust} constructs the auxiliary estimator, and Appendix~\ref{app:action-moments} transfers its moment bounds to $\B_n$. The fourth moment controls boundary localization. UI2 controls the witness sums in Lemma~\ref{lem:ui}.
Assumption~\ref{ass:information} and posterior orthogonality also give
\[
 \frac{\kappa v}{n}\le\kappa q_n
 \le\kappa\E\norm{\B_n-\bb_\theta}^2
 \le g_n\le\frac{C_0}{n}\qquad(n\ge n_0).
\]
The bound on $g_n$ controls retention loss. Proposition~\ref{prop:motion} combines the gradient witness, boundary localization and UI2 to obtain a uniform positive-probability lower bound on benchmark motion.

\Needspace{5\baselineskip}
\subsection{Allocation and characterization}\label{sec:allocation-main}
Fix $c\in\CC$ and $G\in\HH$. The allocation assigns tracking-error allowances to action-distance scales. We use the following indices:
\begin{center}
\begin{tabularx}{\linewidth}{@{}llX@{}}
\toprule
Index & Range & Meaning \\
\midrule
$n$ & $n\ge0$ & Number of observations available before a prediction. \\
$J$ & $J\ge1$ & Horizon exponent, with $T=2^J$. \\
$j$ & $j\ge0$ & Observation epoch $2^j\le n<2^{j+1}$. \\
$i$ & $i\ge1$ & Distance scale $2^{-i}$, with charge $d_i=c(2^{-i})$. \\
$m$ & $m\ge1$ & Distance-prefix length in the allocation program. \\
\bottomrule
\end{tabularx}
\end{center}
All indices are integers. The allowance $a_i\ge0$ is indexed by distance, whereas $e_n(A)$ is indexed by observation count. A window $[n_\ell,m_\ell]$ has deterministic endpoints and internal edges $n_\ell+1,\ldots,m_\ell$. The endpoint $m_\ell$ is an observation count, distinct from the distance-prefix length $m$. We use $k$ for a retention-block count and $p$ for a tested distance prefix.

\paragraph{Why the allocation cost is reciprocal.}
Divide $[N,2N)$, $N=2^j$, into $k\le N$ nearly equal retention blocks with endpoints $t_0=N<\cdots<t_k=2N$. The calculation in Section~\ref{sec:proof-attainment} uses \eqref{eq:risk-energy} and \eqref{eq:basic-action-bounds} to show
\[
 \sum_{\ell=0}^{k-1}\sum_{n=t_\ell}^{t_{\ell+1}-1}
 \E[r_n(\B_{t_\ell})-r_n(\B_n)]
 \le C_0(k^{-1}+N^{-1}).
\]
A block-exit jump has action displacement \(\Delta_\ell=\norm{\B_{t_{\ell+1}}-\B_{t_\ell}}\). It is ordinary at distance scale \(2^{-i}\) when \(\Delta_\ell\le2^{-i}\). For \(a_i>0\) and \(k=\lceil a_i^{-1}\rceil\le N\), we have \(k^{-1}\le a_i\) and
\[
 \sum_{\ell=0}^{k-1}c(\Delta_\ell)\ind_{\{\Delta_\ell\le2^{-i}\}}
 \le kd_i\le(1+a_i^{-1})d_i=d_i+\frac{d_i}{a_i}.
\]
Thus $a_i$ controls excess through the block count. Smaller allowances require more blocks and increase the reciprocal charge $d_i/a_i$.

The target bounds cumulative excess at every horizon, so the allocation constrains every prefix. The converse distributes a required displacement over individual observation edges, whereas the construction charges direct jumps between checkpoint actions. Sections~\ref{sec:proof-converse}--\ref{sec:proof-attainment} relate the two index systems explicitly. After finitely many initial indices, the converse copies each epoch allowance to distance indices $2j+s$ and $2j+s+1$ for a fixed integer $s\ge0$. The construction assigns $a_i$ to epochs $4i,\ldots,4i+3$. In each direction, every source prefix is no longer than its target prefix. Monotonicity of $G$ therefore controls the budget up to a fixed factor.

\paragraph{The allocation program.}
Define $\phi_i(x)=d_i/x$ for $x>0$, with $\phi_i(0)=0$ if $d_i=0$ and $\phi_i(0)=+\infty$ otherwise. The finite and infinite allocation programs are
\begin{equation}\label{eq:allocation}
\begin{aligned}
 \Alloc_m(c,G)&=\min\left\{\sum_{i=1}^m\phi_i(a_i):a_i\ge0,\ 
                    \sum_{i=1}^{p}a_i\le G(p)\ (1\le p\le m)\right\},\\
 \Alloc(c,G)&=\inf\left\{\sum_{i\ge1}\phi_i(a_i):a_i\ge0,\ 
                    \sum_{i=1}^ma_i\le G(m)\ (m\ge1)\right\}.
\end{aligned}
\end{equation}
Each constraint bounds the total allowance through a distance prefix. Zero allowance has finite cost only at a zero charge weight.

\paragraph{Decreasing-weight dual.}
For decreasing nonnegative weights $(\omega_i)_{i=1}^m$, define
\begin{equation}\label{eq:dual}
 \Dual(c,G)=\sup_{\substack{m\ge1,\ \omega_1\ge\cdots\ge \omega_m\ge0\\\omega_1>0}}
 \frac{\left(\sum_{i=1}^m\sqrt{d_i\omega_i}\right)^2}
 {\sum_{i=1}^m(\omega_i-\omega_{i+1})G(i)},\qquad \omega_{m+1}=0.
\end{equation}
The denominator is at least $G(1)\omega_1>0$. Each difference $\omega_i-\omega_{i+1}$ weights a prefix constraint.

\Needspace{15\baselineskip}
\begin{theorem}[Prescribed-efficiency characterization]\label{thm:main}
Under the experiment and Assumptions~\ref{ass:curvature}--\ref{ass:information}, fix $(c,G)\in\CC\times\HH$. The following are equivalent:
\begin{enumerate}
\item[(i)] $(c,G)$ is attainable by an admissible policy, allowing independent private randomization.
\item[(ii)] $\Alloc(c,G)<\infty$.
\item[(iii)] $\Dual(c,G)<\infty$.
\item[(iv)] A horizon-independent policy retaining Bayes actions at deterministic checkpoints satisfies \eqref{eq:attain} and $\E[V_{c,\infty}(A)^q]<\infty$ for every $q>0$.
\end{enumerate}
Moreover, $\Alloc(c,G)=\sup_m\Alloc_m(c,G)=\Dual(c,G)$ in the extended nonnegative reals. When finite, the infinite infimum is attained.
\end{theorem}
One schedule works for all $q>0$. Its times may depend on the fixed model, prior, charge and profile. The values $\Alloc$ and $\Dual$ test feasibility, not the minimum expected charge. Theorem~\ref{thm:revisions}(b) gives a charge-independent schedule for each fixed power profile.

\paragraph{Example.}
For \(G(m)=\sqrt m\), take \(a_i=\sqrt i-\sqrt{i-1}\), so that \(\sum_{i=1}^m a_i=\sqrt m\) and \(\sum_{i\ge1}\phi_i(a_i)=\sum_{i\ge1}d_i(\sqrt i+\sqrt{i-1})\).
For the logarithmic charge $c_\alpha$ from Section~\ref{sec:intro}, this series converges exactly when $\alpha>3/2$. Corollary~\ref{cor:power} proves that this is also necessary, including for adaptive policies.

The programs in \eqref{eq:allocation} are also defined for positive finite-valued nondecreasing $G$ outside $\HH$. Rescaling $a_i$ by $t>0$ gives $\Alloc(c,tG)=t^{-1}\Alloc(c,G)$, which absorbs the coefficient $C_A$ in \eqref{eq:attain}.

\paragraph{Proof route.}
Proposition~\ref{prop:motion} converts action uncertainty into motion on selected windows. Section~\ref{sec:proof-converse} uses this motion to derive an allocation from any attaining policy, proving (i)$\Rightarrow$(ii). Section~\ref{sec:proof-attainment} constructs a retention schedule, proving (ii)$\Rightarrow$(iv). The implication (iv)$\Rightarrow$(i) follows by definition. Appendix~\ref{app:allocation} proves (ii)$\Leftrightarrow$(iii) by finite duality and compactness.

\paragraph{Why decreasing weights test all prefixes.}
For a finite-cost feasible allocation and $\omega_1\ge\cdots\ge \omega_m\ge0$, with $\omega_1>0$ and $\omega_{m+1}=0$, summation by parts and Cauchy--Schwarz give
\begin{equation}\label{eq:weak-duality}
\begin{aligned}
 \sum_{i=1}^m \omega_i a_i
 &=\sum_{p=1}^m(\omega_p-\omega_{p+1})\sum_{i=1}^p a_i
 \le\sum_{p=1}^m(\omega_p-\omega_{p+1})G(p),\\
 \left(\sum_{i=1}^m\sqrt{d_i\omega_i}\right)^2
 &\le\left(\sum_{i=1}^m\phi_i(a_i)\right)
       \left(\sum_{i=1}^m \omega_i a_i\right).
\end{aligned}
\end{equation}
At $a_i=0$, finite cost forces $d_i=0$, so these terms vanish. Divide by the positive weighted budget and take the infimum over feasible length-$m$ allocations. Each quotient is then at most $\Alloc_m(c,G)\le\Alloc(c,G)$. Taking the supremum over $m$ and the weights in \eqref{eq:dual} gives $\Dual(c,G)\le\Alloc(c,G)$. In the finite Lagrangian, $\omega_i=\sum_{p=i}^m\lambda_p$ for nonnegative prefix multipliers $\lambda_p$. Appendix~\ref{app:allocation} proves equality by finite duality and compactness.

\section{Feasibility thresholds and revision complexity}\label{sec:consequences}
All results in this section use the statistical assumptions of Theorem~\ref{thm:main}. Asymptotic statements concern sufficiently large horizons. Appendix~\ref{app:profiles} contains the proofs.

\subsection{Dini criteria for excess profiles}
Define the ordinary and square-root Dini functionals
\[
 \DiniOne(c)=\int_0^1\frac{c(x)}x\,\dd x,\qquad
 \DiniHalf(c)=\int_0^1\frac{\sqrt{c(x)}}x\,\dd x.
\]
The subscripts record the powers of $c$ in the integrands. Monotonicity of $c$ connects these near-zero integrals to the distance weights from Section~\ref{sec:allocation-main}:
\[
 \DiniOne(c)<\infty\ \Longleftrightarrow\ \sum_{i\ge1}d_i<\infty,
 \qquad
 \DiniHalf(c)<\infty\ \Longleftrightarrow\ \sum_{i\ge1}\sqrt{d_i}<\infty.
\]
Indeed, integrate over the intervals $[2^{-(i+1)},2^{-i}]$ and bound $c$ by its endpoint values. Appendix~\ref{app:endpoints} gives the comparison explicitly.
\begin{corollary}[Bounded and sublogarithmic excess]\label{cor:dini}
For fixed $c\in\CC$, a policy with $\EE_\infty(A)<\infty$ and $\E V_{c,\infty}(A)<\infty$ exists if and only if $\DiniHalf(c)<\infty$. A policy with $\EE_T(A)=o(\log T)$ and $\E V_{c,\infty}(A)<\infty$ exists if and only if $\DiniOne(c)<\infty$. Constructive policies can have every positive lifetime-cost moment and may depend on the fixed charge.
\end{corollary}
For \(G\equiv1\), the allocation has the exact value \(\Alloc(c,1)=(\sum_i\sqrt{d_i})^2\), giving the square-root threshold. The sublogarithmic statement quantifies over profiles: for a policy \(A\) with \(\EE_T(A)=o(\log T)\), set \(G_A(J)=\max\{1,\EE_{2^J}(A)\}\). Nonnegative excess makes \(G_A\) nondecreasing, and \(G_A(J)/J\le 1/J+\EE_{2^J}(A)/J\to0\), so \(G_A\in\HH\) and \(\EE_{2^J}(A)\le G_A(J)\).
Conversely, any profile in $\HH$ gives $\EE_T=o(\log T)$ by \eqref{eq:all-horizons}. Appendix~\ref{app:endpoints} proves the allocation identity and constructs a suitable profile from $\DiniOne(c)<\infty$.

\begin{corollary}[Weighted-Dini criterion]\label{cor:power}
For $c\in\CC$ and $0<r<1$, a policy with $\EE_T(A)=O((\log T)^r)$ and $\E V_{c,\infty}(A)<\infty$ exists exactly when
\begin{equation}\label{eq:weighted-dini}
 \sum_{i\ge1}i^{1-r}c(2^{-i})<\infty
 \quad\Longleftrightarrow\quad
 \int_0^1 c(x)[\log(e/x)]^{1-r}\frac{\dd x}{x}<\infty.
\end{equation}
For $c_\alpha(x)=[\log(e/x)]^{-\alpha}$ on $0<x\le1$, with $c_\alpha(0)=0$ and $\alpha>0$, this is $\alpha+r>2$.
\end{corollary}

\Needspace{14\baselineskip}
\subsection{Revision complexity}
Benchmark motion also forces actual revisions. The lower bound below depends on tracking error alone.
\begin{theorem}[Revision complexity]\label{thm:revisions}
\leavevmode
\begin{enumerate}
\item[(a)] If an admissible policy satisfies $\EE_{2^J}(A)\le C_AG(J)$ eventually for $G\in\HH$, then
$\E N_A(2^J)\ge K_AJ^2/G(J)$ for all sufficiently large $J$, where $K_A>0$ depends only on model, prior and $C_A$. The threshold for sufficiently large $J$ may depend on the policy and profile. No finite-cost assumption is needed.

\item[(b)] For \(0<r<1\), every policy with \(\EE_T(A)=O((\log T)^r)\) has \(\E N_A(T)=\Omega((\log T)^{2-r})\). For each such \(r\), one horizon-independent deterministic-checkpoint policy satisfies
\[
 \EE_T(A)=O((\log T)^r),\qquad
 N_A(T)=O((\log T)^{2-r})\quad\text{pathwise}.
\]
For every fixed charge satisfying \eqref{eq:weighted-dini}, this same policy has $\E[V_{c,\infty}(A)^q]<\infty$ for all $q>0$. Its checkpoint times depend on $r$, not on $c$. The moment bounds may depend on $c$.

\item[(c)] For fixed $c\in\CC$ and $\beta\in\R$, a policy with $\EE_\infty(A)<\infty$, $\E V_{c,\infty}(A)<\infty$, and $\E N_A(T)=O((\log T)^2(\log\log T)^\beta)$ exists if and only if $\DiniHalf(c)<\infty$ and $\beta>1$. One deterministic-checkpoint policy then has the revision bound pathwise and all positive lifetime-cost moments.
\end{enumerate}
\end{theorem}
Appendix~\ref{app:revisions} proves the lower bound by counting revisions on a fixed fraction of low-error epochs and applying Cauchy--Schwarz.
For the upper bound in part (b), use $k_j=\lceil(j+1)^{1-r}\rceil$ retention blocks in observation epoch $j\ge4$. At each checkpoint the policy issues the current predictive Bayes action. The checkpoint count bounds $N_A(T)$, since a scheduled update may leave the action unchanged. Part (c) uses a cap with summable reciprocals.

For example, $c_{3/2}$ and $r=3/5$ permit excess $O((\log T)^{3/5})$ with $O((\log T)^{7/5})$ actual revisions pathwise. Every competing policy with that excess order requires the same revision order in expectation. The target $r=1/2$ is infeasible for this charge.

\subsection{Critical profiles}
We write $a_T=\omega(b_T)$ when $a_T/b_T\to\infty$ for an eventually positive sequence $b_T$.
\begin{corollary}[Critical logarithmic profiles]\label{cor:critical}
For $1<\alpha<2$, every policy with $\E V_{c_\alpha,\infty}(A)<\infty$ satisfies $\EE_T(A)=\omega((\log T)^{2-\alpha})$. For every real $\beta$,
\begin{equation}\label{eq:critical-one}
 \exists A:\ \EE_T(A)=O((\log T)^{2-\alpha}(\log\log T)^\beta),\quad
 \E V_{c_\alpha,\infty}(A)<\infty
 \quad\Longleftrightarrow\quad\beta>1.
\end{equation}
At $\alpha=2$, every finite-expected-cost policy satisfies $\EE_T(A)=\omega((\log\log T)^2)$, and for $\gamma\ge0$,
\begin{equation}\label{eq:critical-two}
 \exists A:\ \EE_T(A)=O((\log\log T)^\gamma),\quad
 \E V_{c_2,\infty}(A)<\infty
 \quad\Longleftrightarrow\quad\gamma>2.
\end{equation}
Constructive policies can have all positive lifetime-cost moments.
\end{corollary}
The vanishing tail of the allocation cost strengthens a lower bound at the critical order to the strict $\omega$ statement. Appendix~\ref{app:critical} constructs the epoch allowances, proves the tail inequality and evaluates the logarithmic sums.

Table~\ref{tab:thresholds} collects the preceding charge thresholds.
\begin{table}[H]
\centering
\caption{Feasibility for logarithmic charges $c_\alpha$. Equality is infeasible at every strict threshold.}\label{tab:thresholds}
\begin{tabular}{lll}\toprule
Required cumulative excess & Criterion & Result\\\midrule
Some $o(\log T)$ profile & $\alpha>1$ & Corollary~\ref{cor:dini}\\
$O(1)$ & $\alpha>2$ & Corollary~\ref{cor:dini}\\
$O((\log T)^r)$, $0<r<1$ & $\alpha+r>2$ & Corollary~\ref{cor:power}\\
$O((\log T)^{2-\alpha}(\log\log T)^\beta)$, $1<\alpha<2$ & $\beta>1$ & Corollary~\ref{cor:critical}\\
$O((\log\log T)^\gamma)$, $\alpha=2$, $\gamma\ge0$ & $\gamma>2$ & Corollary~\ref{cor:critical}\\\bottomrule
\end{tabular}
\end{table}

\section{Verification of assumptions and model examples}\label{sec:applications}
A local score criterion verifies action information in two logistic decision models. We then calibrate the excess under squared loss.

\subsection{A local criterion for action information}
A local change in the optimal action can force order-$1/n$ posterior action variance. The following score criterion requires regularity only on an open parameter set. Write $\bb(\theta)=\bb_\theta$ for the action map and $\partial_\vvec$ for differentiation in a unit parameter direction $\vvec$. The operator $\Jac$ denotes the Jacobian, so $\Jac\bb(\theta)\vvec$ is the directional derivative.
\Needspace{5\baselineskip}
\begin{proposition}[Local-score action information]\label{prop:local-score}
Let $d_{\mathrm{par}}$ be the parameter dimension. Let the parameter space be a subset of $\R^{d_{\mathrm{par}}}$ containing a nonempty open set $U$, and let $\bb_\theta\in\R^d$ be bounded and measurable. Suppose:
\begin{enumerate}
\item The restriction of the probability prior to $U$ has a positive $C^1$ Lebesgue density: $\pi|_U(\dd \theta)=\varpi(\theta)\,\dd \theta$. This measure has mass $\pi(U)$ and is not renormalized.
\item On $U$, the conditionally iid law $P_\theta$ has a positive jointly measurable density $p_\theta$ on a common support, relative to a sigma-finite measure $\nu$. On a common $\nu$-full set, $\theta\mapsto p_\theta(z)$ is $C^1$ on $U$.
\item For a unit direction $\vvec$, the score $s_\vvec=\partial_\vvec\log p_\theta$ satisfies $\E_\theta s_\vvec(\theta,Z)=0$, and $I_\vvec(\theta)=\E_\theta s_\vvec(\theta,Z)^2$ is locally bounded on $U$. The map $\bb$ is $C^1$ on $U$, with $\Jac\bb(\theta_0)\vvec\ne0$ at some $\theta_0\in U$.
\end{enumerate}
Then there is $v_0>0$ such that
\begin{equation}\label{eq:local-info}
 q_n=\E\norm{\bb_\theta-\E[\bb_\theta\mid\G_n]}^2\ge v_0/n,\qquad n\ge1.
\end{equation}
The differentiability and score conditions concern only $U$. By conditional orthogonality, the same lower bound applies to the mean-square error of any estimator based on the observations and an independent private tape.
\end{proposition}
The score and nonzero action derivative use the same direction. When the score assumptions hold in every unit direction, $\Jac\bb(\theta_0)\ne0$ suffices.

Appendix~\ref{app:local-score} proves the criterion by projecting the action onto a direction of nonzero derivative and integrating a compactly supported score under the original prior.

\subsection{Logistic decisions with finite second covariate moments}\label{sec:logistic}
For symmetric matrices, $\preceq$ denotes Loewner order and $\Id_d$ is the identity, with the dimension omitted when clear. The two models differ in their latent optima: without regularization the optimum equals the parameter, while fixed ridge keeps the optimum bounded even for an unbounded parameter prior.
\paragraph{Coordinates.}
Throughout this subsection, $\w$ is a physical action and $\wnorm$ its normalized coordinate. Write $\ell_z$ for the physical loss. Let $K_0$ be compact with diameter $D_{\phys}>0$, and let $\widetilde c$ be a bounded nonnegative nondecreasing charge on $[0,D_{\phys}]$ with $\widetilde c(0)=0$. Define $R_\theta^\phys(\w)=\E_\theta\ell_Z(\w)$ and $r_n^\phys(\w)=\int R_\vartheta^\phys(\w)\Pi_n(\dd\vartheta)$, with minimizers $\bb_\theta^\phys$ and $\B_n^\phys$ over $K_0$. The coordinate dictionary is
\begin{center}
\begin{tabularx}{\linewidth}{@{}l l X@{}}
\toprule
Quantity & Physical coordinates & Normalized coordinates \\
\midrule
Action set & $K_0$ & $K=K_0/D_{\phys}$ \\
Action & $\w\in K_0$ & $\wnorm=\w/D_{\phys}\in K$ \\
Optima & $\bb_\theta^\phys,\ \B_n^\phys$ & $\bb_\theta=\bb_\theta^\phys/D_{\phys},\ \B_n=\B_n^\phys/D_{\phys}$ \\
Sample loss & $\ell_z(\w)$ & $f_z(\wnorm)=\ell_z(D_{\phys}\wnorm)$ \\
Charge & $\widetilde c$ on $[0,D_{\phys}]$ & $c(x)=\widetilde c(D_{\phys}x)$, $0\le x\le1$ \\
\bottomrule
\end{tabularx}
\end{center}
The action coordinates change while the latent parameter, observations and sampling law remain fixed. Superscript $\phys$ also marks the physical gradient witness. If $\widetilde A=(D_{\phys}\A_n)_{n\ge0}$ and $\widetilde\EE_T$ denotes physical excess, substitution gives
\begin{equation}\label{eq:physical-dictionary}
 \B_n^\phys=D_{\phys}\B_n,\qquad
 V_{\widetilde c,\infty}(\widetilde A)=V_{c,\infty}(A),\qquad
 \widetilde\EE_T(\widetilde A)=\EE_T(A).
\end{equation}
The risk and gradient transformations are
\[
\begin{aligned}
 R_\theta(\wnorm)&=R_\theta^\phys(D_{\phys}\wnorm),
 &r_n(\wnorm)&=r_n^\phys(D_{\phys}\wnorm),\\
 \nabla_{\wnorm}R_\theta(\wnorm)
 &=D_{\phys}\nabla_\w R_\theta^\phys(D_{\phys}\wnorm),
 &\xiobs_z(\wnorm)&=D_{\phys}\xiobs_z^\phys(D_{\phys}\wnorm).
\end{aligned}
\]
Normalization multiplies both curvature constants by $D_{\phys}^2$ and preserves loss values. Below, $\bb_\theta,\B_n,\A_n$ denote normalized actions, whereas their physical counterparts carry $\phys$ or a tilde. The derivatives $\nabla_\w$ and $\nabla_{\wnorm}$ specify which action coordinates are used.

\paragraph{Unregularized decisions.}
Let $\sigma(t)=(1+e^{-t})^{-1}$. Take a compact, convex, full-dimensional physical action set $K_0\subset\R^d$ and a prior with $\pi(\interior K_0)=1$. Assume that on some open ball $U\subset\interior K_0$, the restricted prior $\pi|_U$ has a positive $C^1$ Lebesgue density. The support may approach $\partial K_0$. The conditionally iid observations are pairs $Z=(\X,Y)$ with
\begin{equation}\label{eq:logistic-law}
 \X\sim Q,\qquad Y\mid\X,\theta\sim\Bernoulli(\sigma(\X^\top\theta)),
 \qquad \E_Q\norm\X^2<\infty,\quad \E_Q[\X\X^\top]\succ0.
\end{equation}
Here \(Q\) is fixed and parameter-independent, and the physical loss is \(\ell_{\x,y}(\w)=\log(1+e^{\x^\top\w})-y\x^\top\w\). With \(R_0=\sup_{\w\in K_0}\norm\w\), the population Hessian is \(\mathbf H(\w)=\E_Q[\sigma'(\X^\top\w)\X\X^\top]\), and the observable witness is \(\xiobs^{\phys}_{(\x,y)}(\w)=\x\{\sigma(\x^\top\w)-y\}\).
They satisfy
\begin{equation}\label{eq:logistic-curvature}
 \mu_0\Id\preceq\mathbf H(\w)\preceq L_0\Id\quad(\w\in K_0),\qquad 0<\mu_0\le L_0<\infty.
\end{equation}
For each unit \(\vvec\), \(\sigma'>0\) and \(\E_Q[(\vvec^\top\X)^2]>0\). Continuity of the Hessian on the compact action set makes the bounds uniform. The individual sample Hessian can still have rank one. The envelope and witness satisfy \(0\le\ell_{\X,Y}(\w)\le\log2+R_0\norm\X\) and \(\norm{\xiobs^{\phys}_{(\X,Y)}(\w)}\le\norm\X\), while
\[
 \sup_{\theta,\w}\E_\theta\!\big[\norm{\xiobs^{\phys}_Z(\w)}^2
       \ind_{\{\norm{\xiobs^{\phys}_Z(\w)}>t\}}\big]
 \le\E_Q\!\big[\norm\X^2\ind_{\{\norm\X>t\}}\big]\longrightarrow0.
\]
Thus UI2 holds without a higher moment. On the prior-full interior set, conditioning on \(\X\) gives \(\nabla R_\theta^\phys(\w)=\E_Q[\X\{\sigma(\X^\top\w)-\sigma(\X^\top\theta)\}]\). Since \(\nabla R_\theta^\phys(\theta)=0\), strong convexity implies \(\bb_\theta^\phys=\theta\). For a unit parameter direction \(\vvec\), the score
\(s_\vvec(\theta,Z)=(\vvec^\top\X)\{Y-\sigma(\X^\top\theta)\}\)
satisfies \(\E_\theta[s_\vvec\mid\X]=0\) and
\[
 \E_\theta s_\vvec^2
 =\E_Q[(\vvec^\top\X)^2\sigma'(\X^\top\theta)]
 \le\E_Q\norm\X^2.
\]
The normalized action derivative is \(\Jac\bb(\theta)=\Id/D_{\phys}\).
Proposition~\ref{prop:local-score} gives action information. Population interiority supplies alternative (I), even when predictive optima lie on the boundary. Appendix~\ref{app:logistic} supplies the differentiation and measurability details.
With $D_{\phys}=\diam K_0$, the normalized constants are $\mu=D_{\phys}^2\mu_0$, $L=D_{\phys}^2L_0$ and $M_2=D_{\phys}^2\E_Q\norm\X^2$.

\paragraph{Fixed-ridge decisions.}
Keep the observation law \eqref{eq:logistic-law}, now with a prior on $\R^d$ whose restriction to some open ball has a positive $C^1$ Lebesgue density. The prior may have unbounded support. For fixed $\lambda>0$, use
$\ell_{\x,y}(\w)=\log(1+e^{\x^\top\w})-y\x^\top\w+\lambda\norm\w^2/2$.
The ridge term belongs to the decision loss. Define $R_\theta^\phys$ and $r_n^\phys$ on $\R^d$ using this loss, and let their global minimizers be
\[
 \bb_\theta^\phys=\argmin_{\w\in\R^d}R_\theta^\phys(\w),\qquad
 \B_n^\phys=\argmin_{\w\in\R^d}r_n^\phys(\w).
\]
Generally $\bb_\theta^\phys\ne\theta$. The stationary equations in Appendix~\ref{app:ridge} bound both minimizer norms by $\E_Q\norm\X/\lambda$, so the centered closed ball of radius $R>\E_Q\norm\X/\lambda$ contains them in its interior. The expectation is under the fixed covariate law $Q$ and imposes no moment assumption on the parameter prior. On that ball the physical witness
$\xiobs^{\phys}_{(\x,y)}(\w)=\x\{\sigma(\x^\top\w)-y\}+\lambda\w$ is uniformly square-integrable, dominated in squared norm by $(\norm\X+\lambda R)^2$. Population Hessians are bounded above and below by positive constants, with lower bound $\lambda\Id$. Both boundary alternatives hold.

Write \(\bb^\phys(\theta)=\bb_\theta^\phys\), and define \(\mathbf H_\theta=\nabla_\w^2R_\theta^\phys(\bb_\theta^\phys)\succeq\lambda\Id\) and \(\mathbf I_\theta=\E_Q[\sigma'(\X^\top\theta)\X\X^\top]\succ0\). Differentiating the stationary equation gives
\[
\begin{aligned}
 0&=\Jac_\theta\{\nabla_\w R_\theta^\phys(\bb_\theta^\phys)\}
   =\mathbf H_\theta\Jac\bb^\phys(\theta)-\mathbf I_\theta,\\
 \Jac\bb^\phys(\theta)&=\mathbf H_\theta^{-1}\mathbf I_\theta.
\end{aligned}
\]
Use the coordinate dictionary with \(K_0=\{\w:\norm\w\le R\}\) and \(D_{\phys}=2R\). The normalized action map is \(\bb(\theta)=\bb^\phys(\theta)/(2R)\), so \(\Jac\bb(\theta)=(2R)^{-1}\mathbf H_\theta^{-1}\mathbf I_\theta\ne0\). Both matrices are invertible, so Proposition~\ref{prop:local-score} gives action information, including under nondegenerate Gaussian priors. The normalized set is \(K=\{\wnorm:\norm{\wnorm}\le1/2\}\), and \eqref{eq:physical-dictionary} preserves excess and charges. Appendix~\ref{app:ridge} gives the full verification.

\paragraph{Second moments without higher moments.}
Appendix~\ref{app:heavy-tail} gives an explicit covariate law with $\E_Q\norm\X^2<\infty$, $\E_Q[\X\X^\top]\succ0$, and $\E_Q\norm\X^{2+\delta}=\infty$ for every $\delta>0$. The natural logistic witness also has no moment above two, with or without fixed ridge.

\subsection{Bernoulli--Brier calibration}\label{sec:calibration}
Return to the scalar Bernoulli experiment of Section~\ref{sec:intro}, with Brier loss $f_y(w)=(w-y)^2$. Write $m_n=\E[p\mid\G_n]$. Then \(b_p=p\), \(B_n=m_n=(S_n+1)/(n+2)\), and \(g_n=q_n=1/[6(n+2)]\asymp n^{-1}\) for \(n\ge1\). Both optima are stationary and the sample gradient is a bounded witness, so all assumptions hold. For the linear charge $c_{\mathrm{lin}}(x)=x$, always issuing the current Bayes action incurs
\[
 \E V_{c_{\mathrm{lin}},T}(B)
 =\frac13\sum_{k=2}^{T}\frac1k\sim\frac{\log T}{3}.
\]
Since $\DiniHalf(c_{\mathrm{lin}})=\int_0^1x^{-1/2}\,\dd x=2$, Corollary~\ref{cor:dini} gives one deterministic retention policy with bounded cumulative expected excess and lifetime-cost moments of every positive order. Thus finite lifetime movement is compatible with a finite total expected loss increase relative to the continuously updated Bayes policy. Appendix~\ref{app:calibration} derives these formulas and the exact relation to expected hindsight regret.

\section{Proof of the characterization}\label{sec:proof}
We prove the probabilistic implications in Theorem~\ref{thm:main}. Appendix~\ref{app:allocation} gives the allocation duality.

\subsection{Information forces benchmark motion}\label{sec:proof-motion}
Assumption~\ref{ass:information} gives $q_n\ge v/n$ eventually, while \eqref{eq:basic-action-bounds} bounds the second and fourth moments of $\B_n-\bb_\theta$. Lemma~\ref{lem:geometry} (Appendix~\ref{app:localization}) combines these bounds with the boundary alternatives. Let $N_{\mathrm{loc}}$ be the threshold in that lemma and, under (I), let $\rho>0$ be its boundary margin. For $n\ge N_{\mathrm{loc}}$, define
\[
 \loc_n=\begin{cases}
 \Omega,&\text{under (S)},\\
 \{\dist(\B_n,\rbd K)\ge\rho\},&\text{under (I)}.
 \end{cases}
\]
Lemma~\ref{lem:geometry} gives a finite constant $K_{\mathrm{geo}}>0$ such that
\[
\begin{aligned}
 \E[\ind_{\loc_n}\norm{\B_n-\bb_\theta}^2]&\ge v/(2n),\\
 \ind_{\loc_n}\nabla r_n(\B_n)&=0,\\
 \ind_{\loc_n}\norm{\nabla r_m(\B_n)}&\le K_{\mathrm{geo}}\norm{\B_m-\B_n}\qquad(m\ge n).
\end{aligned}
\]
The first inequality retains posterior uncertainty on stationary events, and the third converts a future gradient into action displacement. Strong monotonicity also gives
\[
 \begin{aligned}
 \mu\norm{\B_n-\bb_\theta}^2
 &\le\ip{\nabla R_\theta(\B_n)-\nabla R_\theta(\bb_\theta)}{\B_n-\bb_\theta}\\
 &=\ip{\nabla R_\theta(\B_n)}{\B_n-\bb_\theta}
 \le\norm{\nabla R_\theta(\B_n)}\norm{\B_n-\bb_\theta}.
 \end{aligned}
\]
Thus $\E\norm{\ind_{\loc_n}\nabla R_\theta(\B_n)}^2\ge\mu^2v/(2n)$, for every $n\ge N_{\mathrm{loc}}$. The risk-energy bound controls the sum of future predictive gradient second moments over disjoint windows. Selecting windows with small second moments lets us turn the gradient signal into motion with positive probability.

\begin{proposition}[Uniform selected-window motion]\label{prop:motion}
Under Assumptions~\ref{ass:curvature}--\ref{ass:information}, there are $a_w,\eta\in(0,1]$ and one deterministic $N_*<\infty$ such that, for every integer $N\ge N_*$, $1\le M\le\lfloor N/6\rfloor$, and deterministic integer windows with disjoint internal edge sets (adjacent windows may share an endpoint)
$N\le n_1<m_1\le n_2<\cdots\le n_M<m_M<2N$ satisfying $h_\ell=m_\ell-n_\ell\ge N/(6M)$, at least $M/2$ intervals form a deterministic subset on which
\[
 \Pp\{\norm{\B_{m_\ell}-\B_{n_\ell}}\ge a_w/\sqrt{MN}\}\ge\eta.
\]
The constants and threshold are independent of the endpoints, charge, profile and competing policy.
\end{proposition}

\begin{proof}
\emph{Covariance signal.} Fix $N\ge N_*:=\lceil\max\{6,N_{\mathrm{loc}}\}\rceil$. For a window $N\le n<m<2N$, put $h=m-n$ and freeze the witness at $\B_n$:
\[
 \U=\ind_{\loc_n}\nabla R_\theta(\B_n),\quad
 \Uhat=\E[\U\mid\G_m],\quad
 \Svec=\ind_{\loc_n}\sum_{t=n+1}^{m}\xiobs_{Z_t}(\B_n).
\]
The vector $\U$ is $\sigma(\theta)\vee\G_n$-measurable, whereas $\Uhat$ and $\Svec$ are $\G_m$-measurable. Lemma~\ref{lem:gradients} (Appendix~\ref{app:kernels}), stationarity on $\loc_n$ and the tower property give
\[
\begin{aligned}
 \E[\U\mid\G_n]&=\ind_{\loc_n}\nabla r_n(\B_n)=0,\\
 \Uhat=\E[\U\mid\G_m]&=\ind_{\loc_n}\nabla r_m(\B_n),\\
 \E[\Svec\mid\theta,\G_n]&=h\U,\\
 \E\ip{\Uhat}{\Svec}
 &=\E\ip{\E[\U\mid\G_m]}{\Svec}
 =\E\ip{\U}{\Svec}\\
 &=\E\ip{\U}{\E[\Svec\mid\theta,\G_n]}
 =h\E\norm{\U}^2\\
 &\ge \mu^2h\E[\ind_{\loc_n}\norm{\B_n-\bb_\theta}^2]
 \ge \frac{\mu^2v}{4}\frac hN.
\end{aligned}
\]
\emph{Window selection.} Write $\Uhat_\ell$ for $\Uhat$ on window $[n_\ell,m_\ell]$. For disjoint internal edges, the aggregate form of \eqref{eq:risk-energy} in Appendix~\ref{app:action-moments} and the localized gradient bound yield
\[
\begin{aligned}
 \sum_{\ell=1}^{M}\E\norm{\Uhat_\ell}^2
 &\le K_{\mathrm{geo}}^2\sum_{\ell=1}^{M}\E\norm{\B_{m_\ell}-\B_{n_\ell}}^2\\
 &\le \frac{K_{\mathrm{geo}}^2}{\kappa}\sum_{\ell=1}^{M}(g_{n_\ell}-g_{m_\ell})
 \le\frac{K_{\mathrm{geo}}^2C_0}{\kappa N}.
\end{aligned}
\]
Put \(C_{\mathrm{en}}=1+K_{\mathrm{geo}}^2C_0/\kappa\) and \(\zeta_\ell=N^2\E\norm{\Uhat_\ell}^2/h_\ell\). Since \(h_\ell\ge N/(6M)\), we have
\[
 \sum_{\ell=1}^M\zeta_\ell
 \le6MN\sum_{\ell=1}^M\E\norm{\Uhat_\ell}^2
 \le6C_{\mathrm{en}}M.
\] Define \(\mathcal I=\{\ell:\zeta_\ell\le12C_{\mathrm{en}}\}\). Then \(12C_{\mathrm{en}}|\mathcal I^c|\le6C_{\mathrm{en}}M\).
Consequently $|\mathcal I|\ge M/2$ and $\E\norm{\Uhat_\ell}^2\le12C_{\mathrm{en}}h_\ell/N^2$ on $\mathcal I$. The set is deterministic because each $\zeta_\ell$ is an expected moment.

\emph{Positive-probability motion.} To control the witness sum, write
\[
 \W_t=\xiobs_{Z_t}(\B_n)-\nabla R_\theta(\B_n),\qquad
 \Svec=\ind_{\loc_n}\sum_{t=n+1}^m\W_t+h\U.
\]
Conditionally on \((\theta,\G_n)\), the \(\W_t\) are centered and independent. Lemma~\ref{lem:ui} (Appendix~\ref{app:innovation}) controls their normalized sum by truncation and the mean term by \eqref{eq:basic-action-bounds}. Since \(h\le N\le n\), it gives a constant \(C_S\) and a deterministic nondecreasing modulus \(\ui\), with \(\ui(t)\to0\) as \(t\downarrow0\), such that, for every event \(F\),
\[
 \E\norm{\Svec}^2\le C_S h,\qquad
 \E[\norm{\Svec}^2\ind_F]\le h\,\ui(\Pp(F)).
\]
This modulus prevents the witness energy from concentrating on a vanishing-probability event. For a selected window, put $K_{\mathrm{inf}}=\mu^2v/4$ and $C_{\mathrm{sel}}=\max\{1,C_S,12C_{\mathrm{en}}\}$. For $a>0$, let $F=\{\norm{\Uhat}\ge a\sqrt{h}/N\}$. Splitting the covariance across $F^c$ and $F$ gives
\[
\begin{aligned}
 K_{\mathrm{inf}}\frac hN
 &\le\E\ip{\Uhat}{\Svec}\\
 &=\E[\ip{\Uhat}{\Svec}\ind_{F^c}]
   +\E[\ip{\Uhat}{\Svec}\ind_F]\\
 &\le\frac{a\sqrt h}{N}(\E\norm{\Svec}^2)^{1/2}
   +(\E\norm{\Uhat}^2)^{1/2}(\E[\norm{\Svec}^2\ind_F])^{1/2}\\
 &\le\frac hN\{a\sqrt{C_{\mathrm{sel}}}
       +\sqrt{C_{\mathrm{sel}}\,\ui(\Pp(F))}\}.
\end{aligned}
\]
Choose $a=\min\{1,K_{\mathrm{inf}}/(4\sqrt{C_{\mathrm{sel}}})\}$ and then $\eta\in(0,1)$ with $\ui(\eta)\le K_{\mathrm{inf}}^2/(16C_{\mathrm{sel}})$. Such an $\eta$ exists because $\ui(t)\to0$. If $\Pp(F)\le\eta$, the right side is at most $K_{\mathrm{inf}}h/(2N)$, a contradiction. Hence $\Pp(F)>\eta$. Since $\Uhat=0$ outside $\loc_n$, on $F$ we have
\[
 \norm{\B_m-\B_n}\ge\frac{\norm{\Uhat}}{K_{\mathrm{geo}}}
 \ge\frac{a\sqrt h}{K_{\mathrm{geo}}N}
 \ge\frac{a}{K_{\mathrm{geo}}\sqrt{6MN}}.
\]
Take $a_w=\min\{1,a/(K_{\mathrm{geo}}\sqrt6)\}$. The constants depend only on the model and prior, and $\mathcal I$ depends only on deterministic endpoints and expected moments.
\end{proof}

\subsection{Necessity: from an attaining policy to a feasible allocation}\label{sec:proof-converse}
Fix an attaining policy, including its independent private tape, and put $e_n=e_n(A)$. Its deterministic epoch error and random internal charge are
\[
 Q_j=\sum_{n=2^j}^{2^{j+1}-1}e_n(A),\qquad
 W_{c,j}=\sum_{n=2^j+1}^{2^{j+1}-1}c(\norm{\A_n-\A_{n-1}}).
\]
Window endpoints will minimize expected errors and hence remain deterministic for randomized policies. Quadratic growth and disjoint internal edges give
\[
 \kappa\sum_{j=0}^{J}Q_j\le\EE_{2^{J+1}}(A)\le C_AG(J+1),
 \qquad \sum_{j\ge0}\E W_{c,j}\le\E V_{c,\infty}(A)<\infty.
\]
\paragraph{Window charges.}
For $N=2^j\ge N_*$ and $1\le M\le\lfloor N/6\rfloor$, take $M$ disjoint three-block groups, each of length $3b$ with $b=\lfloor N/(3M)\rfloor$. Within each group, minimize the deterministic $e_n$ over the first and third blocks to obtain endpoints $(n_\ell,m_\ell)$. They satisfy $m_\ell-n_\ell\ge N/(6M)$ and
\[
 \sum_{\ell=1}^M(e_{n_\ell}+e_{m_\ell})\le\frac{6M}{N}Q_j.
\]
Put $\delta_w=a_w/\sqrt{MN}$ and
$F_\ell=\{\max_{v\in\{n_\ell,m_\ell\}}\norm{\A_v-\B_v}>\delta_w/4\}$. Markov's inequality gives
\[
 \sum_{\ell=1}^M\Pp(F_\ell)
 \le\frac{16}{\delta_w^2}\sum_{\ell=1}^M(e_{n_\ell}+e_{m_\ell})
 \le\frac{96M^2}{a_w^2}Q_j.
\]
On the benchmark-motion event outside $F_\ell$, the triangle inequality and the path-length bound give
\[
\begin{aligned}
 \norm{\A_{m_\ell}-\A_{n_\ell}}
 &\ge \norm{\B_{m_\ell}-\B_{n_\ell}}
       -\norm{\A_{m_\ell}-\B_{m_\ell}}
       -\norm{\A_{n_\ell}-\B_{n_\ell}}
 \ge\delta_w/2,\\
 \sum_{t=n_\ell+1}^{m_\ell}c(\norm{\A_t-\A_{t-1}})
 &\ge c\left(\max_{n_\ell<t\le m_\ell}\norm{\A_t-\A_{t-1}}\right)\\
 &\ge c\!\left(\frac{\norm{\A_{m_\ell}-\A_{n_\ell}}}{m_\ell-n_\ell}\right)
 \ge c(\delta_w/(2N)).
\end{aligned}
\]
Proposition~\ref{prop:motion} selects a deterministic set $\mathcal I$ with $|\mathcal I|\ge\chi M$, where $\chi=1/2$. Disjoint internal edges and $C_w=96/a_w^2$ give
\begin{equation}\label{eq:window-sum-main}
\begin{aligned}
 \E W_{c,j}
 &\ge c(\delta_w/(2N))\sum_{\ell\in\mathcal I}
       \Pp\{\norm{\A_{m_\ell}-\A_{n_\ell}}\ge\delta_w/2\}\\
 &\ge c(\delta_w/(2N))
       \left\{\eta|\mathcal I|-\sum_{\ell\in\mathcal I}\Pp(F_\ell)\right\}\\
 &\ge c(\delta_w/(2N))
       \left\{\chi\eta M-\frac{16}{\delta_w^2}\sum_{\ell=1}^M(e_{n_\ell}+e_{m_\ell})\right\}\\
 &\ge (\chi\eta M-C_wM^2Q_j)
       c\!\left(\frac{a_w}{2\sqrt M}2^{-3j/2}\right).
\end{aligned}
\end{equation}
The extra factor $N^{-1}$ inside the charge comes from distributing a net displacement over at most $N$ observation edges. Appendix~\ref{app:converse} verifies the integer endpoints and records this bound as Lemma~\ref{lem:window-charge}.
\paragraph{Epoch allowances.}
Choose \(\alpha_*,\lambda_0>0\) with
\(C_w\alpha_*+\lambda_0<\chi\eta\), and fix an integer \(j_0\)
satisfying \(2^{j_0}\ge\max\{N_*,12\}\).
Put \(\delta_*=a_w/2\), and define
\(\good=\{j\ge j_0:Q_j\le\alpha_*/2\}\) and
\(M_j=\lfloor\min\{\alpha_*/Q_j,2^j/6\}\rfloor\) for \(j\in\good\),
with \(\alpha_*/0=\infty\).
Since both arguments of the minimum are at least two on
\(\good\), we have \(M_jQ_j\le\alpha_*\) and
\begin{equation}\label{eq:rounding}
\begin{aligned}
M_j
&\ge\frac12\min\{\alpha_*/Q_j,2^j/6\},\\
M_j^{-1}
&\le\max\{2Q_j/\alpha_*,12\,2^{-j}\}\\
&\le 2Q_j/\alpha_*+12\,2^{-j},\\
\frac{a_w}{2\sqrt{M_j}}2^{-3j/2}
&=\delta_*2^{-2j}\sqrt{\frac{2^j}{M_j}}\\
&\ge\delta_*2^{-2j}.
\end{aligned}
\end{equation}
Monotonicity of \(c\) and the local charge inequality yield
\(\E W_{c,j}\ge\lambda_0M_jc(\delta_*2^{-2j})\) for \(j\in\good\).
Summing these bounds over disjoint internal epoch edges gives
\begin{equation}\label{eq:good-cost}
\begin{aligned}
 \sum_{j\in\good}M_jc(\delta_*2^{-2j})
 &\le\lambda_0^{-1}\sum_{j\in\good}\E W_{c,j}\\
 &\le\lambda_0^{-1}\E V_{c,\infty}(A)<\infty.
\end{aligned}
\end{equation}
The charge bound covers only good epochs. To assign allowance one elsewhere, we need both a prefix bound on the omitted epochs and summability of their charge weights. The count satisfies
\[
 \#(\{0,\ldots,J\}\setminus\good)
 \le j_0+\frac2{\alpha_*}\sum_{j=0}^JQ_j
 \le j_0+\frac{2C_A}{\alpha_*\kappa}G(J+1)=o(J).
\]
Write \(d_j^{\mathrm{ep}}=c(\delta_*2^{-2j})\) and \(N_{\mathrm{good}}(J)=\#(\good\cap\{0,\ldots,J\})\). Since \(G(J+1)=o(J)\), choose \(J_1\) so that the bound above is at most \((J+1)/2\) for \(J\ge J_1\). The choice \(D_0=(J_1+1)/2\) also covers the earlier prefixes, giving \(N_{\mathrm{good}}(J)\ge\tfrac12(J+1)-D_0\) for all \(J\ge0\).
Since $d_j^{\mathrm{ep}}-d_{j+1}^{\mathrm{ep}}\ge0$, summation by parts and the prefix-count bound give
\[
\begin{aligned}
 \sum_{\substack{j\le J\\j\in\good}}d_j^{\mathrm{ep}}
 &=N_{\mathrm{good}}(J)d_J^{\mathrm{ep}}
   +\sum_{j=0}^{J-1}N_{\mathrm{good}}(j)
      (d_j^{\mathrm{ep}}-d_{j+1}^{\mathrm{ep}})\\
 &\ge\frac12\left\{(J+1)d_J^{\mathrm{ep}}
       +\sum_{j=0}^{J-1}(j+1)(d_j^{\mathrm{ep}}-d_{j+1}^{\mathrm{ep}})\right\}
       -D_0d_0^{\mathrm{ep}}\\
 &=\frac12\sum_{j=0}^Jd_j^{\mathrm{ep}}-D_0d_0^{\mathrm{ep}}.
\end{aligned}
\]
Consequently,
\[
\begin{aligned}
 \sum_{j=0}^Jd_j^{\mathrm{ep}}
 &\le2\sum_{\substack{j\le J\\j\in\good}}d_j^{\mathrm{ep}}+2D_0d_0^{\mathrm{ep}}\\
 &\le2\sum_{\substack{j\le J\\j\in\good}}M_jd_j^{\mathrm{ep}}+2D_0d_0^{\mathrm{ep}}\\
 &\le2\lambda_0^{-1}\E V_{c,\infty}(A)+2D_0d_0^{\mathrm{ep}}<\infty.
\end{aligned}
\]
Letting $J\to\infty$ proves $\sum_{j\ge0}d_j^{\mathrm{ep}}<\infty$. This is where the converse uses $G(J)=o(J)$.
\Needspace{5\baselineskip}
Set $r_j^{\mathrm{ep}}=M_j^{-1}$ on $\good$ and $r_j^{\mathrm{ep}}=1$ elsewhere. The prefix count controls the added allowance, and summability of $d_j^{\mathrm{ep}}$ controls its added cost. For a bad epoch $j\ge j_0$, one has $1<2Q_j/\alpha_*$. Hence, with $C_1=2C_A/(\alpha_*\kappa)$ and $C_2=j_0+24$,
\begin{align}
 \sum_{j=0}^{J}r_j^{\mathrm{ep}}
 &\le j_0+\frac2{\alpha_*}\sum_{j=0}^J Q_j
   +12\sum_{j\ge0}2^{-j}\notag\\
 &\le C_1G(J+1)+C_2,\label{eq:epoch-prefix}\\
 \sum_{j\ge0}\frac{c(\delta_*2^{-2j})}{r_j^{\mathrm{ep}}}
 &=\sum_{j\in\good}M_jd_j^{\mathrm{ep}}
   +\sum_{j\notin\good}d_j^{\mathrm{ep}}<\infty.\label{eq:epoch-cost}
\end{align}
\paragraph{Distance allocation.}
To convert \eqref{eq:epoch-prefix}--\eqref{eq:epoch-cost} to distance weights, choose an integer \(s\ge0\) with \(2^{-s}\le\delta_*\). Then \(2^{-(2j+s+r)}\le\delta_*2^{-2j}\) for \(r=0,1\), so monotonicity of \(c\) bounds both new weights by the original one. Initialize \(\widetilde a_i=1\) for \(1\le i\le s+1\), and set \(\widetilde a_{2j+s}=\widetilde a_{2j+s+1}=r_j^{\mathrm{ep}}\) for \(j\ge1\). The corresponding weights satisfy \(c(2^{-(2j+s+r)})\le c(\delta_*2^{-2j})\) for \(r=0,1\).
Each epoch contributes at most two allowances to a distance prefix, including a prefix ending inside a pair. Put $C_{\mathrm{pref}}=s+1+2C_1+2C_2$ and $p_m=\lfloor(m-s)/2\rfloor$. For $m\ge s+2$, $p_m+1\le m$, and hence
\[
\begin{aligned}
 \sum_{i\ge1}\frac{c(2^{-i})}{\widetilde a_i}
 &\le\sum_{i=1}^{s+1}c(2^{-i})
       +2\sum_{j\ge1}\frac{c(\delta_*2^{-2j})}{r_j^{\mathrm{ep}}}<\infty,\\
 \sum_{i=1}^m\widetilde a_i
 &\le s+1+2\sum_{j=0}^{p_m}r_j^{\mathrm{ep}}\\
 &\le s+1+2C_1G(p_m+1)+2C_2\\
 &\le C_{\mathrm{pref}}G(m).
\end{aligned}
\]
For $m\le s+1$, the prefix is $m\le(s+1)G(m)$. Thus $a_i=\widetilde a_i/C_{\mathrm{pref}}$ satisfies every constraint of \eqref{eq:allocation} with finite cost, proving (i)$\Rightarrow$(ii).

\Needspace{9\baselineskip}
\subsection{Sufficiency: one schedule with all cost moments}\label{sec:proof-attainment}
Suppose $\Alloc(c,G)<\infty$. Choose a feasible allocation with $\sum_{i\le m}a_i\le G(m)$ for every $m$ and $\sum_i\phi_i(a_i)<\infty$. Since $G(m)=o(m)$, Lemma~\ref{lem:summable} (Appendix~\ref{app:summable}) gives $\sum_i d_i<\infty$.

\paragraph{Schedule.}
Assign $a_i$ to epochs $j=4i,\ldots,4i+3$. Direct checkpoint jumps use cutoff $2^{-j/4}$, unlike the smaller distance forced along an arbitrary action path in the converse. The mean-square jump bound is $O(2^{-j})$, so this cutoff gives exceptional probability $O(2^{-j/2})$. Capping the number of blocks at $(j+1)^2$ makes every cost-moment estimate summable and adds only the summable excess allowance $(j+1)^{-2}$. Retain $\B_0$ before count $16$. For $j\ge4$, put $i(j)=\lfloor j/4\rfloor$ and set
\[
 k_j=\begin{cases}
 \min\{2^j,(j+1)^2,\lceil a_{i(j)}^{-1}\rceil\},&a_{i(j)}>0,\\
 \min\{2^j,(j+1)^2\},&a_{i(j)}=0,
 \end{cases}
\]
\begin{equation}\label{eq:mesh}
\begin{aligned}
 t_{j,\ell}&=2^j+\lfloor\ell2^j/k_j\rfloor,\qquad 0\le\ell\le k_j,\\
 \A_n&=\B_{t_{j,\ell}},\qquad t_{j,\ell}\le n<t_{j,\ell+1},\quad 0\le\ell<k_j.
\end{aligned}
\end{equation}
Here $k_j$ counts retention blocks. The map $j=4i+r$ gives $2^{-j/4}\le2^{-i}$, matching ordinary jumps to the charge weight $d_i$. The mesh has positive block lengths, and $t_{j,\ell+1}-t_{j,\ell}\le\lceil2^j/k_j\rceil$. Adjacent epochs share the checkpoint at $2^{j+1}$, where the current Bayes action is issued once. Appendix~\ref{app:full-proof-attainment} verifies the rounding and edge assignment.

\paragraph{Excess.}
For each block $[a,b)$, posterior conditioning and monotonicity of $g_n$ give
\[
\begin{aligned}
 \sum_{n=a}^{b-1}\E[r_n(\B_a)-r_n(\B_n)]
 &=\sum_{n=a}^{b-1}(g_a-g_n)\\
 &\le(b-a)(g_a-g_b).
\end{aligned}
\]
The definition of $k_j$ gives, for $a_{i(j)}>0$,
\[
\begin{aligned}
 k_j^{-1}
 &=\max\{2^{-j},(j+1)^{-2},\lceil a_{i(j)}^{-1}\rceil^{-1}\}\\
 &\le2^{-j}+(j+1)^{-2}+a_{i(j)}.
\end{aligned}
\]
For $a_{i(j)}=0$, the last entry is omitted and the bound still holds.
The block endpoints satisfy $t_{j,0}=2^j$ and $t_{j,k_j}=2^{j+1}$, so
\[
 \sum_{\ell=0}^{k_j-1}(g_{t_{j,\ell}}-g_{t_{j,\ell+1}})
 =g_{t_{j,0}}-g_{t_{j,k_j}}
 =g_{2^j}-g_{2^{j+1}}.
\]
Using this identity and the block-length bound gives
\begin{equation}\label{eq:mesh-excess}
\begin{aligned}
 \EE_{2^J}(A)
 &\le\EE_{16}(A)+\sum_{j=4}^{J-1}\lceil2^j/k_j\rceil(g_{2^j}-g_{2^{j+1}})\\
 &\le\EE_{16}(A)+\sum_{j=4}^{J-1}(2^j/k_j+1)g_{2^j}\\
 &\le\EE_{16}(A)+C_0\sum_{j=4}^{J-1}(k_j^{-1}+2^{-j})\\
 &\le C+C_0\sum_{j=4}^{J-1}a_{\lfloor j/4\rfloor}\\
 &\le C+4C_0\sum_{i=1}^{\lfloor(J-1)/4\rfloor}a_i\\
 &\le C+4C_0G(\lfloor(J-1)/4\rfloor)\le C'G(J),\qquad J\ge5.
\end{aligned}
\end{equation}
The last comparison uses $\lfloor(J-1)/4\rfloor\le J$, so it also covers a horizon ending inside a four-epoch group.

\paragraph{Lifetime charge.}
At every block exit, put \(\Delta_{j,\ell}=\norm{\B_{t_{j,\ell+1}}-\B_{t_{j,\ell}}}\) and \(H_{j,\ell}=\{\Delta_{j,\ell}>2^{-j/4}\}\). For \(a_i>0\) and \(r\in\{0,1,2,3\}\), the ordinary-charge bound is
\[
\begin{aligned}
 k_{4i+r}c(2^{-(4i+r)/4})
 &\le\lceil a_i^{-1}\rceil d_i\\
 &\le d_i+\frac{d_i}{a_i}=d_i+\phi_i(a_i).
\end{aligned}
\]
If $a_i=0$, finite allocation cost forces $d_i=0$ and the left side is zero. Summing the four epochs assigned to each $i$ gives
\[
 \sum_{j\ge4}k_jc(2^{-j/4})
 \le4\sum_{i\ge1}\{d_i+\phi_i(a_i)\}<\infty.
\]
Exceptional jumps satisfy, by \eqref{eq:risk-energy}, Markov's inequality and \eqref{eq:basic-action-bounds},
\begin{equation}\label{eq:exception-prob}
 \Pp(H_{j,\ell})\le2^{j/2}\E\Delta_{j,\ell}^2
 \le\kappa^{-1}2^{j/2}(g_{t_{j,\ell}}-g_{t_{j,\ell+1}})
 \le (C_0/\kappa)2^{-j/2}.
\end{equation}
There is one initialization edge at observation count $16$. Every subsequent adjustment occurs at a block exit. The exit edge at $2^{j+1}$ belongs to epoch $j$. For ordinary and exceptional jumps,
\(
 c(\Delta_{j,\ell})\le c(2^{-j/4})+c(1)\ind_{H_{j,\ell}}.
\)
Consequently, for any finite $J\ge4$,
\begin{equation}\label{eq:cost-decomposition}
 V_{c,2^{J+1}+1}(A)\le c(1)+\sum_{j=4}^Jk_jc(2^{-j/4})
 +c(1)\sum_{j=4}^J\sum_{\ell<k_j}\ind_{H_{j,\ell}}.
\end{equation}
For $q\ge1$ write $\norm{Y}_{L^q}=(\E|Y|^q)^{1/q}$. Minkowski, $\norm{\ind_H}_{L^q}=\Pp(H)^{1/q}$ and $k_j\le(j+1)^2$ give
\[
\begin{aligned}
 \norm{V_{c,2^{J+1}+1}(A)}_{L^q}
 &\le c(1)+\sum_{j=4}^{J}k_jc(2^{-j/4})
   +c(1)\sum_{j=4}^{J}\sum_{\ell<k_j}\Pp(H_{j,\ell})^{1/q}\\
 &\le c(1)+4\sum_{i\ge1}\{d_i+\phi_i(a_i)\}\\
 &\quad+c(1)(C_0/\kappa)^{1/q}\sum_{j\ge4}(j+1)^22^{-j/(2q)}=:K_q<\infty.
\end{aligned}
\]
The bound is uniform in $J$. Monotone convergence and Jensen's inequality give
\begin{equation}\label{eq:all-moments}
\begin{aligned}
 \E[V_{c,\infty}(A)^q]
 &=\lim_{J\to\infty}\norm{V_{c,2^{J+1}+1}(A)}_{L^q}^q
 \le K_q^q<\infty &&(q\ge1),\\
 \E[V_{c,\infty}(A)^q]
 &\le(\E V_{c,\infty}(A))^q<\infty &&(0<q<1).
\end{aligned}
\end{equation}
The schedule is independent of $q$, proving (ii)$\Rightarrow$(iv). Lemmas~\ref{lem:infinite} and~\ref{lem:duality} (Appendices~\ref{app:infinite}--\ref{app:dual}) give (ii)$\Leftrightarrow$(iii), while (iv)$\Rightarrow$(i) follows by definition.

\Needspace{10\baselineskip}
\section{Discussion and limitations}\label{sec:discussion}
The attainable region is unchanged by adaptive timing or private randomization. For power profiles, a charge-independent checkpoint schedule also achieves the sharp revision order.

Action information rules out a common optimum across latent states, for which retention has zero excess and zero adjustment charge. A fixed positive revision fee gives $c(0+):=\lim_{x\downarrow0}c(x)>0$, and the constant-weight dual grows at least as $c(0+)m^2/G(m)$. A zero-cost neighborhood makes only finitely many $d_i$ positive and allows bounded excess. A charge supported only at $x=1$ has $\Alloc(c,G)=0$, though an initialization edge may still carry a charge.

Sublinearity of $G$ enters the converse through the density of low-error epochs. The fixed-fee example in Appendix~\ref{app:unrestricted} has $G(m)=m^3$ and a finite allocation value, yet no policy attains that profile. The assumptions apply to Bayesian prediction in fixed-dimensional action spaces, with exogenous observations and smooth population risks. Appendix~\ref{app:metric} isolates three probabilistic premises under which the allocation reduction also holds in metric action spaces. Nonsmooth population risks, action-dependent observations and misspecified posteriors call for corresponding motion and retention estimates. Two questions remain: whether uniform second-moment boundedness alone can replace UI2, and how to compute checkpoint times from finite allocation certificates.

\label{page:main-end}
\label{page:references-start}

\clearpage
\appendix
\label{page:appendices-start}
\section{Statistical foundations and action moments}\label{app:foundations}
Lemma~\ref{lem:gradients} establishes conditioning identities. Lemma~\ref{lem:accuracy} constructs an auxiliary estimator, and Appendix~\ref{app:action-moments} transfers its moment bounds to $\B_n$. Lemmas~\ref{lem:geometry}--\ref{lem:ui} in Appendix~\ref{app:motion} then supply the localization and uniform integrability used in Proposition~\ref{prop:motion}. Observation counts are deterministic throughout Appendices~\ref{app:foundations}--\ref{app:motion}.

\subsection{Measurable minimizers and posterior kernels}\label{app:kernels}
A continuous $\mu$-strongly convex function $F$ minimized at $\x\in K$ satisfies, for $\y\in K$ and $0<t<1$,
\begin{align*}
 0&\le F((1-t)\x+t\y)-F(\x)\\
  &\le t\{F(\y)-F(\x)\}-\frac\mu2t(1-t)\norm{\y-\x}^2.
\end{align*}
Hence
\[
 F(\y)-F(\x)
 \ge\sup_{0<t<1}(\mu/2)(1-t)\norm{\y-\x}^2
 =(\mu/2)\norm{\y-\x}^2.
\]
For a jointly measurable field of continuous strongly convex objectives $F_\omega$, compactness gives a unique minimizer $\x_\omega$. Fix a common countable dense sequence $(\uvec_i)$ in $K$. Put
$m(\omega)=\inf_iF_\omega(\uvec_i)=F_\omega(\x_\omega)$ and
$i_j=\min\{i:F_\omega(\uvec_i)<m(\omega)+1/j\}$. Then
\[
 \{i_j=i\}=\{F_\omega(\uvec_i)<m+1/j\}
       \cap\bigcap_{k<i}\{F_\omega(\uvec_k)\ge m+1/j\},
\]
\[
 \norm{\uvec_{i_j}-\x_\omega}^2
 \le\frac2\mu\{F_\omega(\uvec_{i_j})-m(\omega)\}
 <\frac2{\mu j}\longrightarrow0.
\]
Thus the minimizer is measurable. Assumption~\ref{ass:curvature} supplies the population versions. For a convergent action sequence $\w_k\to\w$, population continuity and the bound $|R_u(\w_k)-R_u(\w)|\le2\int H_0(z)P_u(\dd z)$ permit dominated convergence under $\Pi_n$. The dominating function is posterior-integrable almost surely, since its unconditional expectation is finite. Thus predictive risks are continuous even though sample losses need not be continuous. For each $n$, the event on which $\Pi_n(\Theta_0)=1$ and the envelope is posterior-integrable belongs to $\G_n$ and has probability one. Define $\B_n$ by minimization on this event and as a fixed action in $K$ on its complement. The preceding construction makes $\B_n$ $\G_n$-measurable. Intersect these events over $n$ only to obtain simultaneous assertions. This intersection does not redefine the actions.

Write $\Borel(S)$ for the Borel sigma-field of a space $S$. For deterministic integers $n\ge0$, $m\ge n$, $t>n$, let $\W$ be $\F_n$-measurable. For $A\in\Borel(\Theta)$, $Q\in\G_m$ and $E_\Xi\in\sigma(\Xi)$, independence of the tape gives
\begin{align*}
 \E[\ind_Q\ind_{E_\Xi}\ind_A(\theta)]
 &=\Pp(E_\Xi)\E[\ind_Q\ind_A(\theta)]
 =\Pp(E_\Xi)\E[\ind_Q\Pi_m(A)]\\
 &=\E[\ind_Q\ind_{E_\Xi}\Pi_m(A)].
\end{align*}
A monotone-class argument in $Q\cap E_\Xi$ yields $\Law(\theta\mid\F_m)=\Pi_m$. For $Q\in\F_m$ and $D\in\Borel(K)$,
\begin{align*}
 \E[\ind_Q\ind_A(\theta)\ind_D(\W)]
 &=\E[\ind_Q\ind_D(\W)\E[\ind_A(\theta)\mid\F_m]]\\
 &=\E\left[\ind_Q\int\ind_A(u)\ind_D(\W)\Pi_m(\dd u)\right].
\end{align*}
Functional monotone classes and integrable truncation give the random-action identity
\[
 \E[h(\theta,\W)\mid\F_m]=\int h(u,\W)\Pi_m(\dd u)
\]
for jointly measurable $h$ whenever the evaluated field is integrable. The envelope ensures this for $h(\theta,\w)=R_\theta(\w)$.
Similarly, conditional iid sampling gives, for $Q\in\sigma(\theta)\vee\F_n$ and $C\in\Borel(\mathcal Z)$,
\begin{align*}
 \E[\ind_Q\ind_D(\W)\ind_C(Z_t)]
 &=\E[\ind_Q\ind_D(\W)P_\theta(C)]\\
 &=\E\left[\ind_Q\int\ind_D(\W)\ind_C(z)P_\theta(\dd z)\right].
\end{align*}
Therefore
\[
 \E[k(Z_t,\W)\mid\theta,\F_n]=\int k(z,\W)P_\theta(\dd z)
\]
for integrable jointly measurable $k$. Nonnegative truncation also covers conditional moment and tail fields. On an auxiliary extension, an independent uniform variable samples $\theta'_n$ from the posterior kernel, so $\Law(\theta,\theta'_n\mid\F_n)=\Pi_n\otimes\Pi_n$. This posterior copy is used only in proofs.

\begin{lemma}[Common predictive gradients and conditioning]\label{lem:gradients}
Under Assumptions~\ref{ass:curvature}--\ref{ass:witness}, population and predictive gradients have common jointly measurable versions bounded in norm by \(\Hgrad=\sqrt{M_2}\), and predictive gradients are \(L\)-Lipschitz. For deterministic integers \(n\ge0\), \(m\ge n\), \(t>n\), and an \(\F_n\)-measurable action \(\W\), we have
\[
 \E[\nabla R_\theta(\W)\mid\F_m]=\nabla r_m(\W),\qquad
 \E[\xiobs_{Z_t}(\W)\mid\theta,\F_n]=\nabla R_\theta(\W).
\]
The future witness has conditional second moment at most \(M_2\) and conditional square-tail expectation at most \(\tau(R)\) at cutoff \(R\).
\end{lemma}
\begin{proof}[Proof of Lemma~\ref{lem:gradients}]
Write $\Theta_0$ for the fixed prior-full set. Define
\[
 H_\theta^0=\int H_0(z)P_\theta(\dd z),\qquad
 \hvec(\theta,\w)=\int\xiobs_z(\w)P_\theta(\dd z)=\nabla R_\theta(\w)
 \quad(\theta\in\Theta_0),
\]
and extend $\hvec$ by zero outside $\Theta_0$. Kernel measurability and Cauchy--Schwarz give
\begin{gather*}
 \norm{\hvec(\theta,\w)}^2\le\int\norm{\xiobs_z(\w)}^2P_\theta(\dd z)\le M_2,\\
 \norm{\hvec(\theta,\uvec)-\hvec(\theta,\w)}\le L\norm{\uvec-\w}
       \quad(\theta\in\Theta_0),\\
 \E\int H_u^0\Pi_n(\dd u)=\E H_\theta^0=\E H_0(Z_1)<\infty,
 \qquad \E\Pi_n(\Theta_0^c)=\pi(\Theta_0^c)=0.
\end{gather*}
Set $\mathbf h_n(\w)=\int\hvec(u,\w)\Pi_n(\dd u)$. The event
\[
 \Omega_0=\bigcap_{n\ge0}\left\{\Pi_n(\Theta_0)=1,\quad
                            \int H_u^0\Pi_n(\dd u)<\infty\right\}
\]
has probability one. On it, define
$T_n(\uvec,\w)=r_n(\uvec)-r_n(\w)-\ip{\mathbf h_n(\w)}{\uvec-\w}$. For $\uvec,\w\in K$,
\begin{align*}
 T_n(\uvec,\w)
 &=\int\left[R_v(\uvec)-R_v(\w)-\ip{\nabla R_v(\w)}{\uvec-\w}\right]\Pi_n(\dd v)\\
 &=\int\int_0^1\ip{\nabla R_v(\w+s(\uvec-\w))-\nabla R_v(\w)}{\uvec-\w}\,\dd s\,\Pi_n(\dd v),\\
 |T_n(\uvec,\w)|
 &\le\int\int_0^1Ls\norm{\uvec-\w}^2\,\dd s\,\Pi_n(\dd v)
 =\frac L2\norm{\uvec-\w}^2,\\
 T_n(\uvec,\w)&\ge\int\frac\mu2\norm{\uvec-\w}^2\Pi_n(\dd v)
 =\frac\mu2\norm{\uvec-\w}^2.
\end{align*}
The common remainder bound justifies these kernel integrations. Moreover,
\begin{align*}
 \norm{\mathbf h_n(\w)}&\le\int\norm{\hvec(v,\w)}\Pi_n(\dd v)\le\Hgrad,\\
 \norm{\mathbf h_n(\uvec)-\mathbf h_n(\w)}
 &\le\int\norm{\hvec(v,\uvec)-\hvec(v,\w)}\Pi_n(\dd v)\le L\norm{\uvec-\w}.
\end{align*}
For $\w\in\ri K$, $\vvec\in\R^d$ and sufficiently small $s\ne0$,
\[
 \left|\frac{r_n(\w+s\vvec)-r_n(\w)}s-\ip{\mathbf h_n(\w)}{\vvec}\right|
 \le\frac{L|s|}2\norm{\vvec}^2\longrightarrow0.
\]
Thus $\nabla r_n=\mathbf h_n$ on the relative interior, with a continuous extension to $K$.
The random-action identity applies coordinatewise to the bounded common field:
\[
 \E[\nabla R_\theta(\W)\mid\F_m]
 =\E[\hvec(\theta,\W)\mid\F_m]
 =\int\hvec(v,\W)\Pi_m(\dd v)=\nabla r_m(\W).
\]
The future-observation identity and nonnegative truncation give
\begin{align*}
 \E[\xiobs_{Z_t}(\W)\mid\theta,\F_n]
 &=\int\xiobs_z(\W)P_\theta(\dd z)=\nabla R_\theta(\W),\\
 \E[\norm{\xiobs_{Z_t}(\W)}^2\mid\theta,\F_n]
 &=\int\norm{\xiobs_z(\W)}^2P_\theta(\dd z)\le M_2,\\
 \E[\norm{\xiobs_{Z_t}(\W)}^2\ind_{\{\norm{\xiobs_{Z_t}(\W)}>R\}}
          \mid\theta,\F_n]
 &=\int\norm{\xiobs_z(\W)}^2\ind_{\{\norm{\xiobs_z(\W)}>R\}}P_\theta(\dd z)
 \le\tau(R).
\end{align*}
These identities evaluate the common kernel fields at observable actions $\W$. Appendix~\ref{app:action-moments} treats latent-action evaluations pathwise.
\end{proof}

For a $\G_n$-measurable $\W$, the tower property gives the same identities with $\F_m,\F_n$ replaced by $\G_m,\G_n$.

\subsection{Robust auxiliary estimation}\label{app:robust}
The auxiliary estimator is used only in the proof. Five conditionally independent block estimates turn second-moment accuracy into the fourth-moment bound needed below.
Fix an initial action $\winit\in K$ and let $\Proj_K(\vvec)=\argmin_{\w\in K}\norm{\vvec-\w}^2$ be Euclidean projection. On each deterministic block relabeled $\widetilde Z_1,\widetilde Z_2,\ldots$, initialize $\iter_0=\winit$ and use
\begin{equation}\label{eq:projected-recursion}
 \iter_l=\Proj_K\left(\iter_{l-1}-\frac{2\xiobs_{\widetilde Z_l}(\iter_{l-1})}{\mu(l+5)}\right),\qquad l\ge1.
\end{equation}
For \(n<5\), set \(\bhat_n=\winit\). For \(n\ge5\), partition the first \(n\) observations into five consecutive deterministic blocks of lengths \(\ell_r\) differing by at most one. Run \eqref{eq:projected-recursion} on each block, restarting at \(\winit\), and write \(\iter_{n,r}=\iter_{\ell_r}^{(r)}\) for its terminal candidate. For each \(r\in\{1,\ldots,5\}\), let \(\radmed_r\) be the third smallest of the five distances \(\norm{\iter_{n,r}-\iter_{n,k}}\), including \(k=r\). Set \(r_\star=\min\argmin_{1\le r\le5}\radmed_r\) and \(\bhat_n=\iter_{n,r_\star}\). The index rule resolves ties. If at least three candidates are close to the target, so is the selected candidate.

\begin{lemma}[Auxiliary accuracy and posterior moments]\label{lem:accuracy}
Under Assumptions~\ref{ass:curvature}--\ref{ass:witness}, there is a $\G_n$-measurable estimator $\bhat_n\in K$ and a finite $C_4$, uniform on the fixed prior-full set, such that, for $n\ge1$,
\[
 \begin{aligned}
 \E_\theta\norm{\bhat_n-\bb_\theta}^2&\le\frac{\sqrt{C_4}}n,
 &\E_\theta\norm{\bhat_n-\bb_\theta}^4&\le\frac{C_4}{n^2},\\
 \E\norm{\bb_\theta-\bb_{\theta'_n}}^4&\le\frac{C_{\mathrm{copy}}}{n^2},
 &0\le g_n&\le\frac{C_0}n.
 \end{aligned}
\]
Here $C_4$ depends only on $\mu,M_2,D_K$, $C_{\mathrm{copy}}$ is a universal multiple of $C_4$, and $C_0=L\sqrt{C_4}/2$ is admissible.
\end{lemma}
\begin{proof}[Proof of Lemma~\ref{lem:accuracy}]
Write $a=2/\mu$ in the recursion \eqref{eq:projected-recursion}.

\emph{One-block accuracy.} For $l\ge1$, write $t=l+5$, $\xiobs_l=\xiobs_{\widetilde Z_l}(\iter_{l-1})$,
$\mathcal H_{l-1}^{\mathrm{block}}=\sigma(\widetilde Z_1,\ldots,\widetilde Z_{l-1})$ and
$\E_{\theta,l-1}=\E_\theta[\,\cdot\mid\mathcal H_{l-1}^{\mathrm{block}}]$.
Nonexpansiveness gives
\begin{align*}
 \norm{\iter_l-\bb_\theta}^2
 &=\norm{\Proj_K(\iter_{l-1}-a\xiobs_l/t)-\Proj_K(\bb_\theta)}^2\\
 &\le\norm{\iter_{l-1}-\bb_\theta-a\xiobs_l/t}^2\\
 &=\norm{\iter_{l-1}-\bb_\theta}^2
  -\frac{2a}t\ip{\iter_{l-1}-\bb_\theta}{\xiobs_l}
  +\frac{a^2}{t^2}\norm{\xiobs_l}^2.
\end{align*}
Conditional iid sampling yields
\[
 \E_{\theta,l-1}\xiobs_l=\int\xiobs_z(\iter_{l-1})P_\theta(\dd z)
 =\nabla R_\theta(\iter_{l-1}),\qquad
 \E_{\theta,l-1}\norm{\xiobs_l}^2
 =\int\norm{\xiobs_z(\iter_{l-1})}^2P_\theta(\dd z)\le M_2.
\]
Both boundary alternatives imply $\nabla R_\theta(\bb_\theta)=0$, so
\begin{align*}
 \E_{\theta,l-1}\norm{\iter_l-\bb_\theta}^2
 &\le\norm{\iter_{l-1}-\bb_\theta}^2
  -\frac{2a}t\ip{\iter_{l-1}-\bb_\theta}{\E_{\theta,l-1}\xiobs_l}
  +\frac{a^2}{t^2}\E_{\theta,l-1}\norm{\xiobs_l}^2\\
 &\le\norm{\iter_{l-1}-\bb_\theta}^2
  -\frac{2a}t\ip{\iter_{l-1}-\bb_\theta}{\nabla R_\theta(\iter_{l-1})-\nabla R_\theta(\bb_\theta)}
  +\frac{a^2M_2}{t^2}\\
 &\le(1-2a\mu/t)\norm{\iter_{l-1}-\bb_\theta}^2+\frac{a^2M_2}{t^2}\\
 &=(1-4/t)\norm{\iter_{l-1}-\bb_\theta}^2+\frac{a^2M_2}{t^2}.
\end{align*}
For $u_l=\E_\theta\norm{\iter_l-\bb_\theta}^2$,
\[
 u_l=\E_\theta[\E_{\theta,l-1}\norm{\iter_l-\bb_\theta}^2]
 \le(1-4/t)u_{l-1}+a^2M_2/t^2.
\]
Choose $C_{\mathrm{iter}}$ depending only on $\mu,M_2,D_K$, with $C_{\mathrm{iter}}\ge5D_K^2$ and $3C_{\mathrm{iter}}\ge a^2M_2$.
Since $u_0\le D_K^2\le C_{\mathrm{iter}}/5$ and $1-4/t\ge0$ for $t\ge6$,
\[
\begin{aligned}
 \frac{C_{\mathrm{iter}}}t-(1-4/t)\frac{C_{\mathrm{iter}}}{t-1}
 &=\frac{3C_{\mathrm{iter}}}{t(t-1)}\ge\frac{a^2M_2}{t^2},\\
 u_l&\le(1-4/t)\frac{C_{\mathrm{iter}}}{t-1}
       +\frac{a^2M_2}{t^2}\le\frac{C_{\mathrm{iter}}}t.
\end{aligned}
\]
Induction gives $u_l\le C_{\mathrm{iter}}/(l+5)$.

\emph{Five-block selection.} The five candidates \(\iter_{n,r}=\iter_{\ell_r}^{(r)}\) use disjoint deterministic blocks and are independent given \(\theta\). Since \(\ell_r\ge\lfloor n/5\rfloor\ge n/10\), we have
\[
\begin{aligned}
 \E_\theta\norm{\iter_{n,r}-\bb_\theta}^2
 &\le\frac{C_{\mathrm{iter}}}{\ell_r+5}\le\frac{10C_{\mathrm{iter}}}{n},\\
 \Pp_\theta\{\norm{\iter_{n,r}-\bb_\theta}>u\}
 &\le\min\{1,10C_{\mathrm{iter}}/(nu^2)\}.
\end{aligned}
\]
Finite order statistics and the deterministic tie rule make the selection measurable. Define \(I_{\mathrm{good}}(u)=\{r:\norm{\iter_{n,r}-\bb_\theta}\le u\}\) and \(\mathcal N_r=\{k:\norm{\iter_{n,r}-\iter_{n,k}}\le\radmed_r\}\). If \(|I_{\mathrm{good}}(u)|\ge3\), then
\begin{gather*}
 r,k\in I_{\mathrm{good}}(u)
 \ \Longrightarrow\ 
 \norm{\iter_{n,r}-\iter_{n,k}}
 \le\norm{\iter_{n,r}-\bb_\theta}+\norm{\iter_{n,k}-\bb_\theta}\le2u,\\
 \radmed_{r_\star}=\min_r\radmed_r
 \le\max_{r\in I_{\mathrm{good}}(u)}\radmed_r\le2u,\\
 |\mathcal N_{r_\star}\cap I_{\mathrm{good}}(u)|
 \ge|\mathcal N_{r_\star}|+|I_{\mathrm{good}}(u)|-5\ge1.
\end{gather*}
For $k$ in the intersection,
\[
\begin{aligned}
 \norm{\bhat_n-\bb_\theta}
 &\le\norm{\iter_{n,r_\star}-\iter_{n,k}}+\norm{\iter_{n,k}-\bb_\theta}\\
 &\le\radmed_{r_\star}+u\le3u.
\end{aligned}
\]
With $F_r(u)=\{\norm{\iter_{n,r}-\bb_\theta}>u\}$ and three-element subsets $T\subset\{1,\ldots,5\}$,
\[
 \{\norm{\bhat_n-\bb_\theta}>3u\}
 \subseteq\{|I_{\mathrm{good}}(u)|\le2\}
 =\bigcup_{|T|=3}\bigcap_{r\in T}F_r(u).
\]
Conditional independence therefore gives
\begin{equation}\label{eq:robust-tail}
\begin{aligned}
 \Pp_\theta\{\norm{\bhat_n-\bb_\theta}>3u\}
 &\le\min\left\{1,\sum_{|T|=3}\Pp_\theta\!\left(\bigcap_{r\in T}F_r(u)\right)\right\}\\
 &=\min\left\{1,\sum_{|T|=3}\prod_{r\in T}\Pp_\theta(F_r(u))\right\}\\
 &\le\min\{1,10(10C_{\mathrm{iter}})^3/(n^3u^6)\}.
\end{aligned}
\end{equation}
\emph{Tail integration.}
Rescaling $u$ in \eqref{eq:robust-tail} gives $\Pp_\theta\{\norm{\bhat_n-\bb_\theta}>u\}\le\min\{1,C/(n^3u^6)\}$ for a fixed $C$ depending only on $\mu,M_2,D_K$. Hence
\[
 \begin{aligned}
 \E_\theta\norm{\bhat_n-\bb_\theta}^4
 &=4\int_0^\infty u^3\Pp_\theta\{\norm{\bhat_n-\bb_\theta}>u\}\,\dd u\\
 &\le4\int_0^{n^{-1/2}}u^3\,\dd u
       +\frac{4C}{n^3}\int_{n^{-1/2}}^\infty u^{-3}\,\dd u
 =\frac{1+2C}{n^2}.
 \end{aligned}
\]
Choose \(C_4\ge1+2C\), enlarging it for \(n<5\), where \(\bhat_n=\winit\). Then \(\E_\theta\norm{\bhat_n-\bb_\theta}^4\le C_4/n^2\) and
\[
 \E_\theta\norm{\bhat_n-\bb_\theta}^2
 \le(\E_\theta\norm{\bhat_n-\bb_\theta}^4)^{1/2}
 \le\sqrt{C_4}/n.
\]

\emph{Posterior and risk bounds.} Predictive optimality and the random-action identity yield
\begin{align*}
 g_n&=\E[r_n(\B_n)-R_\theta(\bb_\theta)]\\
 &\le\E[r_n(\bhat_n)-R_\theta(\bb_\theta)]
 =\E[R_\theta(\bhat_n)-R_\theta(\bb_\theta)]\\
 &\le\frac L2\E\norm{\bhat_n-\bb_\theta}^2\le\frac{L\sqrt{C_4}}{2n}.
\end{align*}
The quadratic upper bound uses population stationarity. Given $\F_n$, the posterior parameters have the same marginal law and $\bhat_n$ is fixed. Consequently,
\begin{align*}
 \E[\norm{\bb_\theta-\bb_{\theta'_n}}^4\mid\F_n]
 &\le8\E[\norm{\bb_\theta-\bhat_n}^4+\norm{\bb_{\theta'_n}-\bhat_n}^4\mid\F_n]\\
 &=16\int\norm{\bb_u-\bhat_n}^4\Pi_n(\dd u),\\
 \E\norm{\bb_\theta-\bb_{\theta'_n}}^4
 &\le16\E\norm{\bb_\theta-\bhat_n}^4\le16C_4/n^2.
\end{align*}
Take $C_{\mathrm{copy}}=16C_4$.

\end{proof}

\subsection{Risk energy and Bayes-action moments}\label{app:action-moments}
Equation~\eqref{eq:risk-energy} and Lemma~\ref{lem:accuracy} give an aggregate bound on motion. For ordered deterministic windows with disjoint internal edges and integer endpoints $N\le n_1<m_1\le n_2<\cdots\le n_M<m_M$, where $N,M\ge1$,
\[
\begin{aligned}
 \kappa\sum_{\ell=1}^M\E\norm{\B_{m_\ell}-\B_{n_\ell}}^2
 &\le\sum_{\ell=1}^M(g_{n_\ell}-g_{m_\ell})\\
 &=g_{n_1}-g_{m_M}-\sum_{\ell=1}^{M-1}(g_{m_\ell}-g_{n_{\ell+1}})\\
 &\le g_{n_1}\le g_N\le C_0/N.
\end{aligned}
\]
\Needspace{5\baselineskip}
Both boundary alternatives give $\nabla R_\theta(\bb_\theta)=0$. Predictive optimality on the convex set $K$ gives the variational inequality
\(
 \ip{\nabla r_n(\B_n)}{\w-\B_n}\ge0\quad(\w\in K).
\)
The inequality holds on one probability-one event for every $\w\in K$, including when $\B_n$ is on the boundary. We may therefore substitute the latent action $\w=\bb_\theta$ pathwise. Strong monotonicity yields
\begin{align*}
 \mu\norm{\B_n-\bb_\theta}^2
 &\le\ip{\nabla r_n(\B_n)-\nabla r_n(\bb_\theta)}{\B_n-\bb_\theta}\\
 &\le-\ip{\nabla r_n(\bb_\theta)}{\B_n-\bb_\theta}\\
 &\le\norm{\nabla r_n(\bb_\theta)}\norm{\B_n-\bb_\theta},\\
 \norm{\nabla r_n(\bb_\theta)}
 &=\left\|\int\{\nabla R_u(\bb_\theta)-\nabla R_u(\bb_u)\}\Pi_n(\dd u)\right\|\\
 &\le\int\norm{\nabla R_u(\bb_\theta)-\nabla R_u(\bb_u)}\Pi_n(\dd u)\\
 &\le L\int\norm{\bb_\theta-\bb_u}\Pi_n(\dd u).
\end{align*}
\Needspace{5\baselineskip}
Dividing when the distance is positive, and including the zero-distance case, gives
\[
 \norm{\B_n-\bb_\theta}
 \le\frac L\mu\int\norm{\bb_\theta-\bb_u}\Pi_n(\dd u).
\]
Jensen's inequality and the posterior-copy law give
\begin{align*}
 \E\norm{\B_n-\bb_\theta}^4
 &\le(L/\mu)^4\E\left(\int\norm{\bb_\theta-\bb_u}\Pi_n(\dd u)\right)^4\\
 &\le(L/\mu)^4\E\int\norm{\bb_\theta-\bb_u}^4\Pi_n(\dd u)\\
 &=(L/\mu)^4\E\left[\E\left[\int\norm{\bb_\theta-\bb_u}^4\Pi_n(\dd u)
                 \,\middle|\,\G_n\right]\right]\\
 &=(L/\mu)^4\E\iint\norm{\bb_v-\bb_u}^4\Pi_n(\dd v)\Pi_n(\dd u)\\
 &=(L/\mu)^4\E\norm{\bb_\theta-\bb_{\theta'_n}}^4
 \le C_Bn^{-2},
\end{align*}
where \(C_B=(L/\mu)^4C_{\mathrm{copy}}\). Population quadratic growth gives
\[
 \kappa\E\norm{\B_n-\bb_\theta}^2
 \le\E[R_\theta(\B_n)-R_\theta(\bb_\theta)]
 =g_n\le C_0/n.
\] Setting \(\Ctwo=C_0/\kappa\), the second and fourth moment bounds needed in Lemmas~\ref{lem:geometry}--\ref{lem:ui} are \(\E\norm{\B_n-\bb_\theta}^2\le\Ctwo/n\) and \(\E\norm{\B_n-\bb_\theta}^4\le C_B/n^2\) for \(n\ge1\).
Assumption~\ref{ass:information}, posterior orthogonality and quadratic growth also give
\[
 \frac{\kappa v}{n}\le\kappa q_n
 \le\kappa\{q_n+\E\norm{\B_n-\bar{\bb}_n}^2\}
 =\kappa\E\norm{\B_n-\bb_\theta}^2
 \le g_n\le\frac{C_0}{n}\qquad(n\ge n_0).
\]
Thus $g_n\asymp n^{-1}$.

\section{Localization and uniform integrability for benchmark motion}\label{app:motion}

\subsection{Observable boundary localization}\label{app:localization}

\begin{lemma}[Observable retained information and constrained geometry]\label{lem:geometry}
Under Assumptions~\ref{ass:curvature}--\ref{ass:information} and (I), there are \(\rho>0\) and \(N_{\mathrm{loc}}<\infty\) such that, for \(n\ge N_{\mathrm{loc}}\), the event \(\loc_n=\{\dist(\B_n,\rbd K)\ge\rho\}\in\G_n\) satisfies \(\E[\ind_{\loc_n}\norm{\B_n-\bb_\theta}^2]\ge v/(2n)\). If \(\x\in K\) has relative boundary distance at least \(\rho\) and \(\y\) minimizes a predictive risk \(r=r_m\) on \(K\), then, with \(K_{\mathrm{geo}}=\max\{L,\Hgrad/\rho\}\),
\[
 \norm{\nabla r(\x)}\le K_{\mathrm{geo}}\norm{\y-\x}.
\] Under (S), take \(\loc_n=\Omega\), \(K_{\mathrm{geo}}=L\), and \(N_{\mathrm{loc}}=n_0\). The comparison at \(\x=\B_n\) and \(\y=\B_m\) also covers boundary minimizers \(\B_m\).
\end{lemma}
\begin{proof}
Under (I), choose \(\delta\in(0,1)\) with \(\sqrt{C_B\delta}\le v/2\). Since \(\bb_\theta\in\ri K\) almost surely, the events \(\{\dist(\bb_\theta,\rbd K)\le2^{1-k}\}\) decrease to \(\{\dist(\bb_\theta,\rbd K)=0\}\) as \(k\to\infty\), and their probabilities decrease to zero.
Choose $\rho=2^{-k}$ so this probability is at most $\delta/2$, and set $N_{\mathrm{loc}}=\lceil\max\{n_0,2\Ctwo/(\delta\rho^2)\}\rceil$. For $n\ge N_{\mathrm{loc}}$,
\begin{align*}
 \E\ip{\B_n-\bar{\bb}_n}{\bb_\theta-\bar{\bb}_n}
 &=\E\ip{\B_n-\bar{\bb}_n}{\E[\bb_\theta-\bar{\bb}_n\mid\G_n]}=0,\\
 \E\norm{\B_n-\bb_\theta}^2
 &=\E\norm{\B_n-\bar{\bb}_n}^2+\E\norm{\bb_\theta-\bar{\bb}_n}^2
 \ge q_n\ge v/n.
\end{align*}
The boundary-distance function is $1$-Lipschitz, so
\[
 \loc_n^c\subseteq\{\dist(\bb_\theta,\rbd K)\le2\rho\}
             \cup\{\norm{\B_n-\bb_\theta}>\rho\}.
\]
Markov's and H\"older's inequalities yield
\begin{align*}
 \Pp(\loc_n^c)
 &\le\Pp\{\dist(\bb_\theta,\rbd K)\le2\rho\}
      +\Pp\{\norm{\B_n-\bb_\theta}>\rho\}\\
 &\le\delta/2+\Ctwo/(n\rho^2)\le\delta,\\
 \E[\ind_{\loc_n}\norm{\B_n-\bb_\theta}^2]
 &=\E\norm{\B_n-\bb_\theta}^2-\E[\ind_{\loc_n^c}\norm{\B_n-\bb_\theta}^2]\\
 &\ge q_n-(\E\norm{\B_n-\bb_\theta}^4)^{1/2}\Pp(\loc_n^c)^{1/2}\\
 &\ge\frac{v-\sqrt{C_B\delta}}n\ge\frac v{2n}.
\end{align*}
These choices precede any window selection. For a minimizer $\y$,
\begin{align*}
 \norm{\nabla r(\x)}&=\norm{\nabla r(\x)-\nabla r(\y)}
 \le L\norm{\x-\y},&&\y\in\ri K,\\
 \norm{\nabla r(\x)}&\le\Hgrad\le(\Hgrad/\rho)\norm{\x-\y},&&\y\in\rbd K.
\end{align*}
The first case uses stationarity, the second $\norm{\x-\y}\ge\dist(\x,\rbd K)\ge\rho$. Under (S),
\begin{align*}
 \E[\ind_{\loc_n}\norm{\B_n-\bb_\theta}^2]
 &=q_n+\E\norm{\B_n-\bar{\bb}_n}^2\ge v/n,\\
 \norm{\nabla r_m(\B_n)}
 &=\norm{\nabla r_m(\B_n)-\nabla r_m(\B_m)}
 \le L\norm{\B_n-\B_m}.
\qedhere
\end{align*}
\end{proof}

\subsection{Uniform integrability with a frozen Bayes action}\label{app:innovation}
The event $F$ in the bound below may depend on future observations.
\begin{lemma}[Uniform integrability of frozen witness sums]\label{lem:ui}
Under Assumptions~\ref{ass:curvature}--\ref{ass:witness}, fix integers \(n\ge1\), \(1\le h\le n\), and an event \(E\in\G_n\). Freeze \(\B_n\) and define
\[
\begin{aligned}
 \U&=\ind_E\nabla R_\theta(\B_n),
 &\Svec&=\ind_E\sum_{t=n+1}^{n+h}\xiobs_{Z_t}(\B_n),\\
 \W_t&=\xiobs_{Z_t}(\B_n)-\nabla R_\theta(\B_n),
 &\Tvec_h&=\sum_{t=n+1}^{n+h}\W_t.
\end{aligned}
\]
Conditional on $(\theta,\G_n)$, the innovations $\W_t$ are centered and $\Svec$ has mean $h\U$, with $\Svec=\ind_E\Tvec_h+h\U$. If $\E\norm{\B_n-\bb_\theta}^4\le C_B/n^2$ and $\E\norm{\B_n-\bb_\theta}^2\le\Ctwo/n$, there is a deterministic nondecreasing $\ui:[0,1]\to[0,\infty)$ with $\ui(q)\to0$ as $q\downarrow0$ such that, for every event $F$,
\[
 \E\left[\frac{\norm{\Svec}^2}{h}\ind_F\right]\le\ui(\Pp(F)),\qquad
 \E\norm{\Svec}^2\le C_Sh.
\]
The modulus depends only on $M_2,L,C_B$ and the UI2 tail modulus. $C_S$ depends only on $M_2,L,\Ctwo$. Neither depends on $n,h,E,F$.
\end{lemma}
\begin{proof}[Proof of Lemma~\ref{lem:ui}]

Write $\E_* =\E[\,\cdot\mid\theta,\G_n]$ and $\mathcal T_{n,h}=\{n+1,\ldots,n+h\}$. Conditional iid sampling and Lemma~\ref{lem:gradients} give
\begin{align*}
 \E_*\W_t&=0,\qquad
 \E_*\norm{\W_t}^2
 =\E_*\norm{\xiobs_{Z_t}(\B_n)}^2-\norm{\nabla R_\theta(\B_n)}^2\le M_2,\\
 \E_*\norm{\Tvec_h}^2
 &=\sum_{t=n+1}^{n+h}\E_*\norm{\W_t}^2
   +2\sum_{n<s<t\le n+h}\ip{\E_*\W_s}{\E_*\W_t}
 \le hM_2.
\end{align*}
Since $\nabla R_\theta(\bb_\theta)=0$, one has $\norm\U\le L\ind_E\norm{\B_n-\bb_\theta}$. The cross term vanishes under $\E_*$, so
\begin{align*}
 \E_*\norm\Svec^2
 &=\ind_E\E_*\norm{\Tvec_h}^2
    +2h\ind_E\ip{\E_*\Tvec_h}{\U}+h^2\norm\U^2\\
 &=\ind_E\E_*\norm{\Tvec_h}^2+h^2\norm\U^2,\\
 \E\norm\Svec^2
 &\le hM_2+h^2L^2\Ctwo/n
 \le h(M_2+L^2\Ctwo).
\end{align*}
Thus $C_S=M_2+L^2\Ctwo$ is admissible.

Truncate the centered innovations, control the bounded part by a fourth moment, and control the remainder by the UI2 tail. Fix $R\ge\max\{1,2\Hgrad\}$ and write
\[
 \begin{aligned}
 \V_t^R&=\W_t\ind_{\{\norm{\W_t}\le R\}}-\E_*[\W_t\ind_{\{\norm{\W_t}\le R\}}],\\
 \Qvec_t^R&=\W_t-\V_t^R
 =\W_t\ind_{\{\norm{\W_t}>R\}}-\E_*[\W_t\ind_{\{\norm{\W_t}>R\}}].
 \end{aligned}
\]
Write $\widetilde\V_h^R=h^{-1/2}\sum_{t=n+1}^{n+h}\V_t^R$ and $\widetilde\Qvec_h^R=h^{-1/2}\sum_{t=n+1}^{n+h}\Qvec_t^R$. 

\emph{Truncated innovations.} Put $J_t=\{\norm{\xiobs_{Z_t}(\B_n)}>R/2\}$. Then $\{\norm{\W_t}>R\}\subseteq J_t$ and
$\Pp(J_t\mid\theta,\G_n)\le4\tau(R/2)/R^2$, so
\begin{align*}
 \E_*[\norm{\W_t}^2\ind_{\{\norm{\W_t}>R\}}]
 &\le2\E_*[\norm{\xiobs_{Z_t}(\B_n)}^2\ind_{J_t}]
       +2\Hgrad^2\Pp(J_t\mid\theta,\G_n)\\
 &\le\left(2+\frac{8\Hgrad^2}{R^2}\right)\tau(R/2)\le4\tau(R/2).
\end{align*}
The pairs are conditionally independent over $t$, and both components are centered. Since $\norm{\V_t^R}\le2R$,
\begin{align*}
 \E_*\norm{\V_t^R}^2
 &=\E_*[\norm{\W_t}^2\ind_{\{\norm{\W_t}\le R\}}]
   -\norm{\E_*[\W_t\ind_{\{\norm{\W_t}\le R\}}]}^2\le M_2,\\
 \E_*\norm{\Qvec_t^R}^2
 &=\E_*[\norm{\W_t}^2\ind_{\{\norm{\W_t}>R\}}]
   -\norm{\E_*[\W_t\ind_{\{\norm{\W_t}>R\}}]}^2\le4\tau(R/2),\\
 \E_*\norm{\V_t^R}^4&\le4R^2\E_*\norm{\V_t^R}^2\le4R^2M_2.
\end{align*}
For indices $i,j,k,l\in\mathcal T_{n,h}$, define
$\gamma_{ijkl}=\ip{\V_i^R}{\V_j^R}\ip{\V_k^R}{\V_l^R}$ and
$\mathcal C_{-i}=\sigma(\theta,\G_n,\W_s:s\in\mathcal T_{n,h}\setminus\{i\})$.
If \(i\notin\{j,k,l\}\), then \(\E[\V_i^R\mid\mathcal C_{-i}]=\E_*\V_i^R=0\) and
\[
 \E[\gamma_{ijkl}\mid\mathcal C_{-i}]
 =\ip{\E[\V_i^R\mid\mathcal C_{-i}]}{\V_j^R}
   \ip{\V_k^R}{\V_l^R}=0.
\] Consequently, \(\E_*\gamma_{ijkl}=\E_*[\E[\gamma_{ijkl}\mid\mathcal C_{-i}]]=0\).
The same conditioning argument applies whenever any index occurs only once. Grouping the remaining multiplicities gives
\begin{align*}
 \E_*\norm{\sum_{t\in\mathcal T_{n,h}}\V_t^R}^4
 &=\sum_{i,j,k,l\in\mathcal T_{n,h}}\E_*\gamma_{ijkl}\\
 &=\sum_{r\in\mathcal T_{n,h}}\E_*\gamma_{rrrr}
   +\sum_{\substack{r,s\in\mathcal T_{n,h}\\r<s}}
     \E_*[\gamma_{rrss}+\gamma_{ssrr}+\gamma_{rsrs}
                   +\gamma_{rssr}+\gamma_{srrs}+\gamma_{srsr}]\\
 &=\sum_{t\in\mathcal T_{n,h}}\E_*\norm{\V_t^R}^4
   +2\sum_{\substack{s,t\in\mathcal T_{n,h}\\s<t}}
            \E_*\norm{\V_s^R}^2\E_*\norm{\V_t^R}^2\\
 &\hspace{1em}+4\sum_{\substack{s,t\in\mathcal T_{n,h}\\s<t}}
            \E_*[\ip{\V_s^R}{\V_t^R}^2]\\
 &\le4hR^2M_2+6\binom h2 M_2^2.
\end{align*}
The last step uses Cauchy--Schwarz for the inner products and conditional independence for $s\ne t$.
For the normalized sums defined above,
\[
\begin{aligned}
 \E\norm{\widetilde\V_h^R}^4
 &=h^{-2}\E\left[\E_*\norm{\sum_{t\in\mathcal T_{n,h}}\V_t^R}^4\right]\\
 &\le\frac{4R^2M_2}h+\frac{3(h-1)M_2^2}h
 \le K_R:=4R^2M_2+3M_2^2,\\
 \E\norm{\widetilde\Qvec_h^R}^2
 &=h^{-1}\E\left[\sum_{t\in\mathcal T_{n,h}}\E_*\norm{\Qvec_t^R}^2\right]
 \le4\tau(R/2).
\end{aligned}
\]
Since $h^{-1/2}\Tvec_h=\widetilde\V_h^R+\widetilde\Qvec_h^R$, H\"older's inequality yields
\[
 \begin{aligned}
 \E\left[\frac{\norm{\Tvec_h}^2}{h}\ind_F\right]
 &\le2\E[\norm{\widetilde\V_h^R}^2\ind_F]+2\E[\norm{\widetilde\Qvec_h^R}^2\ind_F]\\
 &\le2(\E\norm{\widetilde\V_h^R}^4)^{1/2}\Pp(F)^{1/2}
       +2\E\norm{\widetilde\Qvec_h^R}^2\\
 &\le2\sqrt{K_R}\sqrt{\Pp(F)}+8\tau(R/2).
 \end{aligned}
\]
The earlier second-moment bound also gives $\E[\norm{\Tvec_h}^2\ind_F]/h\le M_2$. Set
\[
 \trmod(q)=\min\left\{M_2,\inf_{R\ge\max\{1,2\Hgrad\}}
       \{2\sqrt{K_R}\sqrt q+8\tau(R/2)\}\right\}.
\]
It is nondecreasing. For each fixed $R$,
\[
 0\le\limsup_{q\downarrow0}\trmod(q)\le8\tau(R/2)
 \xrightarrow[R\to\infty]{}0.
\]
Thus the modulus vanishes uniformly over the frozen windows.
It remains to restore the conditional mean $h\U$. H\"older's inequality and $h\le n$ give
\begin{align*}
 h^2\E\norm\U^4
 &\le L^4h^2\E\norm{\B_n-\bb_\theta}^4
 \le L^4C_B(h/n)^2,\\
 h\E[\norm\U^2\ind_F]
 &\le h(\E\norm\U^4)^{1/2}\Pp(F)^{1/2}
 \le L^2\sqrt{C_B}\sqrt{\Pp(F)}.
\end{align*}
Combining the two parts of $\Svec=\ind_E\Tvec_h+h\U$,
\begin{align*}
 \E\left[\frac{\norm\Svec^2}{h}\ind_F\right]
 &\le2\E\left[\frac{\norm{\Tvec_h}^2}{h}\ind_F\right]
       +2h\E[\norm\U^2\ind_F]\\
 &\le2\trmod(\Pp(F))+2L^2\sqrt{C_B}\sqrt{\Pp(F)}.
\end{align*}
Thus $\ui(q):=2\trmod(q)+2L^2\sqrt{C_B}\sqrt q$ has the required dependencies and tends to zero. 
\end{proof}

\section{Local window charges and prefix density}\label{app:converse}
We verify the deterministic endpoints used in Section~\ref{sec:proof-converse}, then prove the prefix-density identity. We use $Q_j,W_{c,j}$ and $e_n=e_n(A)$ from that section. The constants $a_w,\eta,N_*$ come from Proposition~\ref{prop:motion}, with $\chi=1/2$ and $C_w=96/a_w^2$.

\begin{lemma}[Window-to-charge inequality]\label{lem:window-charge}
For every admissible policy and $c\in\CC$, if $N=2^j\ge N_*$ and $1\le M\le\lfloor N/6\rfloor$ is an integer, then
\begin{equation}\label{eq:window-charge}
 \E W_{c,j}\ge(\chi\eta M-C_wM^2Q_j)
 c\left(\frac{a_w}{2\sqrt M}N^{-3/2}\right).
\end{equation}
No profile or lifetime-cost assumption is needed. The prefactor is positive exactly when $MQ_j<\chi\eta/C_w$.
\end{lemma}
\begin{proof}
For $1\le \ell\le M$, let $b=\lfloor N/(3M)\rfloor$, $u_\ell=N+3(\ell-1)b$, and
$I_\ell=\{u_\ell,\ldots,u_\ell+b-1\}$, $J_\ell=\{u_\ell+2b,\ldots,u_\ell+3b-1\}$. Choose
$n_\ell=\min\argmin_{n\in I_\ell}e_n$ and $m_\ell=\min\argmin_{n\in J_\ell}e_n$. The minimum index resolves ties. Since $N/(3M)\ge2$,
\begin{align*}
 b&=\lfloor N/(3M)\rfloor\ge N/(6M)\ge1,\\
 m_\ell-n_\ell&\ge(u_\ell+2b)-(u_\ell+b-1)=b+1\ge N/(6M),\\
 m_M&\le N+3Mb-1\le2N-1,\\
 m_\ell&\le u_\ell+3b-1=u_{\ell+1}-1<n_{\ell+1}\qquad(1\le \ell<M).
\end{align*}
Each endpoint minimum is at most its block average:
\[
 e_{n_\ell}\le b^{-1}\sum_{n\in I_\ell}e_n,\qquad
 e_{m_\ell}\le b^{-1}\sum_{n\in J_\ell}e_n,
\]
so
\[
 \sum_{\ell=1}^M(e_{n_\ell}+e_{m_\ell})
 \le b^{-1}\sum_{\ell=1}^M\sum_{n\in I_\ell\cup J_\ell}e_n
 \le\frac{6M}NQ_j.
\]
The endpoint estimates above supply the inputs to the Markov and path-length calculations in Section~\ref{sec:proof-converse}. Substitution in \eqref{eq:window-sum-main}, with $N=2^j$, proves \eqref{eq:window-charge}.
\end{proof}

The next lemma records the prefix-density argument from Section~\ref{sec:proof-converse} in the form used in Appendix~\ref{app:allocation}.

\begin{lemma}[Finite prefix-density summation]\label{lem:density}
Let $(u_i)_{i\ge1}$ be nonnegative and nonincreasing. If $S\subset\N$ satisfies
$N_S(m)=\#(S\cap\{1,\ldots,m\})\ge\lambda m-D_0$ for all $m$, with $\lambda>0$ and $0\le D_0<\infty$, then
\[
 \sum_{\substack{i\le m\\i\in S}}u_i\ge\lambda\sum_{i=1}^mu_i-D_0u_1.
\]
Summability on $S$ therefore implies full summability. The statement also holds for indexing from zero with the corresponding prefix convention.
\end{lemma}
\begin{proof}
Since $u_i-u_{i+1}\ge0$, finite summation by parts yields
\begin{align*}
 \sum_{\substack{i\le m\\i\in S}}u_i
 &=N_S(m)u_m+\sum_{i<m}N_S(i)(u_i-u_{i+1})\\
 &\ge\lambda\left(mu_m+\sum_{i<m}i(u_i-u_{i+1})\right)
       -D_0\left(u_m+\sum_{i<m}(u_i-u_{i+1})\right)\\
 &=\lambda\sum_{i=1}^mu_i-D_0u_1.
\end{align*}
If $\sum_{i\in S}u_i<\infty$, then
\[
 \sum_{i\ge1}u_i=\sup_m\sum_{i=1}^mu_i
 \le\lambda^{-1}\left(\sum_{i\in S}u_i+D_0u_1\right)<\infty.
\qedhere
\]
\end{proof}

\Needspace{10\baselineskip}
\section{Integer mesh and lifetime-cost accounting}\label{app:full-proof}\label{app:full-proof-attainment}
We verify the rounding and charge accounting for the schedules in Section~\ref{sec:proof-attainment} and Appendix~\ref{app:revisions}.

The mesh \eqref{eq:mesh} has $1\le k_j\le2^j$. Put $N=2^j$ and $\{x\}=x-\lfloor x\rfloor$. The endpoints are $t_{j,0}=N$ and $t_{j,k_j}=2N$. For $0\le\ell<k_j$,
\[
\begin{aligned}
t_{j,\ell+1}-t_{j,\ell}
&=\left\lfloor\frac{(\ell+1)N}{k_j}\right\rfloor
 -\left\lfloor\frac{\ell N}{k_j}\right\rfloor\\
&=\left\lfloor\frac N{k_j}+\left\{\frac{\ell N}{k_j}\right\}\right\rfloor
\in\left\{\left\lfloor\frac N{k_j}\right\rfloor,
 \left\lceil\frac N{k_j}\right\rceil\right\}.
\end{aligned}
\]
Thus $1\le t_{j,\ell+1}-t_{j,\ell}\le\lceil N/k_j\rceil$. The terminal endpoint $t_{j,k_j}=2^{j+1}=t_{j+1,0}$ is a single checkpoint. The policy issues $\B_{t_{j,k_j}}$ there, with exit charge $c(\norm{\B_{t_{j,k_j}}-\B_{t_{j,k_j-1}}})$ assigned to epoch $j$. The first checkpoint at $16$ has charge $c(\norm{\B_{16}-\B_0})$.

The excess estimate \eqref{eq:mesh-excess} holds at every dyadic horizon, including a horizon that ends after only part of a four-epoch allocation group. Indeed, let $m_J=\lfloor(J-1)/4\rfloor$ for $J\ge5$, so $1\le J-4m_J\le4$. The allocation contribution is exactly
\[
\begin{aligned}
\sum_{j=4}^{J-1}a_{\lfloor j/4\rfloor}
&=4\sum_{i=1}^{m_J-1}a_i+(J-4m_J)a_{m_J}\\
&\le4\sum_{i=1}^{m_J}a_i
\le4G(m_J)\le4G(J).
\end{aligned}
\]
The remaining $2^{-j}$ and $(j+1)^{-2}$ terms in \eqref{eq:mesh-excess} are summable. A larger constant covers $J\le4$.

Let $\Delta_{j,\ell}=\norm{\B_{t_{j,\ell+1}}-\B_{t_{j,\ell}}}$ and $H_{j,\ell}=\{\Delta_{j,\ell}>2^{-j/4}\}$ as in the main proof. For $J_{\max}\ge4$ put $T_{J_{\max}}=2^{J_{\max}+1}+1$. The additional one in $T_{J_{\max}}$ includes the final exit at count $2^{J_{\max}+1}$. Therefore
\[
V_{c,T_{J_{\max}}}(A)
=c(\norm{\B_{16}-\B_0})+
 \sum_{j=4}^{J_{\max}}\sum_{\ell=0}^{k_j-1}c(\Delta_{j,\ell}).
\]
Splitting these jumps at $2^{-j/4}$ gives \eqref{eq:cost-decomposition}.
The finite-sum moment calculation and the limit are given in Section~\ref{sec:proof-attainment}, ending in \eqref{eq:all-moments}.

\section{Infinite allocation and duality}\label{app:allocation}
Here \eqref{eq:allocation} and \eqref{eq:dual} also define $\Alloc((d_i),G)$, $\Alloc_m((d_i),G)$ and $\Dual((d_i),G)$ for arbitrary finite weights $d_i\ge0$ and any nondecreasing sequence $0<G(m)<\infty$. Sublinearity and $G\ge1$ are not needed for compactness or duality. Monotonicity of $d_i$ is used only when explicitly stated. For $d_i=c(2^{-i})$, this notation agrees with the main text.

\subsection{Summable charge weights}\label{app:summable}
The finite prefix-density inequality is Lemma~\ref{lem:density}, proved in Appendix~\ref{app:converse}. We apply it to the small-allowance coordinates.

\begin{lemma}[Sublinear allocation implies summable charges]\label{lem:summable}
If $d_i$ is nonnegative and nonincreasing, $a_i\ge0$, $\sum_{i\le m}a_i=o(m)$ and $\sum_i\phi_i(a_i)<\infty$, then $\sum_i d_i<\infty$.
\end{lemma}
\begin{proof}
Let $S=\{i:a_i\le1\}$, $N_S(m)=\#(S\cap\{1,\ldots,m\})$, and choose $m_0$ with $\sum_{i\le m}a_i\le m/2$ for $m\ge m_0$. Then
\begin{gather*}
 m-N_S(m)=\sum_{i\le m}\ind_{\{a_i>1\}}\le\sum_{i\le m}a_i\le m/2
 \quad(m\ge m_0),\\
 N_S(m)\ge m/2-m_0\quad(m\ge1),\qquad
 d_i\le\phi_i(a_i)\quad(i\in S),
\end{gather*}
where finite cost forces $d_i=0$ when $a_i=0$. Lemma~\ref{lem:density} with $(\lambda,D_0)=(1/2,m_0)$ gives
\[
 \sum_{i\ge1}d_i\le2\sum_{i\in S}d_i+2m_0d_1
 \le2\sum_{i\ge1}\phi_i(a_i)+2m_0d_1<\infty.
\qedhere
\]
\end{proof}

\subsection{Attainment for the infinite allocation program}\label{app:infinite}
\begin{lemma}[Infinite prefix attainment]\label{lem:infinite}
For positive finite-valued nondecreasing $G$ and finite $d_i\ge0$,
$\sup_m\Alloc_m((d_i),G)=\Alloc((d_i),G)$. If finite, this value is attained by one infinite prefix-feasible allocation.
\end{lemma}
\begin{proof}
The finite feasible set
\[
 \feasible_m(G)=
 \left\{(a_1,\ldots,a_m)\in[0,\infty)^m:
          \sum_{i=1}^{p}a_i\le G(p)\ (1\le p\le m)\right\}
\]
is closed and contained in $\prod_{i=1}^m[0,G(i)]$.
The positive point $a_i^\circ=G(1)/(2m)$ satisfies
\[
 \sum_{i=1}^{p}a_i^\circ=\frac{pG(1)}{2m}\le G(1)/2<G(p)\quad(p\le m).
\]
Each \(\phi_i\) is lower semicontinuous, with \(\phi_i(0)=0\) if \(d_i=0\) and \(\phi_i(x)\uparrow\infty\) as \(x\downarrow0\) if \(d_i>0\). Compactness and finiteness at \(\allocvec^\circ\) give a minimizer \(\allocvec^{(m)}=(a_1^{(m)},\ldots,a_m^{(m)})\) of finite value. Truncation gives \(\Alloc_m\le\Alloc_{m+1}\) and \(C:=\sup_m\Alloc_m\le\Alloc((d_i),G)\).
If $C=\infty$, equality follows. For finite $C$, the coordinate bounds $a_i^{(m)}\in[0,G(i)]$ give a diagonal subsequence $m_\ell\ge\ell$ such that $a_i^{(m_\ell)}\to a_i$ for every fixed $i$. For fixed integers $p,q\ge1$, take $\ell$ large enough that $m_\ell\ge\max\{p,q\}$. Lower semicontinuity of the coordinate objectives then gives
\begin{align*}
 \sum_{i=1}^{p}a_i&=\lim_{\ell\to\infty}\sum_{i=1}^{p}a_i^{(m_\ell)}\le G(p),\\
 \sum_{i=1}^q\phi_i(a_i)
 &\le\sum_{i=1}^q\liminf_{\ell\to\infty}\phi_i(a_i^{(m_\ell)})\\
 &\le\liminf_{\ell\to\infty}\sum_{i=1}^q\phi_i(a_i^{(m_\ell)})
 \le\liminf_{\ell\to\infty}\Alloc_{m_\ell}\le C.
\end{align*}
Taking the supremum over $q$ yields
\[
 C\le\Alloc((d_i),G)\le\sum_{i\ge1}\phi_i(a_i)
 =\sup_q\sum_{i=1}^q\phi_i(a_i)\le C.
\qedhere
\]
\end{proof}

\subsection{Decreasing-weight duality}\label{app:dual}
\begin{lemma}[Finite duality and infinite supremum]\label{lem:duality}
For the same weights and $G$, $\Alloc_m((d_i),G)$ equals the quotient in \eqref{eq:dual} optimized over its length-$m$ weights. Hence $\Alloc((d_i),G)=\Dual((d_i),G)$, including infinite values.
\end{lemma}
\begin{proof}
Equation~\eqref{eq:weak-duality} gives weak duality for each finite prefix, also for the arbitrary nonnegative weights of this appendix. To prove equality, attach a nonnegative multiplier $\lambda_p$ to each prefix constraint. Write $\bm{\lambda}=(\lambda_1,\ldots,\lambda_m)$ and set
$\omega_i=\sum_{p=i}^m\lambda_p$, $\omega_{m+1}=0$. Then $\lambda_i=\omega_i-\omega_{i+1}\ge0$ and
\begin{align*}
 \sum_{p=1}^m\lambda_p\sum_{i=1}^{p}a_i
 &=\sum_{p=1}^m\sum_{i=1}^p\lambda_p a_i
 =\sum_{i=1}^ma_i\sum_{p=i}^m\lambda_p
 =\sum_{i=1}^m\omega_i a_i,\\
 \mathcal L(\allocvec,\bm{\lambda})
 &=\sum_{i=1}^m\phi_i(a_i)
    +\sum_{p=1}^m\lambda_p\left(\sum_{i=1}^{p}a_i-G(p)\right)\\
 &=\sum_{i=1}^m\{\phi_i(a_i)+\omega_ia_i\}
    -\sum_{i=1}^m(\omega_i-\omega_{i+1})G(i).
\end{align*}
For \(d,\omega,x>0\), we have
\[
 d/x+\omega x-2\sqrt{d\omega}
 =(\sqrt{d/x}-\sqrt{\omega x})^2\ge0,
\]
with equality at \(x^*=\sqrt{d/\omega}\). Hence \(\inf_{x>0}\{d/x+\omega x\}=2\sqrt{d\omega}\). Write \(\phi_d\) for the extended ratio with numerator \(d\). If \(d=0<\omega\), the infimum is zero at \(x=0\). If \(\omega=0<d\), the infimum is \(\lim_{x\to\infty}d/x=0\) and has no finite minimizer. If \(d=\omega=0\), every \(x\ge0\) minimizes. Thus \(\inf_{x\ge0}\{\phi_d(x)+\omega x\}=2\sqrt{d\omega}\) for \(d,\omega\ge0\).
The finite product domain permits coordinatewise minimization:
\begin{align*}
 \inf_{a_i\ge0}\mathcal L(\allocvec,\bm{\lambda})
 &=\sum_{i=1}^m\inf_{a_i\ge0}\{\phi_i(a_i)+\omega_ia_i\}
    -\sum_{i=1}^m(\omega_i-\omega_{i+1})G(i)\\
 &=2\sum_{i=1}^m\sqrt{d_i\omega_i}-\sum_{i=1}^m(\omega_i-\omega_{i+1})G(i).
\end{align*}
The objective is proper and convex: $\phi_i''(x)=2d_i/x^3\ge0$ for $x>0$, and its effective domain is $\prod_iD_i$, where $D_i=(0,\infty)$ if $d_i>0$ and $D_i=[0,\infty)$ otherwise. The point $a_i^\circ=G(1)/(2m)$ lies in $\ri\prod_iD_i$ and satisfies
\[
 G(p)-\sum_{i=1}^{p}a_i^\circ
 \ge G(1)-\frac{pG(1)}{2m}\ge G(1)/2>0.
\]
Finite Slater duality \citep[Section 5.2.3]{boyd2004} yields
\[
 \Alloc_m=\sup_{\omega_1\ge\cdots\ge \omega_m\ge0}
 \left\{2\sum_{i=1}^m\sqrt{d_i\omega_i}-\sum_{i=1}^m(\omega_i-\omega_{i+1})G(i)\right\}.
\]
For nonzero weights define $S_\omega=\sum_{i=1}^m\sqrt{d_i\omega_i}$ and
$D_\omega=\sum_{i=1}^m(\omega_i-\omega_{i+1})G(i)\ge G(1)\omega_1>0$. Optimizing their common scale gives
\begin{align*}
 \sup_{t\ge0}\{2\sqrt t\,S_\omega-tD_\omega\}
 &=\sup_{t\ge0}\left\{\frac{S_\omega^2}{D_\omega}
                 -D_\omega\left(\sqrt t-\frac{S_\omega}{D_\omega}\right)^2\right\}
 =\frac{S_\omega^2}{D_\omega},\\
 \Alloc_m&=\sup_{\substack{\omega_1\ge\cdots\ge \omega_m\ge0\\\omega_1>0}}
                      \frac{S_\omega^2}{D_\omega}.
\end{align*}
The maximizing scale $t=(S_\omega/D_\omega)^2$ also covers $S_\omega=0$.
Lemma~\ref{lem:infinite} now gives
\[
 \Alloc((d_i),G)=\sup_m\Alloc_m((d_i),G)=\sup_{m,\omega}\frac{S_\omega^2}{D_\omega}=\Dual((d_i),G).
\qedhere
\]
\end{proof}
For $t>0$, the coordinatewise map $(a_i)\mapsto(ta_i)$ takes allocations for $G$ bijectively to allocations for $tG$. Thus $\Alloc((d_i),tG)=t^{-1}\Alloc((d_i),G)$, including infinite values. The generalized domain includes $tG$ even if $tG(1)<1$. This scaling absorbs the multiplicative excess coefficient, whereas Section~\ref{sec:proof-converse} handles the index shift separately.

Constant weights give the necessary test
\begin{equation}\label{eq:constant-test}
 \sup_m\frac{\big(\sum_{i\le m}\sqrt{d_i}\big)^2}{G(m)}<\infty.
\end{equation}
To see its insufficiency, take $G(m)=m^{1/2}$ and
\[
 c(x)=\begin{cases}
 [\log_2(1/x)]^{-3/2},&0<x\le1/2,\\
 1,&1/2<x\le1,\\
 0,&x=0.
 \end{cases}
\]
Then $d_i=i^{-3/2}$ and $\sum_{i\le m}i^{-3/4}\le Cm^{1/4}$, so \eqref{eq:constant-test} is bounded. But
$\sum_i i^{1-1/2}d_i=\sum_i i^{-1}=\infty$. Corollary~\ref{cor:power} gives $\Alloc(c,G)=\infty$. Decreasing weights detect divergence accumulated over disjoint scales that a single terminal-prefix test misses.

\section{Proofs of the profile and revision consequences}\label{app:profiles}
We prove the endpoint and power-profile criteria first, followed by the revision bounds and critical logarithmic corrections. Distance coordinates use $i,m$, observation epochs use $j,J$, and $k$ indexes dyadic blocks of either sequence. In the revision proof, $s_j$ denotes the expected number of actual internal revisions.

\subsection{The integrability endpoints}\label{app:endpoints}
\begin{proof}[Proof of Corollary~\ref{cor:dini}]
For $G\equiv1$, the prefix constraints are equivalent to $\sum_i a_i\le1$. Put $S=\sum_i\sqrt{d_i}$. For any finite-cost feasible allocation, Cauchy--Schwarz gives
\begin{align*}
 \left(\sum_{i=1}^m\sqrt{d_i}\right)^2
 &\le\left(\sum_{i=1}^ma_i\right)
       \left(\sum_{i=1}^m\phi_i(a_i)\right)\\
 &\le\sum_{i=1}^m\phi_i(a_i)
 \le\sum_{i\ge1}\phi_i(a_i).
\end{align*}
When $a_i=0$, finite cost forces $d_i=0$, so that coordinate contributes zero. Letting $m\to\infty$ gives $S^2\le\sum_i\phi_i(a_i)$. This lower bound is automatic for an infinite-cost allocation. If $0<S<\infty$, take $a_i=\sqrt{d_i}/S$, which has total allowance one and cost $S^2$. If $S=0$, take the zero allocation. If $S=\infty$, the lower bound rules out every finite-cost allocation. Thus
\[
 \Alloc(c,1)=\left(\sum_{i\ge1}\sqrt{c(2^{-i})}\right)^2.
\]
For any nonnegative nondecreasing $a$, integral comparison gives
\[
 (\log2)\sum_{i\ge1}a(2^{-i})
 \le\int_0^1a(x)\frac{\dd x}{x}
 \le(\log2)\sum_{i\ge0}a(2^{-i}).
\]
Apply this with $a=\sqrt c$ for the bounded-excess equivalence.

If a finite-expected-cost policy has $\EE_T=o(\log T)$, then
$G(m)=\max\{1,\EE_{2^m}(A)\}$ belongs to $\HH$. Theorem~\ref{thm:main} and Lemma~\ref{lem:summable} give $\sum_i d_i<\infty$, hence $\DiniOne(c)<\infty$.
Conversely, if $\sum_i d_i<\infty$, choose strictly increasing integers $n_\ell\ge\ell$ with
$\sum_{i\ge n_\ell}d_i\le2^{-\ell}$. Put
$k_i=1+\sum_{\ell\ge1}\ind_{\{n_\ell\le i\}}$, $a_i=k_i^{-1}$, and
$G(m)=\max\{1,\sum_{i=1}^ma_i\}$. Since $k_i\to\infty$, Ces\`aro convergence gives $G(m)=o(m)$. Tonelli's theorem gives
\begin{align*}
 \sum_i\phi_i(a_i)&=\sum_i k_id_i
 =\sum_i d_i+\sum_{\ell\ge1}\sum_{i\ge n_\ell}d_i\\
 &\le\sum_i d_i+\sum_{\ell\ge1}2^{-\ell}<\infty.
\end{align*}
Theorem~\ref{thm:main} supplies an attaining policy with all positive lifetime-cost moments.
\end{proof}

\subsection{The prescribed-power criterion}\label{app:power}
\begin{proof}[Proof of Corollary~\ref{cor:power}]
For sufficiency, take \(G(m)=m^r\) and its positive increments \(a_i=i^r-(i-1)^r\). Then
\[
\begin{aligned}
 a_i&=\int_{i-1}^i rx^{r-1}\,\dd x\ge ri^{r-1},\\
 \sum_{i=1}^m a_i&=m^r,\qquad
 \sum_{i\ge1}\phi_i(a_i)\le r^{-1}\sum_{i\ge1}i^{1-r}d_i.
\end{aligned}
\]
Theorem~\ref{thm:main} and \eqref{eq:all-horizons} yield the policy.

For necessity, let $(a_i)$ be a finite-cost prefix allocation and put
$I_k=\{2^k,\ldots,2^{k+1}-1\}$, $k\ge0$. Finite cost forces $a_i>0$ wherever $d_i>0$. Hence
\[
 \left(\sum_{i\in I_k}\sqrt{d_i}\right)^2
 =\left(\sum_{i\in I_k}\sqrt{\phi_i(a_i)}\sqrt{a_i}\right)^2
 \le\left(\sum_{i\in I_k}\phi_i(a_i)\right)
     \left(\sum_{i\in I_k}a_i\right).
\]
Zero-cost coordinates contribute zero, also at $a_i=0$. For a block of positive total allocation,
\begin{equation}\label{eq:power-block}
\begin{aligned}
 \sum_{i\in I_k}\phi_i(a_i)
 &\ge\frac{\big(\sum_{i\in I_k}\sqrt{d_i}\big)^2}{\sum_{i\in I_k}a_i}
 \ge\frac{2^{2k}d_{2^{k+1}}}{G(2^{k+1})}
 =2^{-r}2^{(2-r)k}d_{2^{k+1}}.
\end{aligned}
\end{equation}
A zero-allocation block has all $d_i=0$, and its lower bound is $0\ge0$.
For $k\ge1$, the preceding block estimate gives
\[
 \sum_{i\in I_{k-1}}\phi_i(a_i)
 \ge 2^{-r}2^{(2-r)(k-1)}d_{2^k}
 =2^{(2-r)k-2}d_{2^k}.
\]
There are $2^k$ indices in $I_k$, and both $i^{1-r}\le(2^{k+1})^{1-r}$ and $d_i\le d_{2^k}$ there. Therefore
\[
 \begin{aligned}
 \sum_{i\in I_k}i^{1-r}d_i
 &\le2^k(2^{k+1})^{1-r}d_{2^k}\\
 &=2^{(2-r)k+1-r}d_{2^k}\\
 &\le2^{3-r}\sum_{i\in I_{k-1}}\phi_i(a_i).
 \end{aligned}
\] Summing disjoint blocks yields
\[
 \sum_{i\ge1}i^{1-r}d_i
 \le d_1+2^{3-r}\sum_{k\ge1}\sum_{i\in I_{k-1}}\phi_i(a_i)
 =d_1+2^{3-r}\sum_{i\ge1}\phi_i(a_i)<\infty.
\]
For $2^{-(i+1)}\le x\le2^{-i}$, monotonicity gives
\begin{align*}
 (\log2)(1+i\log2)^{1-r}d_{i+1}
 &\le\int_{2^{-(i+1)}}^{2^{-i}}c(x)[\log(e/x)]^{1-r}\frac{\dd x}{x}\\
 &\le(\log2)(1+(i+1)\log2)^{1-r}d_i.
\end{align*}
Summing and setting $j=i+1$ gives
\begin{align*}
 \frac{(\log2)^{2-r}}{2^{1-r}}\sum_{j\ge2}j^{1-r}d_j
 &\le(\log2)^{2-r}\sum_{i\ge1}i^{1-r}d_{i+1}\\
 &\le\int_0^1c(x)[\log(e/x)]^{1-r}\frac{\dd x}{x}\\
 &\le(\log2)(1+\log2)^{1-r}c(1)
      +(\log2)(1+2\log2)^{1-r}\sum_{i\ge1}i^{1-r}d_i.
\end{align*}
Since \(d_1<\infty\), this proves \eqref{eq:weighted-dini}. For \(c_\alpha\), we have \(d_i=(1+i\log2)^{-\alpha}\) and \((\log2)i\le1+i\log2\le(1+\log2)i\), hence \((1+\log2)^{-\alpha}i^{1-r-\alpha}\le i^{1-r}d_i\le(\log2)^{-\alpha}i^{1-r-\alpha}\).
The series converges exactly when $\alpha+r>2$.
\end{proof}

\subsection{Revision complexity}\label{app:revisions}
\begin{proof}[Proof of Theorem~\ref{thm:revisions}]
Recall $Q_j=\sum_{n=2^j}^{2^{j+1}-1}\E\norm{\A_n-\B_n}^2$.
The constants $a_w,\eta,N_*$ come from Proposition~\ref{prop:motion}, while
$\chi=1/2$, $C_w=96/a_w^2$ and $C_w\alpha_*+\lambda_0<\chi\eta$ are fixed in Section~\ref{sec:proof-converse}. Choose $j_0$ there so $2^j\ge\max\{N_*,12\}$ for $j\ge j_0$. Define
\[
 S_j^{\mathrm{int}}=\sum_{n=2^j+1}^{2^{j+1}-1}\ind_{\{\A_n\ne\A_{n-1}\}},
 \qquad s_j=\E S_j^{\mathrm{int}}.
\]
These internal counts exclude entry edges, so $\sum_{j<J}s_j\le\E N_A(2^J)$. Applying Lemma~\ref{lem:window-charge} to $c_{\mathrm{sw}}(x)=\ind_{\{x>0\}}$ gives
\begin{equation}\label{eq:revision-window}
 s_j\ge\chi\eta M-C_wM^2Q_j,\qquad1\le M\le\lfloor2^j/6\rfloor,
 \quad j\ge j_0.
\end{equation}
The argument of $c_{\mathrm{sw}}$ in \eqref{eq:window-charge} is strictly positive, so this charge is exactly one there. The bound \eqref{eq:revision-window} therefore counts actual revisions and uses no lifetime-cost assumption. For $Q_j\le\alpha_*/2$, choose
$M_j=\lfloor\min\{\alpha_*/Q_j,2^j/6\}\rfloor$, with $\alpha_*/0=\infty$.
The argument inside the floor is at least two, and \eqref{eq:rounding} gives $M_j^{-1}\le C(Q_j+2^{-j})$. Thus, for a fixed $K_{\mathrm{rev}}>0$,
\begin{equation}\label{eq:revision-local}
 s_j\ge\lambda_0M_j\ge\frac{K_{\mathrm{rev}}}{Q_j+2^{-j}}
 \quad\text{if }j\ge j_0,\quad Q_j\le\delta_{\mathrm{rev}}:=\alpha_*/2.
\end{equation}

\emph{Part (a): an arbitrary sublinear profile.} Quadratic growth gives $\sum_{j<J}Q_j\le(C_A/\kappa)G(J)$ eventually. For $\lceil J/2\rceil\ge j_0$, put
$I_J=\{j:\lceil J/2\rceil\le j<J,\ Q_j\le\delta_{\mathrm{rev}}\}$. Then
\begin{align*}
 |I_J|&\ge\lfloor J/2\rfloor-\frac2{\alpha_*}\sum_{j<J}Q_j
 \ge\lfloor J/2\rfloor-\frac{2C_A}{\alpha_*\kappa}G(J)\ge J/3,\\
 \sum_{j\in I_J}(Q_j+2^{-j})
 &\le\frac{C_A}\kappa G(J)+2^{1-\lceil J/2\rceil}
 \le\left(\frac{C_A}\kappa+2\right)G(J).
\end{align*}
The cardinality bound holds eventually because $G(J)=o(J)$. Cauchy--Schwarz now gives
\begin{align*}
 \E N_A(2^J)&\ge\sum_{j\in I_J}s_j
 \ge K_{\mathrm{rev}}\frac{|I_J|^2}{\sum_{j\in I_J}(Q_j+2^{-j})}\\
 &\ge\frac{K_{\mathrm{rev}}}{9(C_A/\kappa+2)}\frac{J^2}{G(J)}.
\end{align*}
Take $K_A=K_{\mathrm{rev}}/[9(C_A/\kappa+2)]$. The threshold beyond which this bound holds may depend on the policy and profile.

\emph{Part (b): power profiles.} For the lower bound, use $G(J)=J^r$ and $J=\lfloor\log_2T\rfloor$. At large $T$, $2^J\le T$ and $J\ge\log T/(2\log2)$, so
\[
 \E N_A(T)\ge\E N_A(2^J)\ge K_AJ^{2-r}
 \ge\frac{K_A}{(2\log2)^{2-r}}(\log T)^{2-r}.
\]
For the upper bound, retain $\B_0$ before count $16$ and use \eqref{eq:mesh} with
$k_j=\min\{2^j,\lceil(j+1)^{1-r}\rceil\}$ for $j\ge4$. This schedule is fixed before choosing $c$. Since $j+1\le2^j$,
\[
 k_j=\lceil(j+1)^{1-r}\rceil,\qquad
 (j+1)^{1-r}\le k_j\le\min\{j+1,1+(j+1)^{1-r}\}.
\]
For $J\ge5$,
\begin{align*}
 \sum_{j=4}^{J-1}k_j^{-1}
 &\le\sum_{j=4}^{J-1}(j+1)^{r-1}
 \le\int_4^Jx^{r-1}\,\dd x=\frac{J^r-4^r}r,\\
 \EE_{2^J}(A)
 &\le C+C_0\sum_{j=4}^{J-1}(k_j^{-1}+2^{-j})
 \le C+C_0\left\{\frac{J^r-4^r}r+\frac18\right\}=O(J^r).
\end{align*}
For \(J_T=\lfloor\log_2T\rfloor\ge4\), monotonicity extends the excess bound to every horizon, giving \(\EE_T(A)\le\EE_{2^{J_T+1}}(A)\lesssim(J_T+1)^r\asymp(\log T)^r\).
The increasing function $x^{1-r}$ gives the pathwise count
\begin{align*}
 N_A(T)&\le1+\sum_{j=4}^{J_T}k_j
 \le1+(J_T-3)+\sum_{j=4}^{J_T}(j+1)^{1-r}\\
 &\le J_T-2+\int_5^{J_T+2}x^{1-r}\,\dd x\\
 &=J_T-2+\frac{(J_T+2)^{2-r}-5^{2-r}}{2-r}
 =O((\log T)^{2-r}).
\end{align*}
A partial epoch is bounded by the full epoch. For $j=4i,\ldots,4i+3$,
$c(2^{-j/4})\le d_i$ and $k_j\le C_ri^{1-r}$. Consequently
\[
 \sum_{j\ge4}k_jc(2^{-j/4})\le4C_r\sum_{i\ge1}i^{1-r}d_i<\infty.
\]
The displacement cutoff remains $2^{-j/4}$. The risk-energy estimate in \eqref{eq:exception-prob} applies to every pair of adjacent checkpoints on this new mesh, so the exceptional-event bound is unchanged. Since $k_j\le j+1$, the series $\sum_j k_j2^{-j/(2q)}$ converges for each fixed $q\ge1$. Together with the ordinary-charge bound above, the same finite-sum Minkowski and monotone-convergence steps in \eqref{eq:all-moments} give every positive cost moment. Neither the mesh nor its actions depend on the choice of charge.

\emph{Part (c): bounded total excess.} For necessity, finite total excess gives $\sum_jQ_j<\infty$. Choose $j_1\ge j_0$ so $Q_j\le\alpha_*/2$ for all $j\ge j_1$. Every such epoch satisfies \eqref{eq:revision-local}, whence
\begin{equation}\label{eq:inverse-revisions}
 \sum_{j\ge j_1}s_j^{-1}\le C\sum_{j\ge j_1}(Q_j+2^{-j})<\infty.
\end{equation}
Here $s_j$ is positive and finite. If $\E N_A(2^J)\le C_NJ^2(\log J)^\beta$ eventually, set
$K_\beta=\{4C_N(\log2)^\beta\}^{-1}$. Define the epoch-index blocks
$\mathcal J_k=\{2^k,\ldots,2^{k+1}-1\}$, in contrast to the distance-index blocks
$I_k$ used in \eqref{eq:power-block}. For large $k$,
\begin{align*}
 \sum_{j\in\mathcal J_k}s_j^{-1}
 &\ge\frac{2^{2k}}{\sum_{j\in\mathcal J_k}s_j}
 \ge\frac{2^{2k}}{\E N_A(2^{2^{k+1}})}\\
 &\ge\frac{2^{2k}}{C_N2^{2(k+1)}\{(k+1)\log2\}^\beta}
 =K_\beta(k+1)^{-\beta}.
\end{align*}
Disjoint-block summation contradicts \eqref{eq:inverse-revisions} if $\beta\le1$. Corollary~\ref{cor:dini} also forces $\DiniHalf(c)<\infty$.

Conversely, let $\beta>1$ and $\DiniHalf(c)<\infty$. Take a finite-cost allocation for $G\equiv1$ and replace the polynomial checkpoint cap by
$k_j^{\mathrm{cap}}=\lceil(j+1)[\log(j+2)]^\beta\rceil$. Set
\[
 k_j=
 \begin{cases}
 \min\{2^j,k_j^{\mathrm{cap}},\lceil a_{\lfloor j/4\rfloor}^{-1}\rceil\},&a_{\lfloor j/4\rfloor}>0,\\
 \min\{2^j,k_j^{\mathrm{cap}}\},&a_{\lfloor j/4\rfloor}=0.
 \end{cases}
\]
Use \eqref{eq:mesh} with these counts, retaining $\B_0$ before count $16$. The cutoff and edge accounting are unchanged. The new cap satisfies
\[
 \sum_{j\ge4}(k_j^{\mathrm{cap}})^{-1}
 \le\sum_{m\ge5}\frac1{m(\log m)^\beta}
 \le\int_4^\infty\frac{\dd x}{x(\log x)^\beta}
 =\frac{(\log4)^{1-\beta}}{\beta-1}<\infty,
\]
\[
 \sum_{j\ge4}(k_j^{-1}+2^{-j})
 \le2\sum_{j\ge4}2^{-j}+\sum_{j\ge4}(k_j^{\mathrm{cap}})^{-1}
       +4\sum_{i\ge1}a_i<\infty.
\]
Equation~\eqref{eq:mesh-excess} gives finite total excess. For $i=\lfloor j/4\rfloor$ and $a_i>0$,
$k_jc(2^{-j/4})\le\lceil a_i^{-1}\rceil d_i\le d_i+\phi_i(a_i)$.
If $a_i=0$, then $d_i=0$. Hence
\[
 \sum_{j\ge4}k_jc(2^{-j/4})\le4\sum_i\{d_i+\phi_i(a_i)\}<\infty.
\]
Let $N_{J_{\max}}^{\mathrm{exc}}=\sum_{j=4}^{J_{\max}}\sum_{\ell<k_j}\ind_{H_{j,\ell}}$ and
$N_\infty^{\mathrm{exc}}=\lim_{J_{\max}}N_{J_{\max}}^{\mathrm{exc}}$, where $H_{j,\ell}=\{\norm{\B_{t_{j,\ell+1}}-\B_{t_{j,\ell}}}>2^{-j/4}\}$ is defined on this new mesh. By \eqref{eq:exception-prob}, $\Pp(H_{j,\ell})\le p_j:=(C_0/\kappa)2^{-j/2}$. For $q\ge1$,
\begin{align*}
 \sup_{J_{\max}}\norm{N_{J_{\max}}^{\mathrm{exc}}}_{L^q}
 &\le\sum_{j\ge4}k_j^{\mathrm{cap}}p_j^{1/q}\\
 &\le C^{1/q}\sum_{j\ge4}\{(j+1)[\log(j+2)]^\beta+1\}2^{-j/(2q)}.
\end{align*}
The summand $v_j$ satisfies
\[
 \frac{v_{j+1}}{v_j}
 =2^{-1/(2q)}\frac{(j+2)[\log(j+3)]^\beta+1}
                       {(j+1)[\log(j+2)]^\beta+1}
 \longrightarrow2^{-1/(2q)}<1.
\]
Minkowski and monotone convergence give all $q\ge1$ moments. Jensen covers $0<q<1$. Equation~\eqref{eq:cost-decomposition} transfers them to lifetime cost without independence of checkpoint edges. The pathwise revision count satisfies, with $J_T=\lfloor\log_2T\rfloor$ for sufficiently large $T$,
\begin{align*}
 N_A(T)&\le1+\sum_{j=4}^{J_T}k_j^{\mathrm{cap}}
 \le J_T-2+[\log(J_T+2)]^\beta\sum_{i=5}^{J_T+1}i\\
 &=J_T-2+[\log(J_T+2)]^\beta
              \left\{\frac{(J_T+1)(J_T+2)}2-10\right\}\\
 &\le2J_T^2[\log(J_T+2)]^\beta\\
 &\le\frac{2^{\beta+1}}{(\log2)^2}(\log T)^2(\log\log T)^\beta
 \qquad\text{for sufficiently large }T.
\end{align*}
This horizon-independent mesh may depend on $c$ and $\beta$.
\end{proof}

\Needspace{12\baselineskip}
\subsection{Critical logarithmic corrections}\label{app:critical}
\begin{lemma}[Tail allocation for a finite-cost policy]\label{lem:tail}
If $\DiniOne(c)<\infty$, every admissible policy with $\E V_{c,\infty}(A)<\infty$ has a positive deterministic sequence $(r_j^{\mathrm{ep}})$ satisfying, with $\delta_*=a_w/2$,
\[
 \sum_{j=0}^Jr_j^{\mathrm{ep}}\le C_1\EE_{2^{J+1}}(A)+C_2,
 \qquad \sum_{j\ge0}\frac{c(\delta_*2^{-2j})}{r_j^{\mathrm{ep}}}<\infty.
\]
These conclusions hold without an excess-profile assumption, with constants depending only on the model and finite initialization.
\end{lemma}
\begin{proof}
Recall $Q_j=\sum_{n=2^j}^{2^{j+1}-1}e_n$ and the fixed constants $\alpha_*,\lambda_0,j_0,\delta_*$ from Section~\ref{sec:proof-converse}. Use $\good=\{j\ge j_0:Q_j\le\alpha_*/2\}$, $M_j=\lfloor\min\{\alpha_*/Q_j,2^j/6\}\rfloor$ there, with $\alpha_*/0=\infty$, and set $r_j^{\mathrm{ep}}=M_j^{-1}$ on $\good$ and $r_j^{\mathrm{ep}}=1$ elsewhere. The Dini condition supplies the summability needed for the omitted epochs. Since $\delta_*\le1$, with $\dep_j=c(\delta_*2^{-2j})$,
\[
 \sum_{j\ge0}\dep_j\le c(1)+\sum_{j\ge1}c(2^{-j})
 \le c(1)+\frac{\DiniOne(c)}{\log2}<\infty.
\]
By \eqref{eq:good-cost},
\begin{align*}
 \sum_{j\ge0}\frac{\dep_j}{r_j^{\mathrm{ep}}}
 &=\sum_{j\in\good}M_j\dep_j+\sum_{j\notin\good}\dep_j\\
 &\le\lambda_0^{-1}\E V_{c,\infty}(A)+\sum_{j\ge0}\dep_j<\infty.
\end{align*}
Summing \eqref{eq:rounding} on good epochs and $1<2Q_j/\alpha_*$ on late bad epochs gives
\[
 \sum_{j=0}^Jr_j^{\mathrm{ep}}
 \le j_0+\frac2{\alpha_*}\sum_{j=0}^JQ_j+12\sum_{j\ge0}2^{-j}
 \le C_2+C_1\EE_{2^{J+1}}(A).
\]
Take $C_1=2/(\alpha_*\kappa)$ and $C_2=j_0+24$.
\end{proof}

\begin{proof}[Proof of Corollary~\ref{cor:critical}]
Use the fixed scale $\delta_*=a_w/2$ from Lemma~\ref{lem:tail}. Put $a=1-\log\delta_*>0$ and $b=2\log2>0$. Then
$\dep_j=c_\alpha(\delta_*2^{-2j})=(a+bj)^{-\alpha}$.
For the positive sequence from Lemma~\ref{lem:tail}, define
$S_{J_{\mathrm{tail}}}=\sum_{j\ge J_{\mathrm{tail}}}\dep_j/r_j^{\mathrm{ep}}$.
These positive tail sums decrease to zero. For a fixed cutoff $J_{\mathrm{tail}}$ and $J\ge J_{\mathrm{tail}}$, Cauchy--Schwarz gives
\begin{align*}
 \left(\sum_{j=J_{\mathrm{tail}}}^J\sqrt{\dep_j}\right)^2
 &=\left(\sum_{j=J_{\mathrm{tail}}}^J
              \sqrt{\dep_j/r_j^{\mathrm{ep}}}\sqrt{r_j^{\mathrm{ep}}}\right)^2\\
 &\le\left(\sum_{j=J_{\mathrm{tail}}}^J\frac{\dep_j}{r_j^{\mathrm{ep}}}\right)
        \left(\sum_{j=J_{\mathrm{tail}}}^Jr_j^{\mathrm{ep}}\right)\\
 &\le S_{J_{\mathrm{tail}}}\{C_1\EE_{2^{J+1}}(A)+C_2\}.
\end{align*}
The decreasing integrand gives
\[
 \int_{J_{\mathrm{tail}}}^{J+1}(a+bx)^{-\alpha/2}\,\dd x
 \le\sum_{j=J_{\mathrm{tail}}}^J\sqrt{\dep_j}
 \le(a+bJ_{\mathrm{tail}})^{-\alpha/2}
       +\int_{J_{\mathrm{tail}}}^J(a+bx)^{-\alpha/2}\,\dd x.
\]
For $0\le u<v$,
\[
 \int_u^v(a+bx)^{-\alpha/2}\,\dd x
 =\begin{cases}
 \displaystyle\frac{(a+bv)^{1-\alpha/2}-(a+bu)^{1-\alpha/2}}
                   {b(1-\alpha/2)},&1<\alpha<2,\\[6pt]
 \displaystyle b^{-1}\log\frac{a+bv}{a+bu},&\alpha=2.
 \end{cases}
\]
At a fixed tail cutoff,
\[
 \sum_{j=J_{\mathrm{tail}}}^J\sqrt{\dep_j}
 \sim\begin{cases}
 \displaystyle\frac{b^{-\alpha/2}}{1-\alpha/2}J^{1-\alpha/2},&1<\alpha<2,\\[5pt]
 b^{-1}\log J,&\alpha=2.
 \end{cases}
\]
Let $\Phi_\alpha(J)=J^{2-\alpha}$ for $1<\alpha<2$ and
$\Phi_2(J)=(\log J)^2$. The squared leading coefficient $A_\alpha>0$ depends only on $\alpha$. Rearranging before taking limits yields
\begin{align*}
 \frac{\EE_{2^{J+1}}(A)}{\Phi_\alpha(J)}
 &\ge\frac{\big(\sum_{j=J_{\mathrm{tail}}}^J\sqrt{\dep_j}\big)^2}
              {C_1S_{J_{\mathrm{tail}}}\Phi_\alpha(J)}
       -\frac{C_2}{C_1\Phi_\alpha(J)},\\
 \liminf_{J\to\infty}\frac{\EE_{2^{J+1}}(A)}{\Phi_\alpha(J)}
 &\ge\frac{A_\alpha}{C_1S_{J_{\mathrm{tail}}}},\\
 \liminf_{J\to\infty}\frac{\EE_{2^{J+1}}(A)}{\Phi_\alpha(J)}
 &\ge\sup_{J_{\mathrm{tail}}\ge0}\frac{A_\alpha}{C_1S_{J_{\mathrm{tail}}}}
 =\infty.
\end{align*}
This uses the order: first $J\to\infty$, then $J_{\mathrm{tail}}\to\infty$.
For $2^J\le T<2^{J+1}$, monotonicity gives respectively
\[
 \frac{\EE_T(A)}{(\log T)^{2-\alpha}}
 \ge\frac{\EE_{2^J}(A)}{\{(J+1)\log2\}^{2-\alpha}},\qquad
 \frac{\EE_T(A)}{(\log\log T)^2}
 \ge\frac{\EE_{2^J}(A)}{[\log\{(J+1)\log2\}]^2}.
\]
The denominator comparisons are
\begin{align*}
 \frac{\{(J+1)\log2\}^{2-\alpha}}{\Phi_\alpha(J-1)}
 &=\left(\frac{J+1}{J-1}\log2\right)^{2-\alpha}
 \longrightarrow(\log2)^{2-\alpha}>0,\\
 \frac{[\log\{(J+1)\log2\}]^2}{\Phi_2(J-1)}
 &=\left(\frac{\log(J+1)+\log\log2}{\log(J-1)}\right)^2\longrightarrow1.
\end{align*}
Both little-omega conclusions therefore hold at arbitrary horizons.

For \eqref{eq:critical-one}, write $r_*=2-\alpha\in(0,1)$ and
$H(x)=x^{r_*}(\log x)^\beta$ at large $x$. Since
\[
 H'(x)=x^{r_*-1}(\log x)^\beta\left(r_*+\frac\beta{\log x}\right)>0
\]
on a tail and $H(x)/x\to0$, choose $m_0\ge3$ with $H'>0$ beyond $m_0$ and $H(m_0)\ge1$. Then $G(m)=H(\max\{m,m_0\})$ lies in $\HH$. At $T=2^m$, the profile in \eqref{eq:critical-one} is $(m\log2)^{r_*}[\log(m\log2)]^\beta\asymp H(m)$ for every fixed real $\beta$. A policy attaining this profile at finite expected cost satisfies $\EE_{2^m}(A)=O(G(m))$ and yields a finite-cost allocation by Theorem~\ref{thm:main}. Recall the distance-index blocks $I_k=\{2^k,\ldots,2^{k+1}-1\}$ used in \eqref{eq:power-block}, and choose $k_0$ with $2^{k_0+1}\ge m_0$. For $k\ge k_0$,
\begin{gather*}
 d_{2^{k+1}}=(1+2^{k+1}\log2)^{-\alpha}
 \ge(1+\log2)^{-\alpha}2^{-\alpha(k+1)},\\
 G(2^{k+1})=2^{r_*(k+1)}\{(k+1)\log2\}^{\beta},\\
 \sum_{i\in I_k}\phi_i(a_i)
 \ge\frac{2^{2k}d_{2^{k+1}}}{G(2^{k+1})}
 \ge\frac{(k+1)^{-\beta}}{4(1+\log2)^\alpha(\log2)^\beta}.
\end{gather*}
For $\beta\le1$,
\[
 \sum_{i\ge1}\phi_i(a_i)
 \ge\sum_{k=k_0}^{N_{\mathrm{tail}}}\sum_{i\in I_k}\phi_i(a_i)
 \ge\frac{\sum_{k=k_0}^{N_{\mathrm{tail}}}(k+1)^{-\beta}}
              {4(1+\log2)^\alpha(\log2)^\beta}
 \longrightarrow\infty.
\]
For $\beta>1$, take $i_0\ge m_0+1$, set $\Delta_i=H(i)-H(i-1)$ for $i\ge i_0$, and choose positive earlier $\Delta_i$. For $x\in[i-1,i]$,
$x^{r_*-1}\ge i^{r_*-1}$, $\log x\ge(\log i)/2$, and $r_*+\beta/\log x\ge r_*$. Thus
\begin{align*}
 \Delta_i&=\int_{i-1}^i x^{r_*-1}(\log x)^\beta
                         \left(r_*+\frac\beta{\log x}\right)\,\dd x
 \ge r_*2^{-\beta}i^{r_*-1}(\log i)^\beta,\\
 \frac{d_i}{\Delta_i}
 &\le\frac{2^\beta}{r_*(\log2)^\alpha}
              \frac{i^{1-r_*-\alpha}}{(\log i)^\beta}
 =\frac{2^\beta}{r_*(\log2)^\alpha}\frac1{i(\log i)^\beta},\\
 \sum_{i\ge i_0}\frac1{i(\log i)^\beta}
 &\le\int_{i_0-1}^\infty\frac{\dd x}{x(\log x)^\beta}
 =\frac{[\log(i_0-1)]^{1-\beta}}{\beta-1}<\infty.
\end{align*}
Set $S_0=\sum_{i<i_0}\Delta_i$ and $C_{\mathrm{alloc}}=1+S_0$. The prefixes obey
\[
 \sum_{i=1}^m\Delta_i\le
 \begin{cases}
 S_0\le C_{\mathrm{alloc}}G(m),&m<i_0,\\
 S_0+H(m)-H(i_0-1)\le S_0+G(m)\le C_{\mathrm{alloc}}G(m),&m\ge i_0.
 \end{cases}
\]
Consequently
\[
 a_i=\Delta_i/C_{\mathrm{alloc}},\qquad
 \sum_{i=1}^ma_i\le G(m),\qquad
 \sum_i\frac{d_i}{a_i}=C_{\mathrm{alloc}}\sum_i\frac{d_i}{\Delta_i}<\infty.
\]
Theorem~\ref{thm:main} proves sufficiency.

For \eqref{eq:critical-two}, the little-omega bound excludes $0\le\gamma\le2$. If $\gamma>2$, use $H(x)=(\log x)^\gamma$, $G(m)=H(\max\{m,m_0\})$ with $m_0\ge3$ and $H(m_0)\ge1$. Choose $i_0\ge m_0+1$, let $\Delta_i=H(i)-H(i-1)$ for $i\ge i_0$, and choose positive earlier $\Delta_i$. At $T=2^m$, the profile in \eqref{eq:critical-two} satisfies $[\log(m\log2)]^\gamma\asymp H(m)$. Then
\begin{align*}
 \Delta_i&=\int_{i-1}^i\frac{\gamma(\log x)^{\gamma-1}}x\,\dd x
 \ge\gamma2^{1-\gamma}\frac{(\log i)^{\gamma-1}}i,\\
 \frac{d_i}{\Delta_i}
 &\le\frac{2^{\gamma-1}}{\gamma(\log2)^2}
                      \frac1{i(\log i)^{\gamma-1}},\\
 \sum_{i\ge i_0}\frac1{i(\log i)^{\gamma-1}}
 &\le\int_{i_0-1}^\infty\frac{\dd x}{x(\log x)^{\gamma-1}}
 =\frac{[\log(i_0-1)]^{2-\gamma}}{\gamma-2}<\infty.
\end{align*}
For these increments, set $S_0=\sum_{i<i_0}\Delta_i$, $C_{\mathrm{alloc}}=1+S_0$ and $a_i=\Delta_i/C_{\mathrm{alloc}}$. For $m<i_0$ the increment sum is at most $S_0$, and for $m\ge i_0$ it equals $S_0+H(m)-H(i_0-1)$. Hence
\[
 \begin{aligned}
 \sum_{i=1}^m a_i
 &\le\frac{S_0+G(m)}{1+S_0}\le G(m)\qquad(m\ge1),\\
 \sum_{i\ge1}\frac{d_i}{a_i}
 &=C_{\mathrm{alloc}}\left\{\sum_{i<i_0}\frac{d_i}{\Delta_i}
                  +\sum_{i\ge i_0}\frac{d_i}{\Delta_i}\right\}<\infty.
 \end{aligned}
\]
Theorem~\ref{thm:main} supplies the policy and all positive lifetime-cost moments. Equation~\eqref{eq:all-horizons} extends its dyadic excess bound to the stated profile at arbitrary horizons.
\end{proof}

\section{Scope of the allocation reduction}\label{app:scope}
We isolate the probabilistic inputs to the allocation reduction, then give counterexamples delimiting the excess profiles and comparator risks covered by the characterization.

\subsection{Probabilistic premises for the reduction}\label{app:metric}
The allocation reduction uses the following three probabilistic premises. In this subsection $A_n,B_n,w$ denote metric-space points and use ordinary type. Work on a probability space $(\Omega,\mathcal A,\Pp)$ with an observation filtration $(\G_n)_{n\ge0}$. Let $(K,\dist_K)$ have diameter at most one. Let $B_n$ be $\G_n$-measurable reference actions and let $\ell_n(w)$ be jointly measurable nonnegative excess losses with $\ell_n(B_n)=0$. As in Section~\ref{sec:model}, enlarge this experiment by a private tape $\Xi$ independent of its original sigma-field $\mathcal A$, and put $\F_n=\G_n\vee\sigma(\Xi)$. An admissible policy consists of $K$-valued, $\F_n$-measurable actions defined at every count without an evaluation horizon as input. Expectations include the independent tape. Require all displayed random distances to be measurable. Assume common versions satisfy:
\begin{enumerate}
\item[(P1)] $\ell_n(w)\ge\kappa\dist_K(w,B_n)^2$ for all $n,w$, with fixed $\kappa>0$.
\item[(P2)] A nonnegative deterministic sequence $(g_n)$, finite $C$ and integer $n_g\ge1$ satisfy $g_n\le C/n$ for $n\ge n_g$ and
$\E\ell_b(B_a)=g_a-g_b$ for deterministic $n_g\le a\le b$. Each fixed initialization risk $\E\ell_n(B_0)$ is finite. In particular, $g_n$ is nonincreasing for $n\ge n_g$.
\item[(P3)] Fixed $\chi,a_w,\eta>0$ and one $N_*<\infty$ work for every integer $N\ge N_*$, every $1\le M\le\lfloor N/6\rfloor$, and every collection of deterministic integer endpoints
$N\le n_1<m_1\le n_2<\cdots\le n_M<m_M<2N$ with $m_\ell-n_\ell\ge N/(6M)$: there is a deterministic subset $I$ with $|I|\ge\chi M$ such that
$\Pp\{\dist_K(B_{m_\ell},B_{n_\ell})\ge a_w/\sqrt{MN}\}\ge\eta$ for $\ell\in I$.
\end{enumerate}
\paragraph{Transfer principle.} Define $\EE_T(A)=\sum_{n<T}\E\ell_n(A_n)$ and use $\dist_K$ in adjustment charges and revision counts. Under (P1)--(P3), for every $(c,G)\in\CC\times\HH$, the four conditions of Theorem~\ref{thm:main} are equivalent with these metric quantities. In condition (iv), the policy retains the reference actions $B_n$ at deterministic checkpoints. The profile and revision consequences of Section~\ref{sec:consequences} also follow. In particular, (P2) supplies the risk-energy identity used by the checkpoint construction, while (P3) supplies the selected-window motion used by the all-policy converse. This transfer does not require a posterior or a Euclidean gradient in the metric model. In the original Euclidean experiment, $\ell_n(\w)=r_n(\w)-r_n(\B_n)$ and $\dist_K(\x,\y)=\norm{\x-\y}$. Assumptions~\ref{ass:curvature}--\ref{ass:information} imply (P1)--(P3) with $\chi=1/2$.

\begin{proof}[Verification of the transfer]
Choose \(j_*\ge4\) with \(2^{j_*}\ge\max\{n_g,N_*,12\}\)
and reduce \(a_w\) to \(\min\{a_w,1\}\).
Premises (P1)--(P2) give
\(0\le\kappa\E\dist_K(B_b,B_a)^2
\le\E\ell_b(B_a)=g_a-g_b\)
for \(n_g\le a\le b\). By (P1),
\[
 \kappa\sum_{j=0}^J Q_j
 \le\EE_{2^{J+1}}(A),\qquad
 Q_j=\sum_{n=2^j}^{2^{j+1}-1}
 \E\dist_K(A_n,B_n)^2.
\]
The geometric step in Lemma~\ref{lem:window-charge}
holds in any metric space:
\begin{align*}
 \dist_K(x_p,x_0)
 &\le\sum_{l=1}^p\dist_K(x_l,x_{l-1})\\
 &\le p\max_l\dist_K(x_l,x_{l-1}),\\
 \sum_{l=1}^pc(\dist_K(x_l,x_{l-1}))
 &\ge c\!\left(\dist_K(x_p,x_0)/p\right).
\end{align*}
Thus its deterministic endpoint selection, Markov bounds and (P3) give \eqref{eq:window-charge}, with $C_w=96/a_w^2$ and the $\chi$ of (P3). Starting at $j_0=j_*$, the rounding, density and scale conversion in Section~\ref{sec:proof-converse} establish necessity.

For sufficiency, retain $B_0$ before $2^{j_*}$ and use the mesh \eqref{eq:mesh}. The initialization risk is finite by (P2). On each block $[a,b)\subset[2^j,2^{j+1}]$,
\begin{align*}
 \sum_{n=a}^{b-1}\E\ell_n(B_a)
 &=\sum_{n=a}^{b-1}(g_a-g_n)\le(b-a)(g_a-g_b),\\
 \Pp\{\dist_K(B_b,B_a)>2^{-j/4}\}
 &\le2^{j/2}\E\dist_K(B_b,B_a)^2
 \le(C/\kappa)2^{-j/2}.
\end{align*}
These are exactly the block-loss and exceptional-jump inputs of Section~\ref{sec:proof-attainment}. Its integer rounding, ordinary-charge grouping, exit-edge accounting and finite-sum moment argument are scalar, so apply unchanged. Starting later discards finitely many epochs and uses each allowance at most four times.

For revisions, apply the local inequality with $c_{\mathrm{sw}}(x)=\ind_{\{x>0\}}$. The resulting \eqref{eq:revision-window} and \eqref{eq:revision-local} are the only stochastic inputs to Appendix~\ref{app:revisions}. The profile criteria use the deterministic allocation in Appendix~\ref{app:allocation}. Lemma~\ref{lem:tail} uses the same window and rounding bounds, so the critical-profile proofs also apply.
\end{proof}

\subsection{Why unrestricted profile growth fails}\label{app:unrestricted}
For $c(x)=\ind_{\{x>0\}}$ and $G(m)=m^3$, the allocation $a_i=i^2$ satisfies
$\sum_{i\le m}a_i\le m^3$ and $\sum_i1/a_i<\infty$. But $G\notin\HH$, and feasibility need not follow.
Return to the scalar uniform-prior Bernoulli--Brier model. Finite expected fixed-fee cost implies finitely many action changes almost surely. Define $A_\infty$ as the eventual constant on that probability-one event and assign it any fixed value on the complement. Then $A_n\to A_\infty$ almost surely. The posterior of $p$ after any finite history and independent tape is atomless. Since $A_n$ is $\F_n$-measurable, $\Pp(p=A_n\mid\F_n)=0$ almost surely, hence $\Pp(p=A_n)=0$. Write $N_A(\infty)=\lim_{T\to\infty}N_A(T)$. On the finite-change event, $A_\infty=A_n$ for some finite integer $n$. Therefore
\[
 \{p=A_\infty\}\cap\{N_A(\infty)<\infty\}
 \subseteq\bigcup_{n\ge0}\{p=A_n\},
\]
and the event on the left has probability zero. Since $m_n=(S_n+1)/(n+2)$ and the conditional strong law gives $S_n/n\to p$ almost surely, also $m_n\to p$ almost surely. Bounded convergence and Ces\`aro averaging yield
\[
 \frac{\EE_T(A)}T=\frac1T\sum_{n<T}\E(A_n-m_n)^2
 \longrightarrow\E(A_\infty-p)^2>0.
\]
Such a policy cannot have $\EE_{2^J}=O(J^3)$. Thus allocation finiteness is insufficient for arbitrary nondecreasing profiles.

\subsection{Population curvature does not control hindsight regret}\label{app:empirical}
Use scalar actions and witnesses. Let $K=[0,1]$, $\theta\sim\Unif[1/4,3/4]$, and $Z=(Y,V)$ with independent $Y\sim\Bernoulli(\theta)$ and $V\sim\Unif[0,1]$ conditional on $\theta$. Represent finite subsets of $[0,1]$ by
\[
 \mathsf{Fin}([0,1])
 =\{\varnothing\}\sqcup\bigsqcup_{k\ge1}
 \{(v_1,\ldots,v_k)\in[0,1]^k:v_1<\cdots<v_k\}.
\]
Each tuple component is Borel, so this countable disjoint union is an uncountable standard Borel space. The Borel isomorphism theorem \citep[Theorem~15.6]{kechris1995} gives a Borel bijection $\iota:\mathsf{Fin}([0,1])\to(0,1/4)$ with Borel inverse. Identify a tuple with its set of entries. Put $S_w=\iota^{-1}(w)$ for $w\in(0,1/4)$ and $S_w=\varnothing$ elsewhere. On the component of length $k$, membership is the finite union of the coordinate equalities $v=v_r$, $1\le r\le k$. Composing this Borel relation with $\iota^{-1}$ makes $\{(w,v):v\in S_w\}$ Borel. Define
\[
 f_{(y,v)}(w)=(w-y)^2+3-3\ind_{\{v\in S_w\}},\qquad
 R_\theta(w)=(w-\theta)^2+\theta(1-\theta)+3.
\]
The loss is jointly measurable and takes values in $[0,4]$. Population curvature and smoothness hold with $\mu=L=2$, $b_\theta=\theta$ is interior, and $\xi_{(y,v)}(w)=2(w-y)$ is a bounded unbiased witness. The tags $V$ are independent of $\theta$ and add no parameter information. Proposition~\ref{prop:local-score} therefore gives $q_n\ge v_0/n$. Thus all main assumptions hold.

At an admissible action $A_n$, the next uniform tag belongs to its finite decoded set with conditional probability zero. Expected Bayes loss is therefore at least three. Sorting the observed tags and deleting repeated entries is a Borel map into $\mathsf{Fin}([0,1])$. Thus $w_T=\iota(\{V_1,\ldots,V_T\})$ is a measurable hindsight choice and gives $f_{Z_t}(w_T)=(w_T-Y_t)^2\le1$ at every observed point. Hence
\[
 \E\left[\sum_{n<T}f_{Z_{n+1}}(B_n)
       -\inf_{w\in K}\sum_{t=1}^Tf_{Z_t}(w)\right]\ge3T-T=2T.
\]
To justify the expectation of the empirical infimum, put $L_T(z_{1:T},w)=\sum_{t=1}^Tf_{z_t}(w)$ and $I_T(z_{1:T})=\inf_{w\in K}L_T(z_{1:T},w)$.
For every $a\in\R$,
\[
 \{z_{1:T}:I_T(z_{1:T})<a\}
 =\operatorname{proj}_{\mathcal Z^T}
   \{(z_{1:T},w):L_T(z_{1:T},w)<a\}.
\]
Here $\operatorname{proj}_{\mathcal Z^T}$ denotes projection onto the data coordinates. The set being projected is Borel. Its projection is analytic and hence universally measurable \citep[Theorem~21.10]{kechris1995}. Since $0\le I_T\le4T$, the expectation is finite in the completed experiment. The Bayes-relative characterization therefore coexists with linear expected regret against the empirical comparator.

\section{Exact Brier calibration}\label{app:calibration}
For the scalar model of Section~\ref{sec:calibration}, define realized regret against the best fixed action in hindsight by
\[
 \Reg_T(A)=\sum_{n=0}^{T-1}(A_n-Y_{n+1})^2
       -\min_{w\in[0,1]}\sum_{t=1}^T(w-Y_t)^2.
\]
Its expectation satisfies
\begin{equation}\label{eq:brier-calibration}
 \E\Reg_T(A)=\frac{H_{T+1}}6+\EE_T(A),\qquad H_t=\sum_{i=1}^t i^{-1}.
\end{equation}
The calculation below includes policies with independent private randomization.

The success count $S_n$ has
\begin{align*}
 \Pp(S_n=s)
 &=\binom ns\int_0^1p^s(1-p)^{n-s}\,\dd p
 =\binom ns\frac{s!(n-s)!}{(n+1)!}=\frac1{n+1},\\
 p\mid S_n=s&\sim\Beta(s+1,n-s+1),\\
 m_n=\E[p\mid S_n]&=\frac{S_n+1}{n+2},\qquad
 \Var(p\mid S_n=s)=\frac{(s+1)(n-s+1)}{(n+2)^2(n+3)}.
\end{align*}
Since $R_p(w)=w^2-2pw+p$, one has $r_n(w)-r_n(m_n)=(w-m_n)^2$ and $B_n=m_n$. Also,
\begin{align*}
 \sum_{s=0}^n(s+1)(n-s+1)
 &=(n+2)\sum_{i=1}^{n+1}i-\sum_{i=1}^{n+1}i^2
 =\frac{(n+1)(n+2)(n+3)}6,\\
 g_n=q_n=\E\Var(p\mid\G_n)
 &=\frac{\sum_{s=0}^n(s+1)(n-s+1)}{(n+1)(n+2)^2(n+3)}
 =\frac1{6(n+2)}.
\end{align*}
Let $\widehat p_T=T^{-1}\sum_{t\le T}Y_t\in[0,1]$. Conditional on $p$, the next observation is independent of $\F_n$, so
\begin{align*}
 \E_p[(A_n-Y_{n+1})^2\mid\F_n]
 &=(A_n-p)^2+2(A_n-p)\E_p[p-Y_{n+1}\mid\F_n]
                    +\E_p[(p-Y_{n+1})^2\mid\F_n]\\
 &=(A_n-p)^2+p(1-p),\\
 \sum_{t\le T}(Y_t-w)^2
 &=\sum_{t\le T}(Y_t-\widehat p_T)^2+T(w-\widehat p_T)^2
 \quad(w\in[0,1]),\\
 \E_p(\widehat p_T-p)^2
 &=T^{-2}\sum_{s,t\le T}\E_p[(Y_s-p)(Y_t-p)]=p(1-p)/T,\\
 \E_p\min_{w\in[0,1]}\sum_{t\le T}(Y_t-w)^2
 &=\sum_{t\le T}\E_p(Y_t-p)^2-T\E_p(\widehat p_T-p)^2\\
 &=Tp(1-p)-p(1-p)=(T-1)p(1-p),\\
 \E_p\Reg_T(A)&=\sum_{n<T}\E_p(A_n-p)^2+Tp(1-p)-(T-1)p(1-p)\\
 &=\sum_{n<T}\E_p(A_n-p)^2+p(1-p).
\end{align*}
Tape independence gives $\E[p\mid\F_n]=m_n$. Therefore
\begin{align*}
 \E[(A_n-m_n)(p-m_n)]
 &=\E[(A_n-m_n)\E[p-m_n\mid\F_n]]=0,\\
 \E(A_n-p)^2&=\E(A_n-m_n)^2+\E(p-m_n)^2,\\
 \E \Reg_T(A)&=\int_0^1p(1-p)\,\dd p+\sum_{n<T}\E(A_n-p)^2\\
 &=\frac16+\sum_{n<T}\left\{\E(A_n-m_n)^2+\frac1{6(n+2)}\right\}
 =\frac{H_{T+1}}6+\EE_T(A).
\end{align*}

\paragraph{Linear adjustment charges.}
Let $c_{\mathrm{lin}}(x)=x$ on $[0,1]$. For $n\ge1$, the predictive law of $Y_n$ given $\G_{n-1}$ is $\Bernoulli(B_{n-1})$. The posterior recursion and conditional absolute deviation give
\begin{align*}
 B_n-B_{n-1}&=\frac{Y_n-B_{n-1}}{n+2},\\
 \E[|B_n-B_{n-1}|\mid\G_{n-1}]
 &=\frac{2B_{n-1}(1-B_{n-1})}{n+2},\\
 \E[B_{n-1}(1-B_{n-1})]
 &=\E[p(1-p)]+\E\Var(p\mid\G_{n-1})\\
 &=\frac16+q_{n-1}=\frac{n+2}{6(n+1)},\\
 \E|B_n-B_{n-1}|
 &=\frac{2}{n+2}\frac{n+2}{6(n+1)}=\frac1{3(n+1)}.
\end{align*}
Thus the policy that always issues the current Bayes action has, for $T\ge1$,
\[
 \E V_{c_{\mathrm{lin}},T}(B)
 =\sum_{n=1}^{T-1}\E|B_n-B_{n-1}|
 =\frac13\sum_{k=2}^{T}\frac1k
 =\frac{H_T-1}{3}\longrightarrow\infty.
\]
For the same charge, \(\DiniHalf(c_{\mathrm{lin}})=\int_0^1x^{-1/2}\,\dd x=2\). Corollary~\ref{cor:dini} and Theorem~\ref{thm:main} therefore supply one horizon-independent deterministic-checkpoint policy \(A\) with \(\EE_\infty(A)<\infty\) and \(\E[V_{c_{\mathrm{lin}},\infty}(A)^q]<\infty\) for every \(q>0\). By \eqref{eq:brier-calibration}, this policy also satisfies \(\E\Reg_T(A)=H_{T+1}/6+O(1)\).

\section{Population interiority without predictive stationarity}\label{app:boundary}
This example satisfies population interiority (I) and action information but not predictive stationarity (S). Let \(p\) have density \(1/[2(b-a)]\) on \([a,b]\cup[1-b,1-a]\), where \(a=1/100\) and \(b=1/50\), and let \(Z_t\mid p\) be iid \(\Bernoulli(p)\). Define symmetric matrices \(\mathbf M_1\) and \(\mathbf M_0\) with diagonal \((1,1/3)\) and off-diagonal entries \(1/2\) and \(-1/2\), respectively. Set \(\uvec_1=(3/4,0)^\top\), \(\uvec_0=-\uvec_1\), and \(K=[-1,1]^2/(2\sqrt2)\), and define
\[
 f_z(\w)=\frac16(2\sqrt2\w-\uvec_z)^\top\mathbf M_z(2\sqrt2\w-\uvec_z).
\]
With $D(p)=1+12p(1-p)$, the population optimizer is
\begin{equation}\label{eq:boundary-target}
 \bb_p=\frac1{2\sqrt2}
       \begin{pmatrix}3(2p-1)/(4D(p))\\18p(1-p)/D(p)\end{pmatrix}.
\end{equation}
Use physical coordinates $\x=2\sqrt2\w$, while gradients remain in normalized coordinates unless indicated. For $\varepsilon_z=2z-1$,
\(
 \y^\top\mathbf M_z\y=(y_1+\varepsilon_zy_2/2)^2+y_2^2/12.
\)
Both matrices are positive definite. Since $\nabla_\w^2f_z=(8/3)\mathbf M_z$ and there are two labels, uniform positive sample curvature and finite smoothness follow. The action set has diameter one. For $\y=\x-\uvec_z$, $|x_i|\le1$ and $|u_{z,1}|=3/4$ imply
\(
 0\le6f_z(\w)\le(1+3/4+1/2)^2+1/12<6.
\)
Thus $0\le f_z\le1$. The affine gradient
$\nabla_\w f_z(\w)=(2\sqrt2/3)\mathbf M_z(\x-\uvec_z)$ is bounded on the compact set, supplying a finite sample Lipschitz constant and a bounded gradient witness.

\Needspace{5\baselineskip}
\subsection{Population and predictive optima}
Write $\delta=2p-1$ and $\mathbf M(p)=(1-p)\mathbf M_0+p\mathbf M_1$, with diagonal $(1,1/3)$ and off-diagonal $\delta/2$. Set
$\hvec(p)=(1-p)\mathbf M_0\uvec_0+p\mathbf M_1\uvec_1=(3\delta/4,3/8)^\top$.
The population stationary equation in physical coordinates is $\mathbf M(p)\x=\hvec(p)$.
Since $\det\mathbf M(p)=1/3-\delta^2/4=D(p)/12>0$,
\begin{align*}
 \x^*(p)&=\mathbf M(p)^{-1}\hvec(p)
 =\frac{12}{D(p)}
       \begin{pmatrix}1/3&-\delta/2\\-\delta/2&1\end{pmatrix}
       \begin{pmatrix}3\delta/4\\3/8\end{pmatrix}\\
 &=\frac1{D(p)}\begin{pmatrix}3\delta/4\\(9/2)(1-\delta^2)\end{pmatrix}
 =\frac1{D(p)}\begin{pmatrix}3(2p-1)/4\\18p(1-p)\end{pmatrix}.
\end{align*}
Division by $2\sqrt2$ yields \eqref{eq:boundary-target}. On the prior support, $s=p(1-p)$ obeys $0<s\le b(1-b)$ and $6b(1-b)<1$, so
\[
 |x^{*,1}(p)|=\frac{3|2p-1|}{4D(p)}<1,\qquad
 0<x^{*,2}(p)=\frac{18s}{1+12s}<1.
\]
The normalized optimizer $\bb_p=\x^*(p)/(2\sqrt2)$ lies in $\ri K$ and is therefore the constrained optimizer.

The risk is affine in $p$, so its predictive stationary equation replaces $p$ by $m_n=\E[p\mid\G_n]$. It need not produce the posterior mean of $\bb_p$. At even $n$, on the event of $n/2$ successes, both likelihood $[p(1-p)]^{n/2}$ and prior are symmetric about $1/2$. Hence $m_n=1/2$. This event has positive probability
$\binom n{n/2}\int[p(1-p)]^{n/2}\pi(\dd p)$ at every finite even $n$. Its unconstrained physical optimizer is \((0,9/8)^\top\). At \(\x=(0,1)^\top\), we have \(\mathbf M(1/2)\x-\hvec(1/2)=(0,-1/24)^\top\) and \(\nabla_\w r_n\big((0,1)^\top/(2\sqrt2)\big)=(0,-\sqrt2/36)^\top\).
For every $\w\in K$, this gradient has nonnegative inner product with
$\w-(0,1)^\top/(2\sqrt2)$. Strong convexity and this variational inequality identify the boundary minimizer, with nonzero gradient. Predictive stationarity therefore fails with positive probability at arbitrarily large deterministic counts.

\subsection{Action information from the local-score criterion}
Use Proposition~\ref{prop:local-score} on $U=(a,b)$ with parameter direction $v=1$. The prior density there is $\varpi(p)=1/[2(b-a)]$. For the Bernoulli likelihood, relative to counting measure on $\{0,1\}$,
\[
 s(p,Z)=\frac{Z-p}{p(1-p)},\qquad
 \E_p s(p,Z)=0,\qquad
 \E_p s(p,Z)^2=\frac1{p(1-p)}\le\frac1{a(1-b)}.
\]
Both the likelihood and the bounded action map \eqref{eq:boundary-target} are $C^1$ on $U$. If $\vartheta(p)$ is the first coordinate of $\bb_p$, differentiation gives
\[
 \vartheta'(p)=\frac{3\{7-12p(1-p)\}}{4\sqrt2D(p)^2}>0.
\]
Proposition~\ref{prop:local-score} therefore gives a fixed \(v_0>0\) such that \(q_n=\E\norm{\bb_p-\E[\bb_p\mid\G_n]}^2\ge v_0/n\) for all \(n\ge1\).
Thus the example also satisfies Assumption~\ref{ass:information}.

\Needspace{5\baselineskip}
\section{Local-score and model verifications}\label{app:models}
The local score argument proves the action-information criterion of Section~\ref{sec:applications}. The remaining subsections verify its hypotheses and the other statistical assumptions for each model.

\subsection{A compactly supported score identity}\label{app:local-score}
\begin{proof}[Proof of Proposition~\ref{prop:local-score}]
Let $\widehat\psi_n$ be any bounded measurable scalar estimator based on $(Z_{1:n},\Xi)$, where $Z_{1:n}=(Z_1,\ldots,Z_n)$. We localize integration by parts to a direction in which the action changes, using the original prior.

\emph{Localized score.} Let $\bm{\zeta}=\Jac\bb(\theta_0)\vvec/\norm{\Jac\bb(\theta_0)\vvec}$ and
$\psi(\theta)=\bm{\zeta}^\top\bb_\theta$.
Continuity gives an open ball $V$ with compact closure $\overline V\subset U$ and
$\inf_V\partial_\vvec\psi>0$. Choose a nonnegative nonzero $h\in C_c^1(V)$, the space of continuously differentiable functions compactly supported in $V$. Extend $h$ by zero outside $V$ and put $a_h=\partial_\vvec h+h\partial_\vvec\log\varpi$. Define
\[
 J_h=\int_Ua_h^2\varpi\,\dd\theta,\qquad
 I_h=\int_Uh^2I_\vvec\varpi\,\dd\theta,\qquad
 m_h=\int_Uh\partial_\vvec\psi\,\varpi\,\dd\theta>0.
\]
The compact set $K_h=\supp h\subset U$ and the local density and information bounds give $J_h,I_h<\infty$. Extend $a_h$ by zero outside $U$. For $\theta\in U$, write $z_{1:n}=(z_1,\ldots,z_n)$ and let $\nu^{\otimes n}$ be the product reference measure. With $L_\theta(z_{1:n})=\prod_{t=1}^np_\theta(z_t)$, define
\[
 T_n=a_h(\theta)+h(\theta)\sum_{t=1}^ns_\vvec(\theta,Z_t),
\]
and set $T_n=0$ outside $U$. This latent score will be paired with the estimation error. Conditional independence and centering give
\[
 \E_\theta[s_\vvec(\theta,Z_s)s_\vvec(\theta,Z_t)]
 =\begin{cases}I_\vvec(\theta),&s=t,\\
 \E_\theta s_\vvec(\theta,Z_s)\E_\theta s_\vvec(\theta,Z_t)=0,&s\ne t,
 \end{cases}
\]
\begin{align*}
 \E_\theta T_n^2
 &=a_h^2+2a_hh\sum_{t=1}^n\E_\theta s_\vvec(\theta,Z_t)
          +h^2\sum_{s,t=1}^n\E_\theta[s_\vvec(\theta,Z_s)s_\vvec(\theta,Z_t)]\\
 &=a_h^2+nh^2I_\vvec(\theta),\\
 \E T_n^2&=\int_U(a_h^2+nh^2I_\vvec)\varpi\,\dd \theta=J_h+nI_h.
\end{align*}
\emph{Integration by parts.} For fixed data $z_{1:n}$ and tape value $\upsilon$, define
$\varepsilon_n(\theta,z_{1:n},\upsilon)=\widehat\psi_n(z_{1:n},\upsilon)-\psi(\theta)$. Thus $\partial_\vvec\varepsilon_n=-\partial_\vvec\psi$.
Product differentiation and the score second moment yield
\begin{align*}
 \partial_\vvec L_\theta&=L_\theta\sum_{t=1}^ns_\vvec(\theta,z_t),\\
 \partial_\vvec(h\varpi L_\theta)
 &=\varpi L_\theta\left(a_h+h\sum_{t=1}^ns_\vvec(\theta,z_t)\right)=T_n\varpi L_\theta,\\
 \int|\partial_\vvec L_\theta|\,\dd \nu^{\otimes n}
 &=\E_\theta\left|\sum_{t=1}^ns_\vvec(\theta,Z_t)\right|\le\sqrt{nI_\vvec(\theta)},\\
 \int_U\int|\partial_\vvec(h\varpi L_\theta)|\,\dd \nu^{\otimes n}\,\dd \theta
 &\le\int_{K_h}\{|\partial_\vvec(h\varpi)|+|h|\varpi\sqrt{nI_\vvec}\}\,\dd \theta<\infty.
\end{align*}
Integrating over the parameter, observations and private tape gives, for each fixed $n$,
\begin{align*}
 &\int\!\int\!\int_U|\varepsilon_n\partial_\vvec(h\varpi L_\theta)|
                    \,\dd \theta\,\dd \nu^{\otimes n}\,\dd P_\Xi\\
 &\quad\le(\norm{\widehat\psi_n}_\infty+\norm\psi_\infty)
       \int_{K_h}\{|\partial_\vvec(h\varpi)|+|h|\varpi\sqrt{nI_\vvec}\}\,\dd \theta<\infty,\\
 &\int\!\int\!\int_U|h\varpi L_\theta\partial_\vvec\psi|
                    \,\dd \theta\,\dd \nu^{\otimes n}\,\dd P_\Xi
 =\int_{K_h}|h|\varpi|\partial_\vvec\psi|\,\dd \theta<\infty.
\end{align*}
Writing $\mathbf n_V$ for the outward normal and $\dd S$ for surface measure, integration by parts on $V$ gives
\begin{align*}
 \int_V\varepsilon_n\partial_\vvec(h\varpi L_\theta)\,\dd \theta
 &=\underbrace{\int_{\partial V}\varepsilon_nh\varpi L_\theta
                     \ip\vvec{\mathbf n_V}\,\dd S}_{=0}
       -\int_Vh\varpi L_\theta\partial_\vvec\varepsilon_n\,\dd \theta\\
 &=\int_Vh\varpi L_\theta\partial_\vvec\psi\,\dd \theta.
\end{align*}
The boundary term vanishes because $h|_{\partial V}=0$. Independence of the tape gives joint law
$L_\theta\varpi\,\dd \theta\,\dd \nu^{\otimes n}\,\dd P_\Xi$ on $U$. Fubini now yields
\begin{align*}
 \E[(\widehat\psi_n-\psi(\theta))T_n]
 &=\int\!\int\left\{\int_V\varepsilon_n\partial_\vvec(h\varpi L_\theta)\,\dd \theta\right\}
                    \nu^{\otimes n}(\dd z_{1:n})P_\Xi(\dd \upsilon)\\
 &=\int_Vh\varpi\partial_\vvec\psi
                  \left\{\int L_\theta\,\dd \nu^{\otimes n}\right\}
                  \left\{\int \dd P_\Xi\right\}\,\dd \theta=m_h.
\end{align*}
\emph{Action-information bound.} Cauchy--Schwarz gives
\[
 0<m_h^2\le\E(\widehat\psi_n-\psi)^2(J_h+nI_h),\qquad
 \E(\widehat\psi_n-\psi)^2\ge\frac{m_h^2}{J_h+nI_h}
                         \ge\frac{m_h^2}{n(J_h+I_h)}.
\]
In particular the denominator is positive, including when $I_h=0$. Take
$\widehat\psi_n=\bm{\zeta}^\top\E[\bb_\theta\mid\G_n]$. Since $\norm{\bm{\zeta}}=1$,
\begin{align*}
 q_n&=\E\norm{\bb_\theta-\E[\bb_\theta\mid\G_n]}^2\\
 &\ge\E\big[\big(\bm{\zeta}^\top
                    \{\bb_\theta-\E[\bb_\theta\mid\G_n]\}\big)^2\big]
 =\E(\widehat\psi_n-\psi)^2
 \ge\frac{m_h^2}{n(J_h+I_h)}.
\end{align*}
This is \eqref{eq:local-info} with $v_0=m_h^2/(J_h+I_h)>0$.
\end{proof}

\subsection{Unregularized logistic verification}\label{app:logistic}
Write $m_1=\E_Q\norm\X\le(\E_Q\norm\X^2)^{1/2}$ and $m_2=\E_Q\norm\X^2<\infty$. For $y\in\{0,1\}$,
\begin{align*}
 \ell_{\x,y}(\w)&=\log(1+e^{(1-2y)\x^\top\w}),\\
 \nabla_\w\ell_{\x,y}(\w)&=\x\{\sigma(\x^\top\w)-y\},\\
 \nabla_\w^2\ell_{\x,y}(\w)&=\sigma'(\x^\top\w)\x\x^\top.
\end{align*}
On a compact action neighborhood of radius $R_+$, the loss is bounded by $\log2+R_+\norm\X$. For $\norm\vvec=1$ and $t\ne0$,
\begin{align*}
 \left|\frac{\ell_{\x,y}(\w+t\vvec)-\ell_{\x,y}(\w)}t\right|
 &=\left|\int_0^1\ip{\nabla\ell_{\x,y}(\w+st\vvec)}\vvec\,\dd s\right|\le\norm\x,\\
 \norm{\frac{\nabla\ell_{\x,y}(\w+t\vvec)-\nabla\ell_{\x,y}(\w)}t}
 &=\norm{\int_0^1\nabla^2\ell_{\x,y}(\w+st\vvec)\vvec\,\dd s}\le\norm\x^2.
\end{align*}
These integrable majorants justify both differentiations. Conditioning on $\X$ gives
\begin{align*}
 \nabla R_\theta^\phys(\w)
 &=\E_Q[\X\{\sigma(\X^\top\w)-\E_\theta[Y\mid\X]\}]\\
 &=\E_Q[\X\{\sigma(\X^\top\w)-\sigma(\X^\top\theta)\}],\\
 \nabla^2 R_\theta^\phys(\w)
 &=\mathbf H(\w):=\E_Q[\sigma'(\X^\top\w)\X\X^\top].
\end{align*}
The same domination gives jointly measurable risks that are \(C^2\) on a common domain. For a unit \(\vvec\), we have
\[
\begin{aligned}
 \E_Q(\vvec^\top\X)^2
 &=\vvec^\top\E_Q[\X\X^\top]\vvec>0,\\
 \vvec^\top\mathbf H(\w)\vvec
 &=\E_Q[\sigma'(\X^\top\w)(\vvec^\top\X)^2]>0.
\end{aligned}
\]
The second inequality uses $Q(\vvec^\top\X\ne0)>0$ and positivity of $\sigma'$ at finite arguments. The quadratic form is continuous in $(\w,\vvec)$ by domination with $\norm\X^2$. With
$\mathbb{S}^{d-1}=\{\vvec\in\R^d:\norm\vvec=1\}$,
\[
 0<\mu_0:=\min_{K_0\times\mathbb{S}^{d-1}}\vvec^\top\mathbf H(\w)\vvec
 \le\max_{K_0\times\mathbb{S}^{d-1}}\vvec^\top\mathbf H(\w)\vvec=:L_0\le m_2.
\]
This proves \eqref{eq:logistic-curvature}. A compact neighborhood gives the required extensions. For the witness $\xiobs^{\phys}_Z(\w)=\nabla\ell_Z(\w)$, we have
$\norm{\xiobs^{\phys}_Z(\w)}\le\norm\X$ and
\[
 \sup_{\theta,\w}\E_\theta[\norm{\xiobs^{\phys}_Z(\w)}^2
                    \ind_{\{\norm{\xiobs^{\phys}_Z(\w)}>t\}}]
 \le\E_Q[\norm\X^2\ind_{\{\norm\X>t\}}]\longrightarrow0.
\]
For $\theta\in\interior K_0$, stationarity at $\w=\theta$ and strong convexity give $\bb_\theta^\phys=\theta$.
Relative to \(Q\) times counting measure on \(\{0,1\}\), the density \(p_\theta(\x,y)=\sigma(\x^\top\theta)^y\{1-\sigma(\x^\top\theta)\}^{1-y}\) is positive and \(C^1\) in \(\theta\). Its directional score is \(s_\vvec(\theta,(\X,Y))=(\vvec^\top\X)\{Y-\sigma(\X^\top\theta)\}\), with \(\E_\theta[s_\vvec\mid\X]=0\) and
\[
 I_\vvec(\theta)
 =\E_Q[(\vvec^\top\X)^2\sigma'(\X^\top\theta)]
 \le m_2\qquad(\norm\vvec=1).
\]
The normalized target on the prior ball is $\bb_\theta=\theta/D_{\phys}$, with derivative $\Id/D_{\phys}$. Proposition~\ref{prop:local-score} gives $q_n\ge v/n$. Normalization multiplies Hessians by $D_{\phys}^2$ and the witness by $D_{\phys}$. Substitution in the loss and path charges gives \eqref{eq:physical-dictionary}, including cost moments.

\paragraph{A covariate law with only a second moment.}\label{app:heavy-tail}
For a centered physical action ball, take \(\X=\varepsilon V\evec_I\), where \(\evec_i\) is the \(i\)th standard basis vector. The three variables are independent, with \(\Pp(\varepsilon=1)=\Pp(\varepsilon=-1)=1/2\), \(I\sim\Unif\{1,\ldots,d\}\), and \(\Pp(V=e^m)=C_V^{-1}e^{-2m}m^{-2}\) for \(m\ge1\), where \(C_V=\sum_{m\ge1}e^{-2m}m^{-2}\in(0,\infty)\). Then
\[
\begin{aligned}
 \E V^2&=C_V^{-1}\sum_{m\ge1}m^{-2}<\infty,\\
 \E V^{2+\delta}&=C_V^{-1}\sum_{m\ge1}e^{\delta m}m^{-2}=\infty
 \qquad(\delta>0).
\end{aligned}
\]
Moreover, \(\E[\X\X^\top]=(\E V^2/d)\Id_d\succ0\).
A uniform prior on a smaller concentric ball inside the physical action ball satisfies the unregularized assumptions. At $\w=0$, the unregularized witness has norm $V/2$ for either label. Adding a fixed ridge term leaves this value unchanged, since its gradient is zero there. Thus the natural witness has UI2 but no moment above two in both models.

\subsection{Fixed-ridge logistic verification under second covariate moments}\label{app:ridge}
Let \(m_1=\E_Q\norm\X\) and \(m_2=\E_Q\norm\X^2\). On \(\norm\w\le R\), an envelope is \(\log2+R\norm\X+\lambda R^2/2\), and \(\norm{\xiobs^{\phys}_Z(\w)}\le U_R:=\norm\X+\lambda R\), where \(\E_QU_R^2\le2m_2+2\lambda^2R^2<\infty\). Consequently,
\[
 \sup_{\theta,\norm\w\le R}\E_\theta[\norm{\xiobs^{\phys}_Z(\w)}^2
 \ind_{\{\norm{\xiobs^{\phys}_Z(\w)}>t\}}]
 \le\E_Q[U_R^2\ind_{\{U_R>t\}}]\longrightarrow0.
\]
The sample loss is at least $\lambda\norm\w^2/2$ and
\[
 \lambda\Id\preceq\mathbf H_\theta(\w)
 :=\lambda\Id+\E_Q[\sigma'(\X^\top\w)\X\X^\top]
 \preceq(\lambda+m_2)\Id.
\]
Thus $R_\theta^\phys$ and $r_n^\phys=\int R_\vartheta^\phys\Pi_n(\dd \vartheta)$ are continuous, coercive and $\lambda$-strongly convex on $\R^d$. Their unique global minimizers are the physical actions $\bb_\theta^\phys$ and $\B_n^\phys$ defined in Section~\ref{sec:logistic}. For \(\mathbf F(\w,\theta)=\nabla R_\theta^\phys(\w)\), we have
\begin{align*}
 \mathbf F(\w,\theta)
 &=\lambda\w+\E_Q[\X\{\sigma(\X^\top\w)-\sigma(\X^\top\theta)\}],\\
 \mathbf F(\bb_\theta^\phys,\theta)&=0,\\
 \lambda\norm{\bb_\theta^\phys}
 &=\norm{\E_Q[\X\{\sigma(\X^\top\theta)-\sigma(\X^\top\bb_\theta^\phys)\}]}\le m_1.
\end{align*}
For the posterior, define the predictive response probability
$\bar\sigma_n(\x)=\int\sigma(\x^\top\vartheta)\Pi_n(\dd \vartheta)\in[0,1]$.
The bound $\norm\X+\lambda R$ on each action ball permits integration and differentiation:
\begin{align*}
 \nabla r_n^\phys(\w)&=\int\mathbf F(\w,\vartheta)\Pi_n(\dd \vartheta)\\
 &=\lambda\w+\E_Q[\X\{\sigma(\X^\top\w)-\bar\sigma_n(\X)\}],\\
 \nabla r_n^\phys(\B_n^\phys)&=0,\\
 \lambda\norm{\B_n^\phys}&=\norm{\E_Q[\X\{\bar\sigma_n(\X)-\sigma(\X^\top\B_n^\phys)\}]}\le m_1.
\end{align*}
For $R>m_1/\lambda$, both minimizers lie in the interior of the physical action ball $K_0=\{\w:\norm\w\le R\}$ and hence also minimize the corresponding risks over $K_0$. Both boundary alternatives hold on common posterior versions, without a moment assumption on the prior parameter.

For $\norm\vvec=1$ and $t\ne0$, the exact directional differences are
\begin{align*}
 \frac{\mathbf F(\w+t\vvec,\theta)-\mathbf F(\w,\theta)}t
 &=\lambda\vvec+\E_Q\int_0^1\sigma'(\X^\top(\w+st\vvec))\X\X^\top\vvec\,\dd s,\\
 \frac{\mathbf F(\w,\theta+t\vvec)-\mathbf F(\w,\theta)}t
 &=-\E_Q\int_0^1\sigma'(\X^\top(\theta+st\vvec))\X\X^\top\vvec\,\dd s.
\end{align*}
The integrand norms are bounded by $\norm\X^2$, giving continuous partial derivatives by Fubini and dominated convergence. At the minimizer write
$\mathbf H_\theta=\mathbf H_\theta(\bb_\theta^\phys)$ and
$\mathbf I_\theta=\E_Q[\sigma'(\X^\top\theta)\X\X^\top]$. For nonzero \(\vvec\), we have
\[
 \vvec^\top\mathbf I_\theta\vvec
 =\E_Q[\sigma'(\X^\top\theta)(\vvec^\top\X)^2]>0.
\]
The partial derivatives are \(\partial_\w\mathbf F(\bb_\theta^\phys,\theta)=\mathbf H_\theta\succeq\lambda\Id\) and \(\partial_\theta\mathbf F(\bb_\theta^\phys,\theta)=-\mathbf I_\theta\). The implicit function theorem gives local \(C^1\) targets that agree on overlaps by uniqueness. Differentiating stationarity yields
\[
\begin{aligned}
 0&=\Jac_\theta\{\mathbf F(\bb_\theta^\phys,\theta)\}
   =\mathbf H_\theta\Jac\bb^\phys(\theta)-\mathbf I_\theta,\\
 \Jac\bb^\phys(\theta)&=\mathbf H_\theta^{-1}\mathbf I_\theta.
\end{aligned}
\] Both matrices are invertible, so the bounded target has a nonzero derivative in every unit direction. The likelihood score computed in Appendix~\ref{app:logistic} and the positive local prior density verify Proposition~\ref{prop:local-score}. No uniform lower bound on \(\mathbf I_\theta\) at infinity is needed. For \(\wnorm=\w/(2R)\), we have \(\Jac\bb(\theta)=(2R)^{-1}\mathbf H_\theta^{-1}\mathbf I_\theta\ne0\). Normalization multiplies the Hessians by \((2R)^2\) and transforms the witness into \(2R\xiobs^{\phys}_z(2R\wnorm)\). Physical and normalized excess agree, and \(c(x)=\widetilde c(2Rx)\) preserves charges.

\end{document}